\documentclass[10pt,twocolumn,letterpaper]{article}

\usepackage{cvpr}
\usepackage{times}
\usepackage{epsfig}
\usepackage{graphicx}
\usepackage{amsmath}
\usepackage{amssymb}
\usepackage{enumerate}
\usepackage{multirow}
\usepackage{color}

\usepackage[pagebackref=true,breaklinks=true,letterpaper=true,colorlinks,bookmarks=false]{hyperref}

\cvprfinalcopy 

\def\cvprPaperID{1776} 

\begin{document}

\title{Simultaneous Stereo Video Deblurring and Scene Flow Estimation}

\author{Liyuan Pan $^{1,2}$, Yuchao Dai$^{2}$, Miaomiao Liu$^{3,2}$ and Fatih Porikli$^{2}$ \\
$^{1}$ School of Automation, Northwestern Polytechnical University, Xi'an, China \\
$^{2}$ Research School of Engineering, Australian National University, Canberra, Australia \\
$^{3}$ Data61, CSIRO, Canberra, Australia \\\tt\small{panliyuan@mail.nwpu.edu.cn, \{yuchao.dai, miaomiao.liu, fatih.porikli\}}@anu.edu.au}

\maketitle


\begin{abstract}

Videos for outdoor scene often show unpleasant blur effects due to the large relative motion between the camera and the dynamic objects and large depth variations. Existing works typically focus monocular video deblurring. In this paper, we propose a novel approach to deblurring from stereo videos. In particular, we exploit the piece-wise planar assumption about the scene and leverage the scene flow information to deblur the image. Unlike the existing approach~\cite{sellent2016stereo} which used a pre-computed scene flow, we propose a single framework to jointly estimate the scene flow and deblur the image, where the motion cues from scene flow estimation and blur information could reinforce each other, and produce superior results than the conventional scene flow estimation or stereo deblurring methods. We evaluate our method extensively on two available datasets and achieve significant improvement in flow estimation and removing the blur effect over the state-of-the-art methods.
\end{abstract}


\section{Introduction }

\begin{figure}
\begin{center}
\begin{tabular}{cc}
\hspace{0.0cm}
\includegraphics[width=0.225\textwidth]{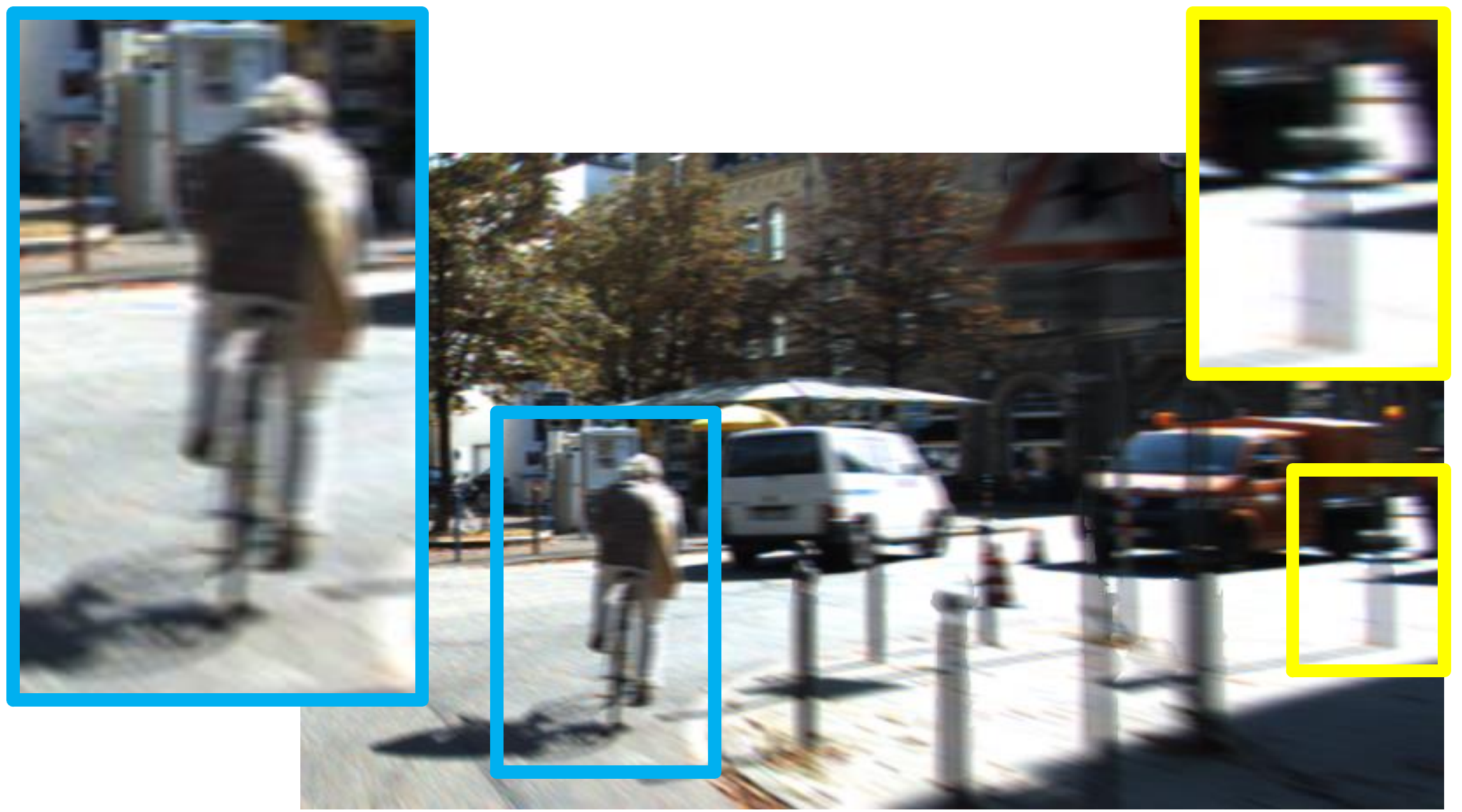}
&\hspace{0.0cm}
\includegraphics[width=0.200\textwidth]{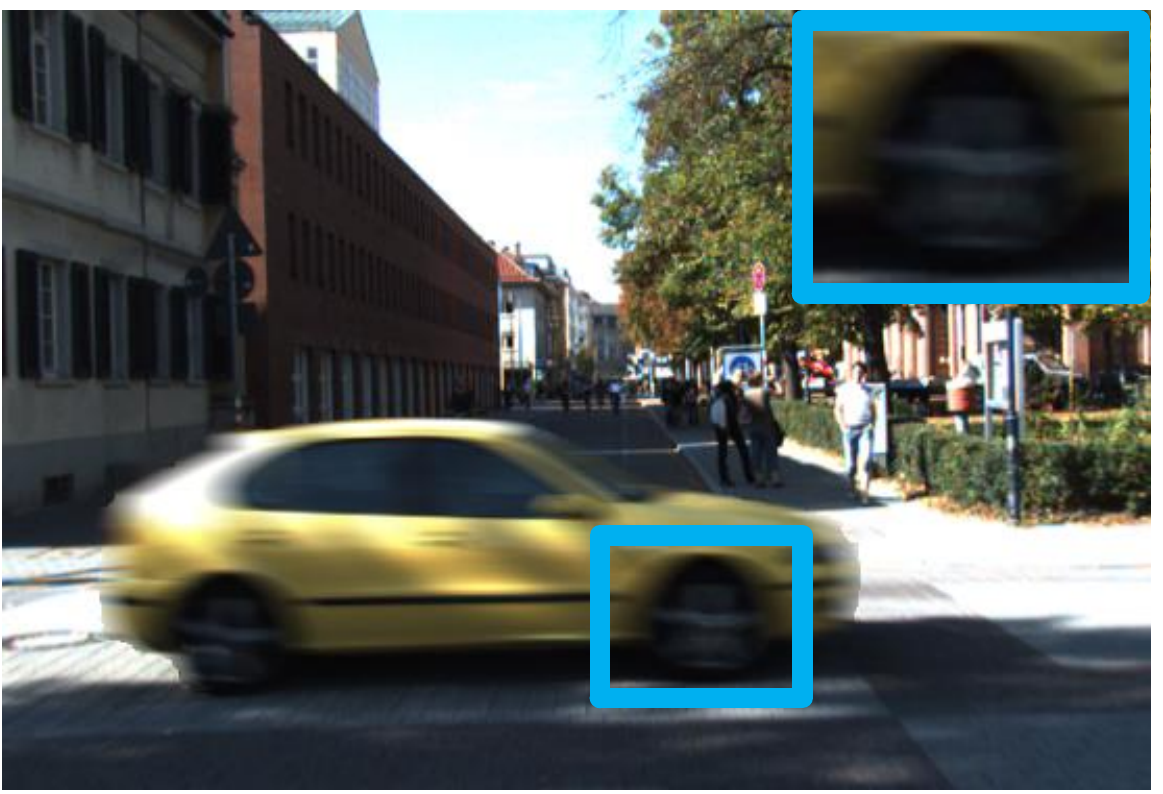}\\
\multicolumn{2}{c}{(a) Original blur images}  \\
\hspace{0.0cm}
\includegraphics[width=0.225\textwidth]{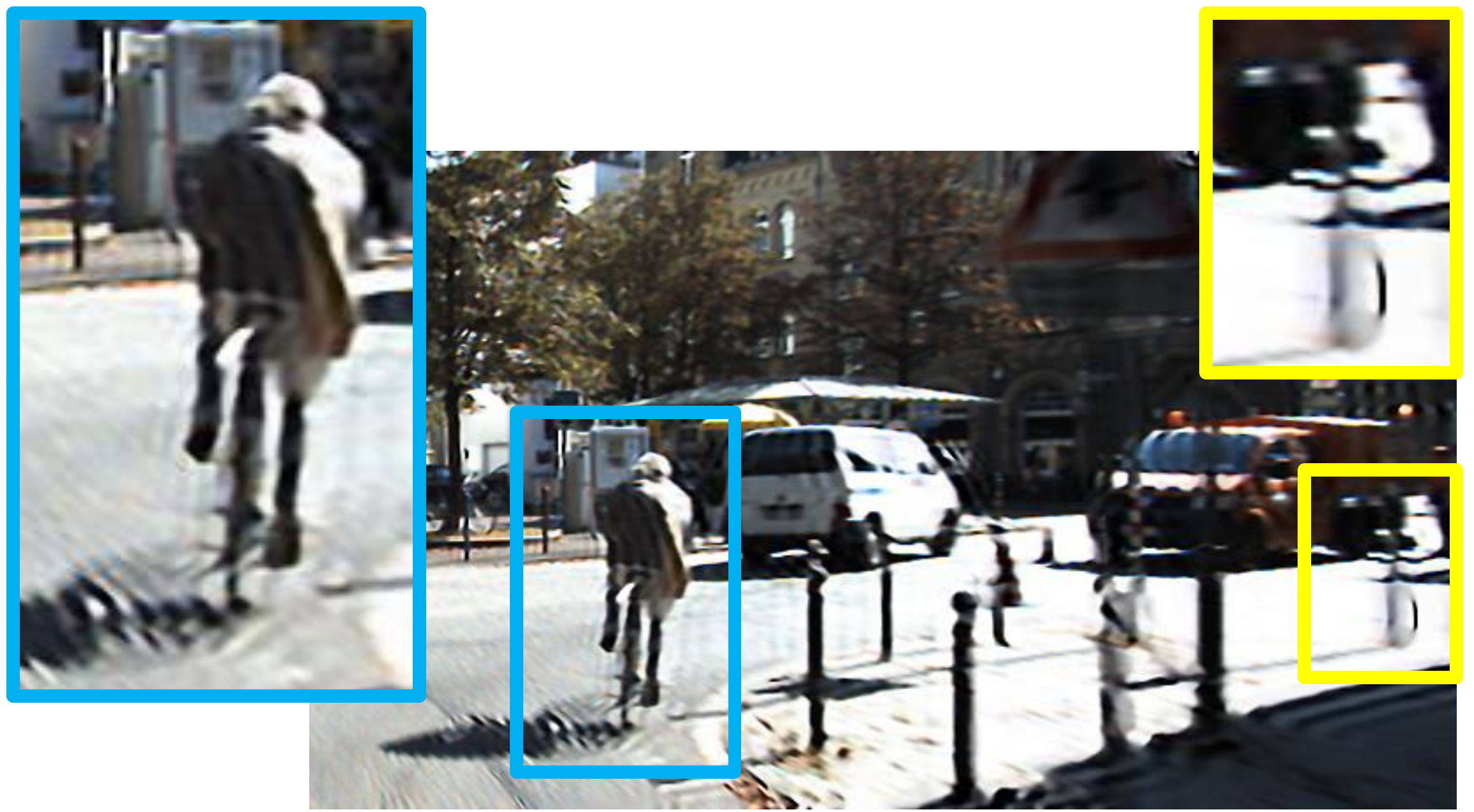}
&\hspace{0.0cm}
\includegraphics[width=0.200\textwidth]{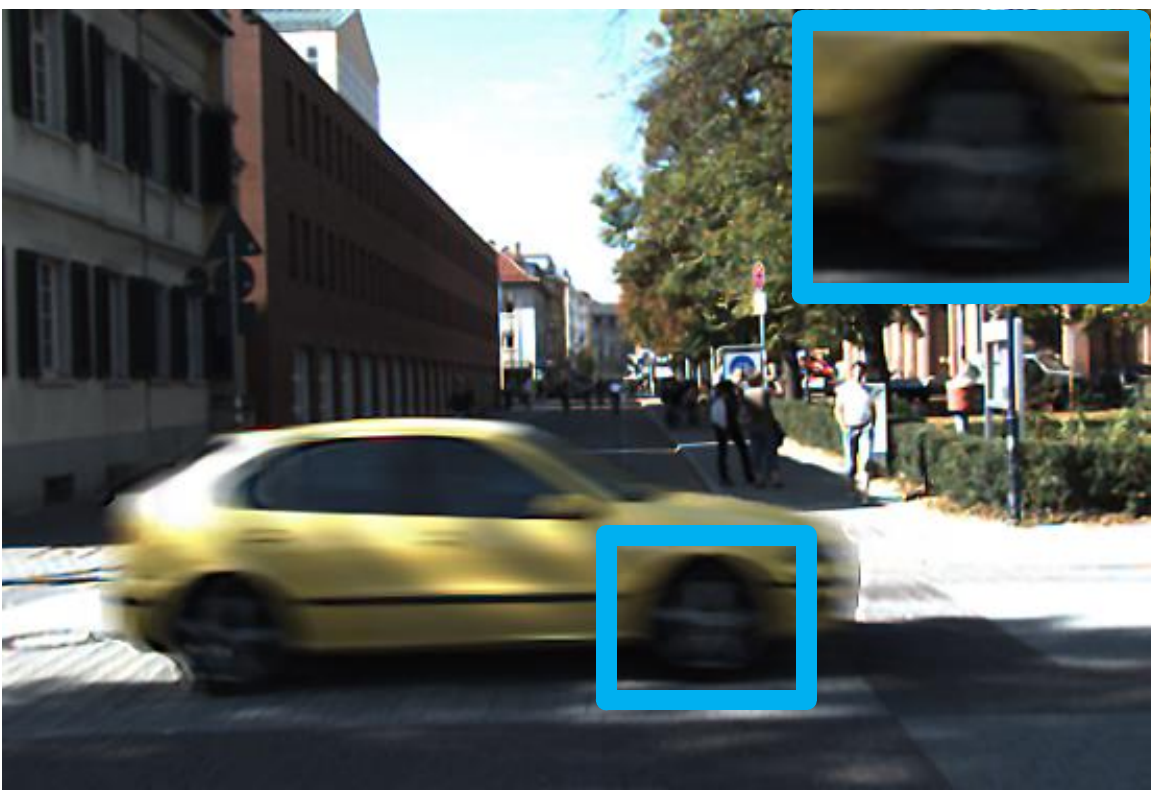}\\
\multicolumn{2}{c}{(b) Kim and Lee CVPR 2015~\cite{hyun2015generalized}}  \\
\hspace{0.0cm}
\includegraphics[width=0.225\textwidth]{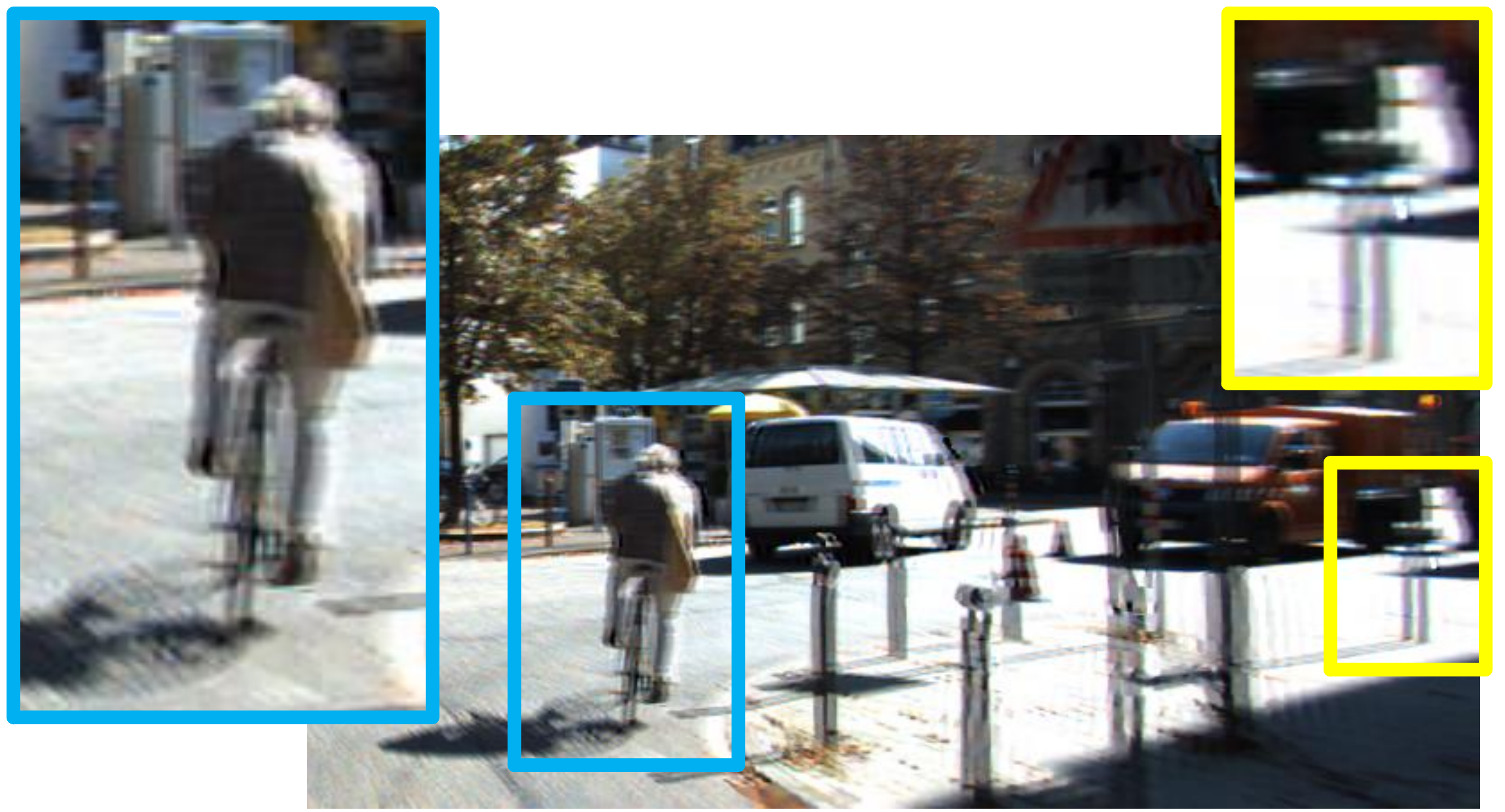}
&\hspace{0.0cm}
\includegraphics[width=0.200\textwidth]{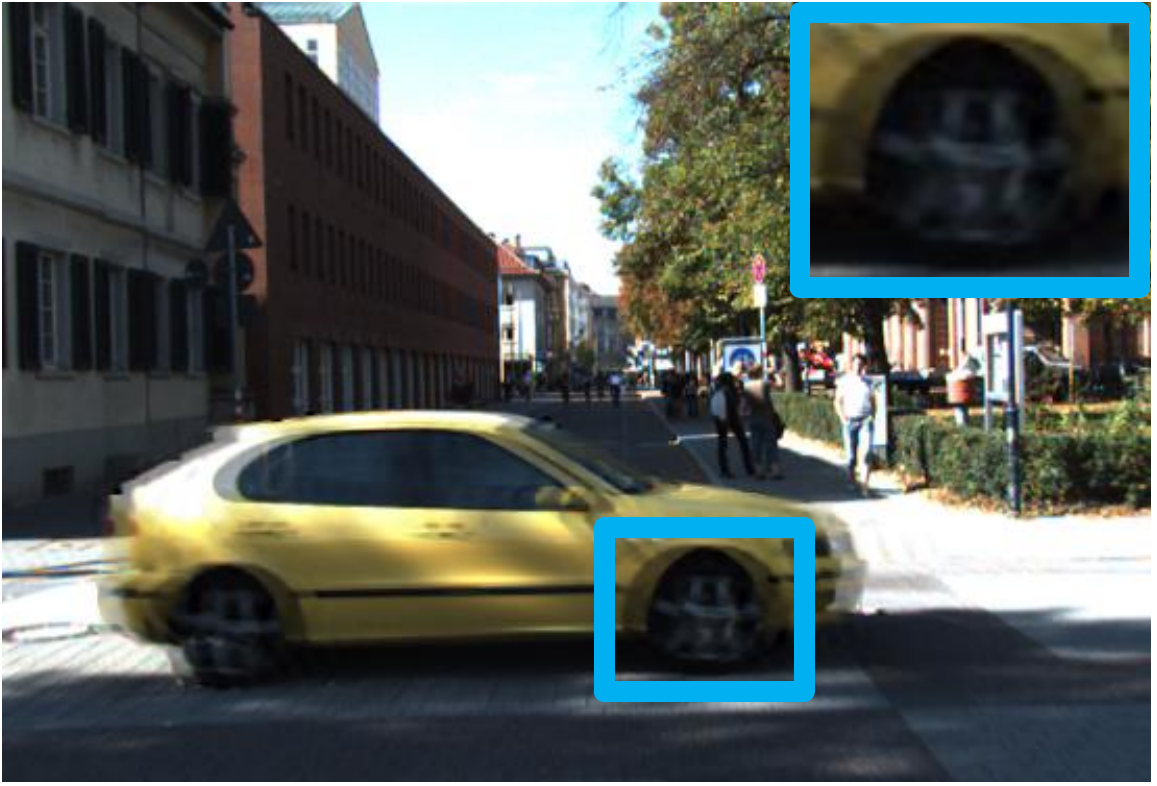}\\
\multicolumn{2}{c}{(c) Sellent \etal ECCV 2016~\cite{sellent2016stereo}} \\
\hspace{0.0cm}
\includegraphics[width=0.225\textwidth]{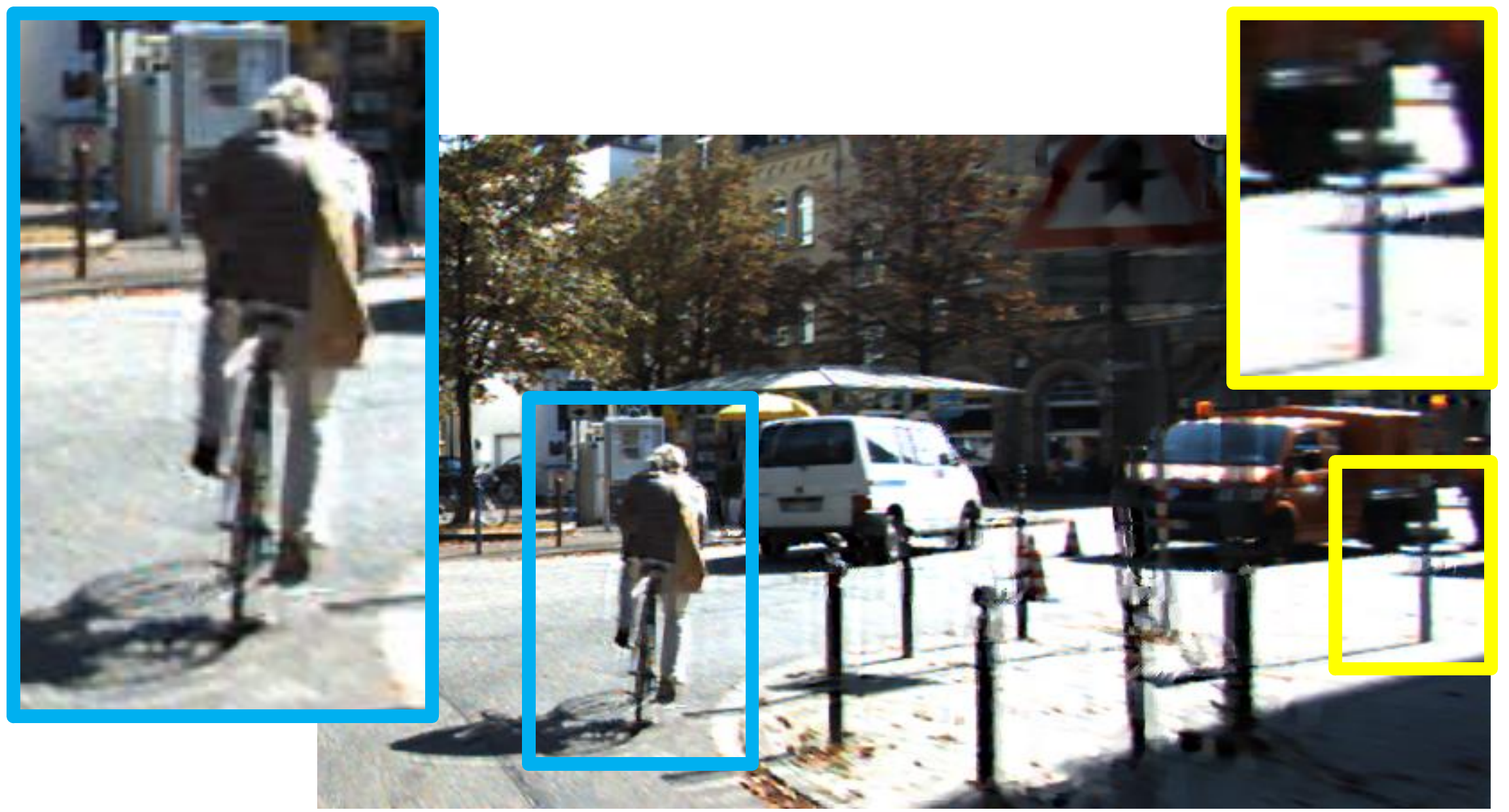}
&\hspace{0.0cm}
\includegraphics[width=0.200\textwidth]{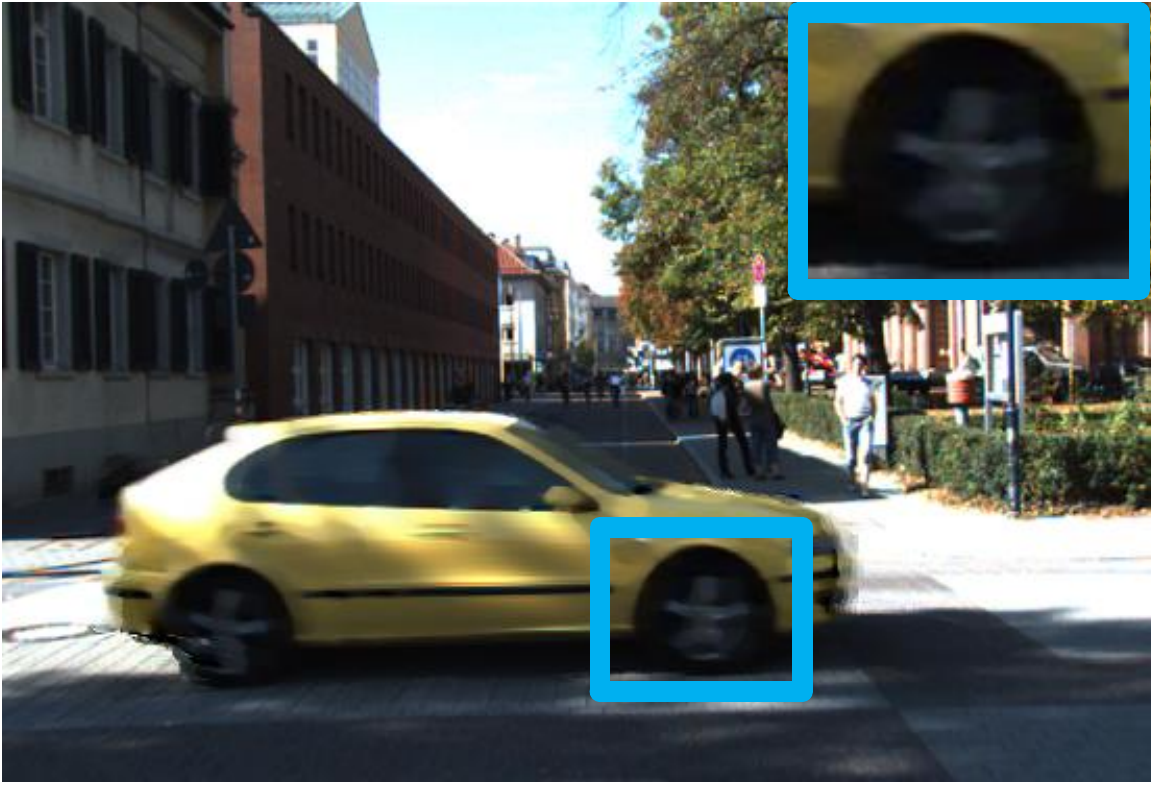}\\
\multicolumn{2}{c}{(a) Ours}  \\
\end{tabular}
\end{center}
\caption{Stereo deblurring results on outdoor scenarios. (a) Two samples from the KITTI autonomous driving benchmark dataset. (b) Deblurring result of \cite{hyun2015generalized}. (c) Deblurring result of  \cite{sellent2016stereo}. (d) Our deblurring result. Compared with both state-of-the-art monocular and stereo debluring methods, our method achieves the best performance especially for large motion regions in the scene. Best viewed in color on screen.
}
\label{fig:deblurresult}
\end{figure}

Image deblurring aims at recovering latent clean images from a single or multiple images, which is a fundamental task in image processing and computer vision. Image blur could be caused by various reasons, for example, optical aberration~\cite{schuler2012blind}, medium perturbation~\cite{kang2007automatic}, temperature variation~\cite{li2011fine}, defocus~\cite{shi2015just}, and motion~\cite{deng2012video, gupta2010single, jia2014mathematical, sun2015learning, Zheng_2013_ICCV}. The blur not only reduces the quality of the image causing loss of important details, but also hampers further analysis. Image deblurring has been extensively studied and various methods have been proposed.

In this work, we focus on image blur caused by motion. Motion blur is widely encountered in real world applications such as autonomous driving~\cite{franke2000real, geiger2012we}. Camera and object motion blur effects become more apparent when the exposure time of the camera increases due to low-light conditions. It is common to model the blur effect using kernels~\cite{jia2014mathematical, lee2013recent}. Under motion blur, the induced blur kernel would be in 2D \cite{hyun2015generalized} or 3D~\cite{sellent2016stereo}. For a scenario where both camera motion and multiple moving objects exist, the blur kernel is, in principle, defined for each pixel. Therefore, conventional blur removal methods, such as \cite{cho2009fast, gupta2010single, michaeli2014blind, xu2013unnatural} cannot be directly applied since they are restricted to a single or a fixed number of blur kernels, making them inferior in tackling general motion blur problems. 

On another front, stereo-based depth and motion estimation have witnessed significant progress over the last decade thanks to the availability of large benchmark datasets such as Middlebury~\cite{scharstein2002taxonomy} and KITTI~\cite{geiger2012we}. These benchmarks provide realistic scenarios with meaningful object classes and associated ground-truth annotations. The success of stereo-based motion estimation naturally prompts more advanced stereo based deblurring solutions, promising more accurate motion estimations to compensate for motion blurs. Very recently, Sellent \etal \cite{sellent2016stereo} proposed to exploit stereo information in aiding the challenging video deblurring task, where a piecewise rigid 3D scene flow representation is used to estimate motion blur kernels via local homographies. It makes a strong assumption that 3D scene flow can be reliably estimated, even under adverse conditions. While they reported favorable results on both synthetic and real data, all the experiments are confined to indoor scenarios.

\begin{figure}
\begin{center}
\begin{tabular}{cc}
\hspace{0.0cm}
\includegraphics[width=0.205\textwidth]{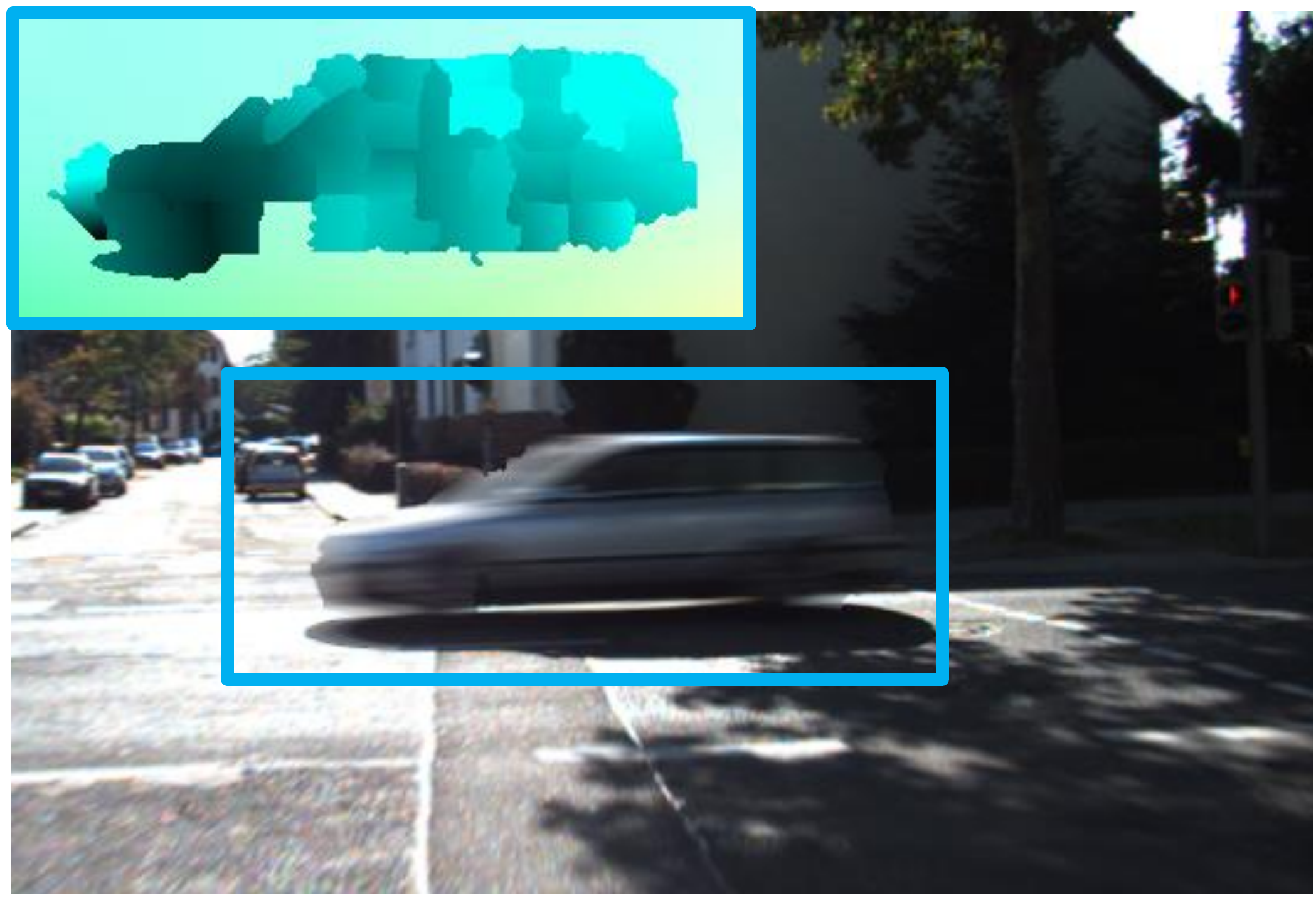}
&\hspace{0.0cm}
\includegraphics[width=0.205\textwidth]{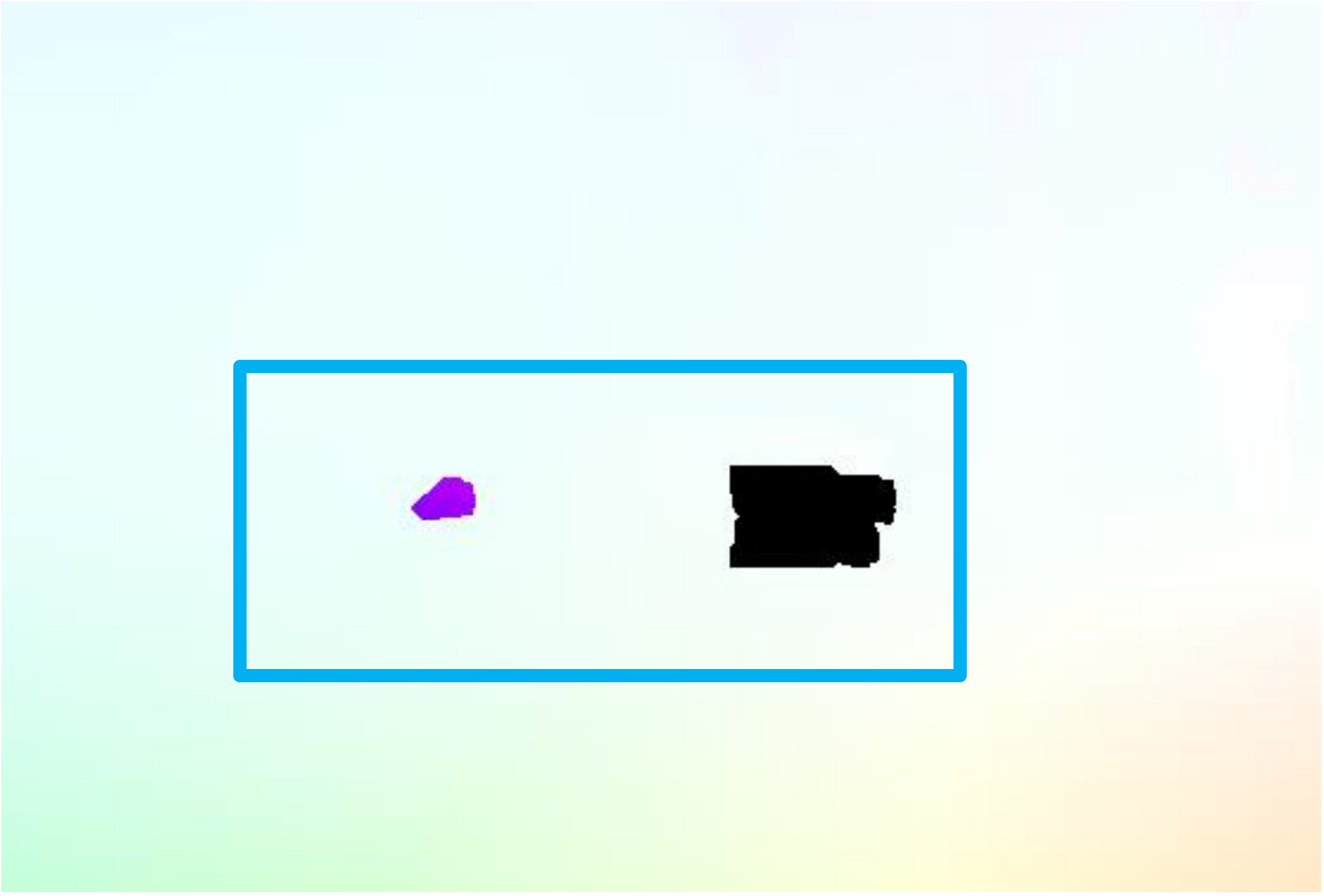}\\
(a) Blurry Image \& GT Flow & (b) Menze CVPR 2015 \cite{menze2015object} \\
\hspace{0.0cm}
\includegraphics[width=0.205\textwidth]{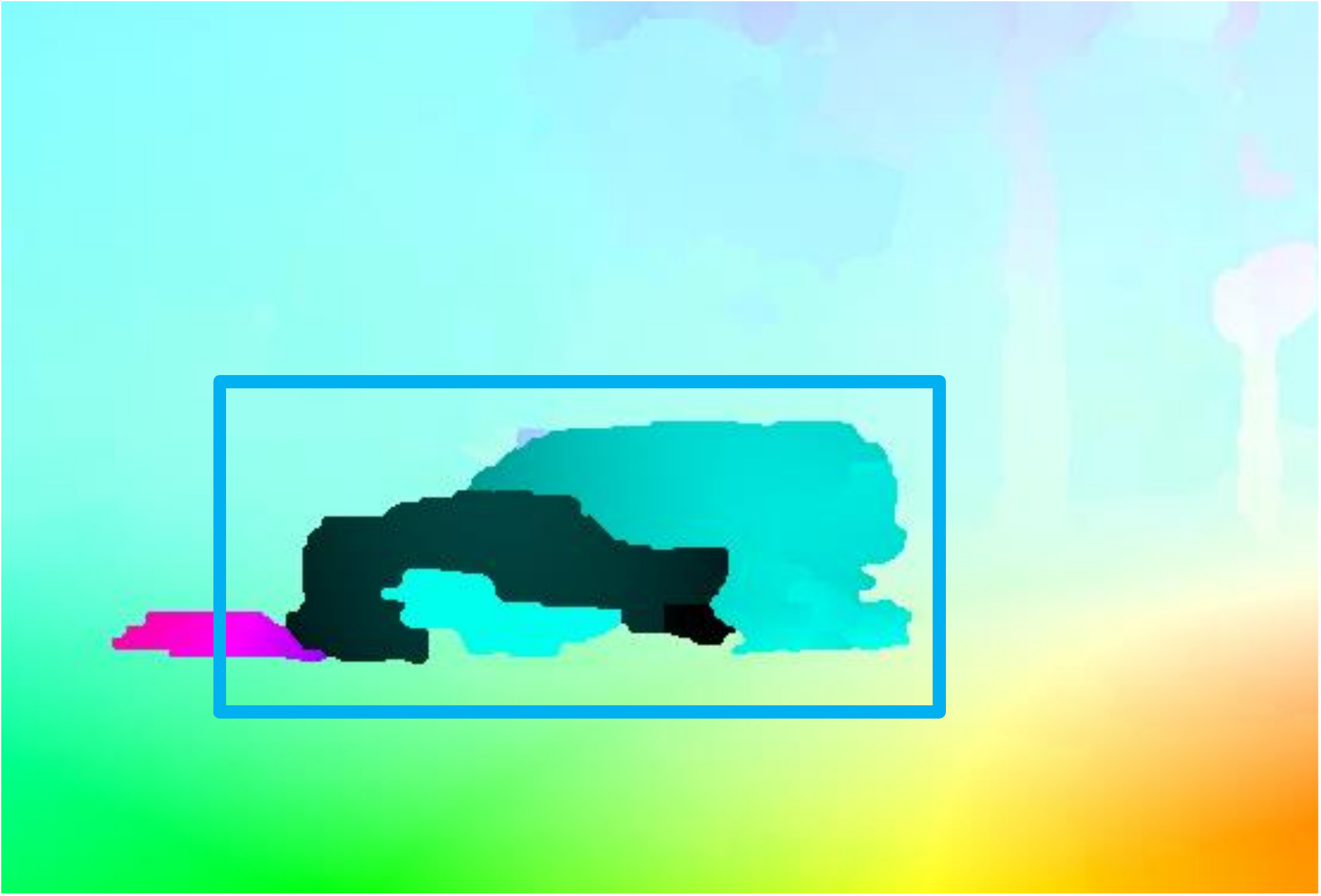}
&\hspace{0.0cm}
\includegraphics[width=0.205\textwidth]{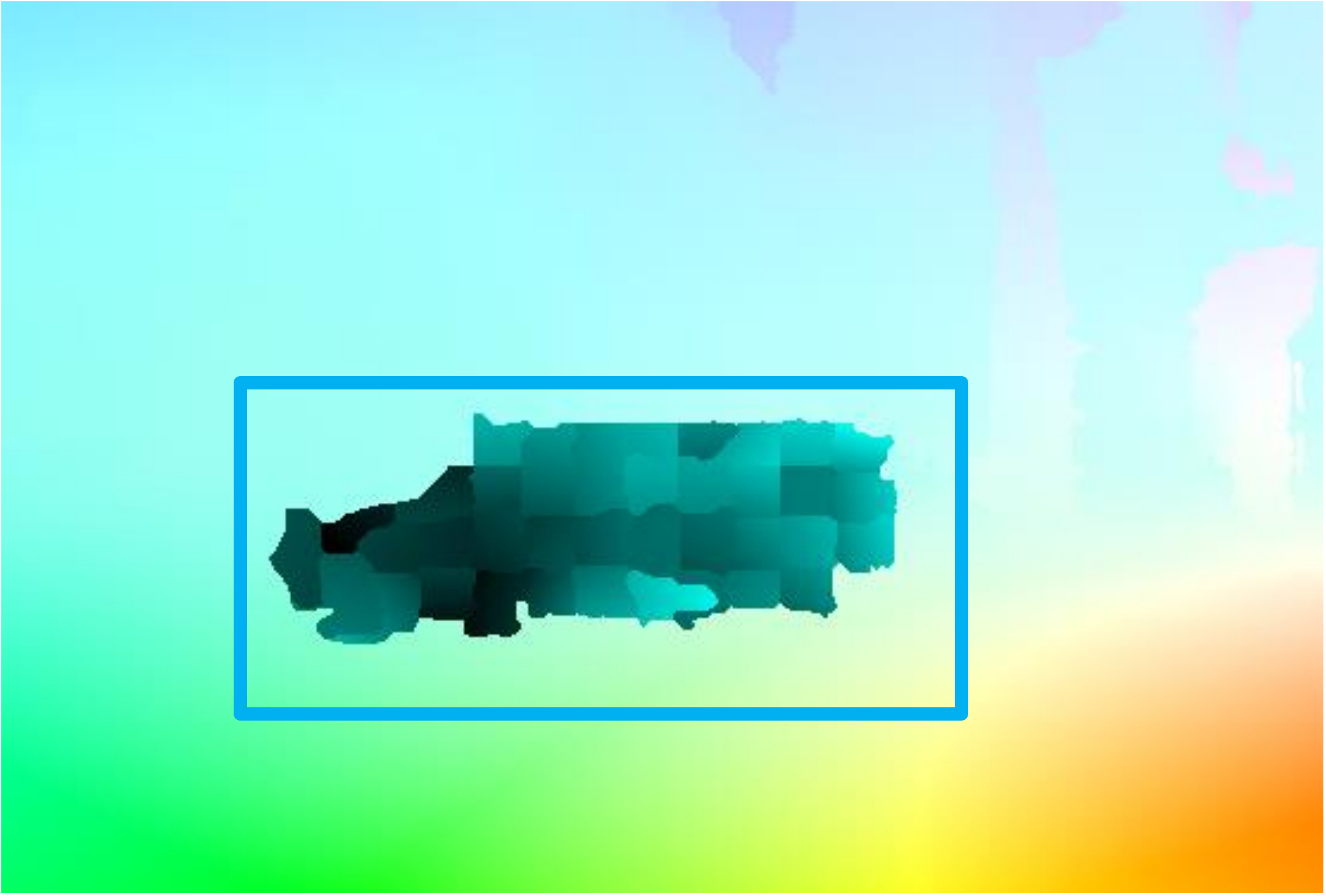}\\
(c) Sellent ECCV 2016 \cite{sellent2016stereo} & (d) Ours\\
\end{tabular}
\end{center}
\caption{
Scene flow estimation results for an outdoor scenarios. (a) A blur frame from the KITTI autonomous driving benchmark datasets. (b) Estimated flow by \cite{menze2015object}. (c) Estimated flow by \cite{sellent2016stereo}. (d) Our flow estimation result. Compared with both these state-of-the-art methods that rank the $1^{st}$ and $2^{nd}$ on the KITTI dataset, our method achieves the best performance especially for large motion region in the scene. Best viewed in color on screen.
}
\label{fig:flowcompare}
\end{figure}

The phenomenon around motion and blur can be viewed as a chicken-egg problem: More effective motion blur removal requires more accurate motion estimation. Yet, the accuracy of motion estimation highly depends on the quality of the images. We would like to argue that, scene flow estimation approaches that make use of color brightness constancy may be hindered by the blur images. In Fig.~\ref{fig:flowcompare}, we compare the scene flow estimation results of the state-of-the-art solutions on different blur images. It could be observed that the scene flow estimation performance deteriorates quickly \wrt the image blur.


Here, we aim to solve the above two problems simultaneously in a unified framework. Our motivation is that motion estimation and video deblur benefit from each other, \ie, better scene flow estimation will lead to a better deblurring result, and a cleaner image will lead to better flow estimation. We tackle a more general blur problem that is not only caused by camera motion but also by moving objects and depth variations in a dynamic scene. We define our problem as ``generalized stereo deblur'', where moving stereo cameras observe a dynamic scene with varying depths. We propose a new pipeline (see Fig.~\ref{fig:framework} for simultaneously estimating the 3D scene flow and deblurring images. Using our formulation, we attain significant improvement in numerous real challenging scenes as illustrated in Fig.~\ref{fig:deblurresult}.

The main contributions of our work are as follows:
\vspace{-2.5mm}
\begin{itemize}

\item We propose a novel joint optimization framework to simultaneously estimate the scene flow and deblurred latent images for dynamic scenes. Our deblurring objective benefits from the improved scene flow estimates and the estimated scene structure. Similarly, the scene flow objective allows deriving more accurate pixel-wise spatially varying blur kernels.


\vspace{-2.5mm}
\item Based on the piece-wise planar assumption, we obtain a structured blur kernel model. More specifically, the optical flows for pixels in the same superpixel are constrained by a single homography (see Section.\ref{sec:3.1}).

\vspace{-2.5mm}
\item As our experiments demonstrate, our method can successfully handle complex real-world scenes depicting fast moving objects, camera motions, uncontrolled lighting conditions, and shadows.
\end{itemize}

\section{Related Work}
Blur removal is an ill-posed problem, thus certain assumptions or additional constraints are required to regularize the solution space. Numerous methods have been proposed to address this problem \cite{hyun2015generalized, jia2014mathematical, sellent2016stereo, sun2015learning}, which can be categorized into two groups: monocular based approaches and binocular based approaches.


Monocular based approaches often assume that the captured scene is static and has a constant depth. Based on these assumptions, uniform or non-uniform blur kernels are estimated from a single image \cite{gupta2010single, hirsch2011fast, hu2014joint}. Hu \etal \cite{hu2014joint} proposed to jointly estimate the depth layering and remove non-uniform blur from a single blur image. While this unified framework is promising, user input for depth layers partition is required, and potential depth values should be known in advance. In practical settings, blur is spatially varying due to camera and object motion, which makes the kernel estimation a difficult problem.

Since blur parameters and the latent image are difficult to be estimated from a single image, the monocular based approaches are extended to video to remove blurs in dynamic scenes~\cite{seok2013dense, whyte2012non}. To this end, Deng \etal \cite{deng2012video} and He \etal \cite{he2010motion} apply feature tracking of a single moving object to obtain 2D displacement-based blur kernels for deblurring. Matsushita \etal \cite{matsushita2006full} and Cho \etal \cite{cho2012video} proposed to exploit the existence of salient sharp frames in videos. Nevertheless, the method of Matsushita \etal \cite{matsushita2006full} cannot remove blurs caused by moving objects. Moreover, the work of Cho \cite{cho2012video} cannot handle fast moving objects which have distinct motions from those of backgrounds. Wulff and Black \cite{wulff2014modeling} proposed a layered model to estimate the different motions of both foreground and background layers. However, these motions are restricted to affine models, and it is difficult to be extended to multi-layer scenes due to the requirement of depth ordering of the layers.


Kim and Lee \cite{kim2014segmentation} proposed a method based on a local linear motion without segmentation. This method incorporates optical flow estimation to guide the blur kernel estimation and is able to deal with certain object motion blur. In \cite{hyun2015generalized}, a new method is proposed to simultaneously estimate optical flow and tackle the case of general blur by minimization a single non-convex energy function. This method represents the state-of-the-art in video deblurring and is used for comparison in the experimental section.


As depth can significantly simplify the deblurring problem, the multi-view methods have been proposed to leverage on depth information. Building upon the work of Ezra and Nayar~\cite{nayar2004motion}, Li \etal~\cite{li2008hybrid} extended the hybrid camera with an additional low-resolution video camera where two low-resolution cameras form a stereo pair and provide a low-resolution depth map. Tai \etal~\cite{tai2008image} used a hybrid camera system to compute a pixel-wise kernel with optical flow. Xu \etal~\cite{xu2012depth} inferred depth from two blur images captured by a stereo camera and proposed a hierarchical estimation framework to remove motion blur caused by in-plane translation. Just recently, Sellent \etal~\cite{sellent2016stereo} proposed a video deblurring technique based on stereo video, where 3D scene flow is estimated from blur images using a piecewise rigid 3D scene flow representation.


\section{Formulation}\label{sec:model}
Our goal is to handle the blurs in stereo videos caused by the motion of the camera, objects, and large depth variations in a scene. To this end, we formulate our problem as a joint estimation of scene flow and image deblurring for dynamic scenes. In particular, we rely on the assumptions that the scene can be approximated by a set of 3D planes~\cite{yamaguchi2013robust} belonging to a finite number of objects\footnote{The background can be regarded as a single 'object' due to the camera motion.} performing rigid motions~\cite{menze2015object}. Based on these assumptions, we define our structured blur kernel as well as the energy functions for deblurring in the following sections.

\subsection{Blur Image Formation based on the Structured Pixel-wise Blur Kernel}
\label{sec:3.1}

\begin{figure}
\begin{center}
\begin{tabular}{cc}
\hspace{0.0cm}
\includegraphics[width=0.225\textwidth]{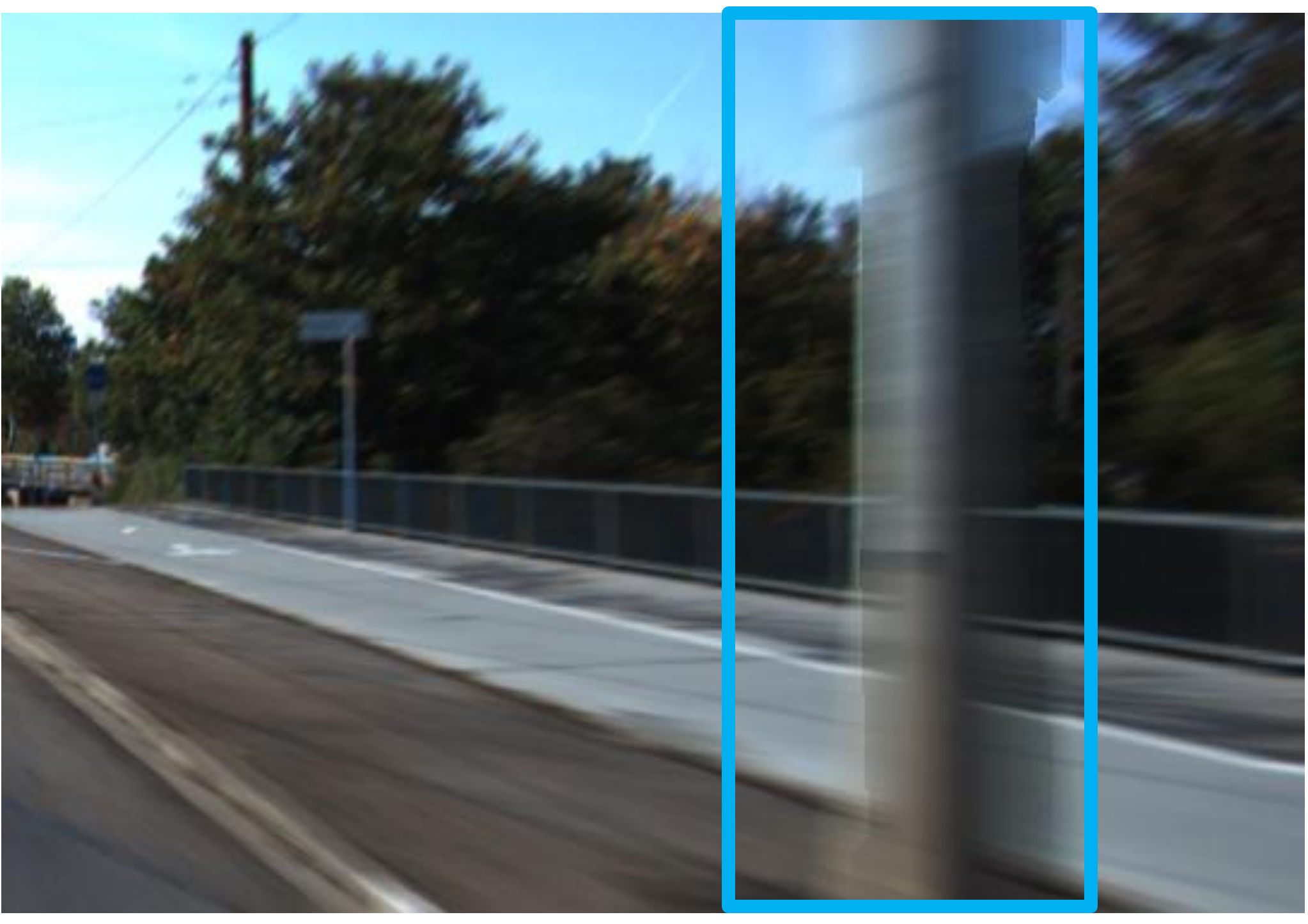}
&\hspace{0.0cm}
\includegraphics[width=0.225\textwidth]{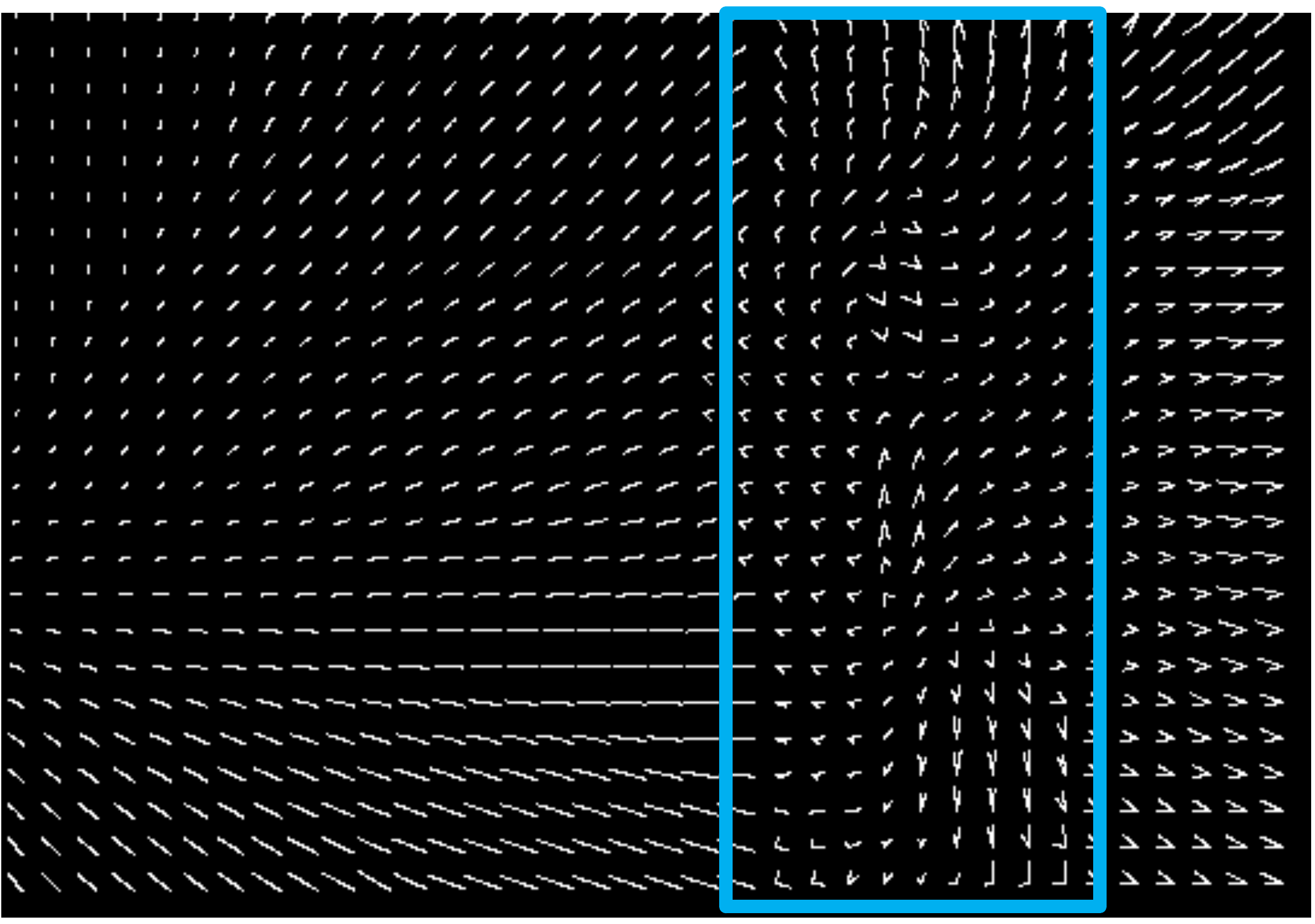}\\
(a) Original Blurry image & (b) Kim and Lee \cite{hyun2015generalized} \\
\hspace{0.0cm}
\includegraphics[width=0.225\textwidth]{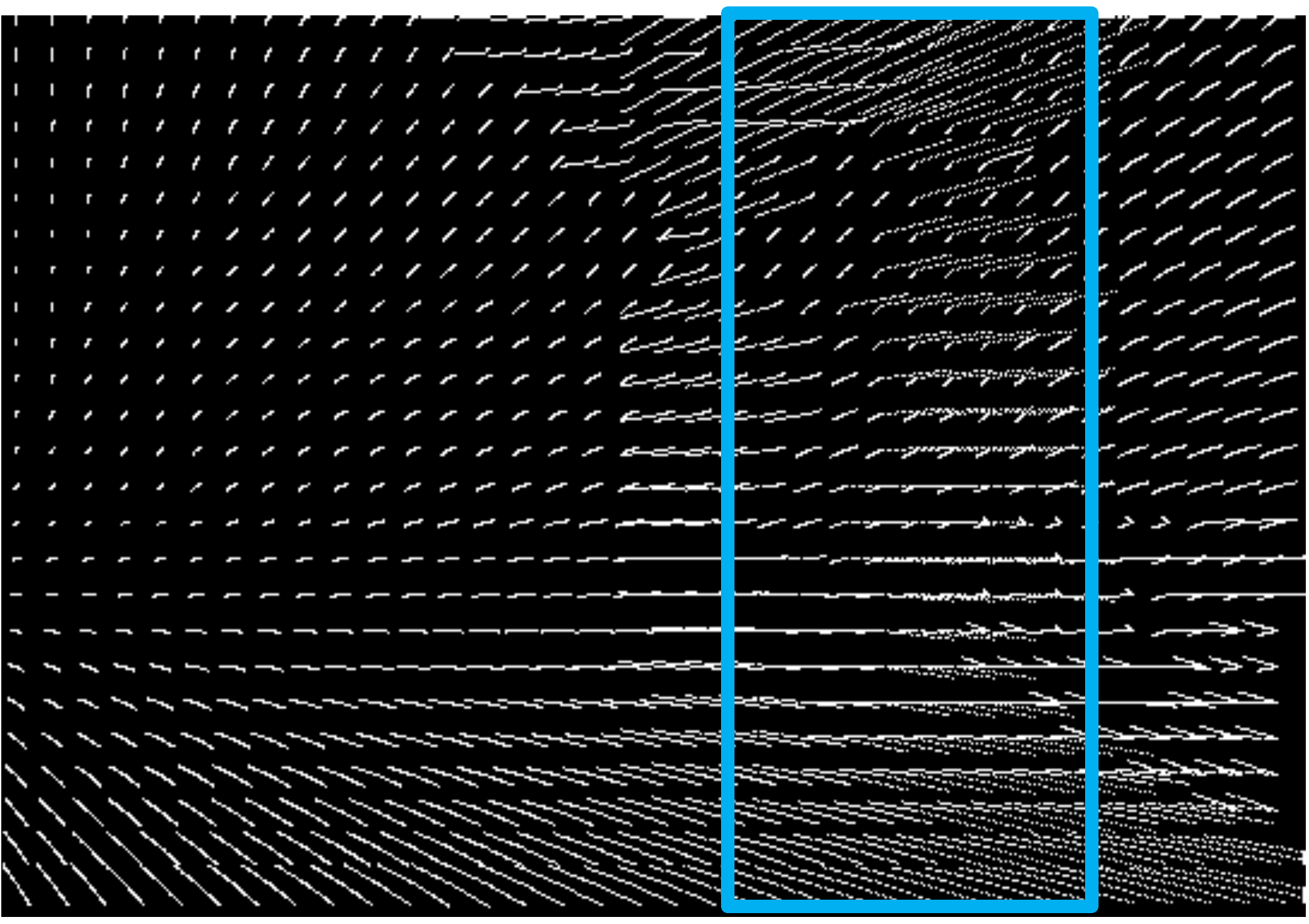}
&\hspace{0.0cm}
\includegraphics[width=0.225\textwidth]{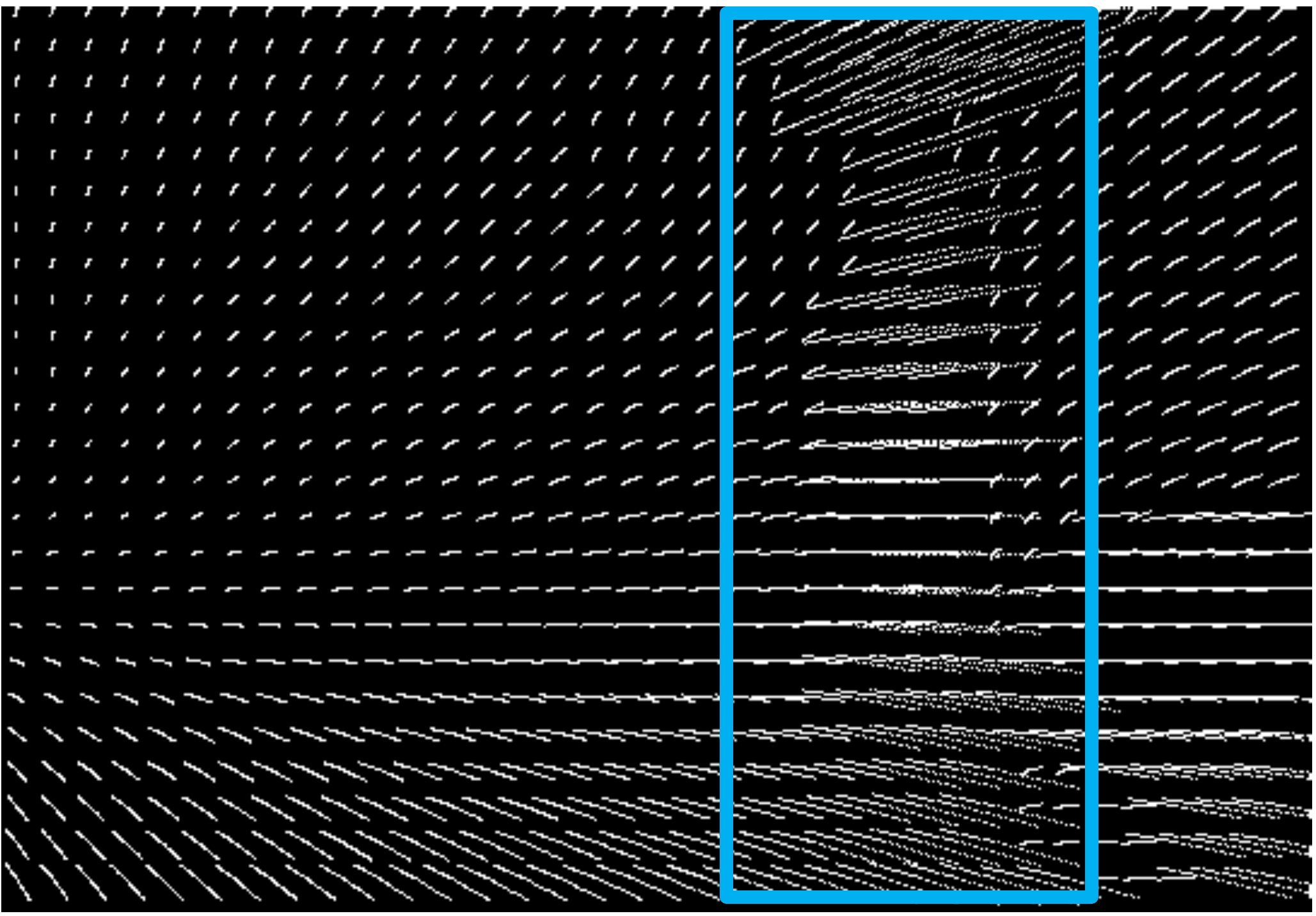}\\
(c) Sellent \etal \cite{sellent2016stereo} & (d) Ours\\
\end{tabular}
\end{center}
\caption{Blur kernel estimation on an outdoor scenario. (a) A blur frame from the KITTI autonomous driving benchmark datasets. (b) Blur kernel of \cite{hyun2015generalized}. (c) Blur kernel of \cite{sellent2016stereo}. (d) Our blur kernel. Compared with these monocular and stereo deblurring methods, our method achieves more accurate blur kernels.}
\label{fig:flowkernel}
\end{figure}

Blur images are formed by the integration of light intensity emitted from the dynamic scene over the aperture time interval of the camera. This defines the image frame in the video sequence as
{\small
\begin{equation} \label{eq:blurmodel}
\mathbf{B}_m({\bf x}) =\frac{1}{\tau} \int^{m+\frac{\tau}{2}}_{m-\frac{\tau}{2}} \mathbf{L}(m,{\bf x})\text{d}m=\frac{1}{\tau} \int^{m+\frac{\tau}{2}}_{m-\frac{\tau}{2}} \mathbf{L}_m({\bf x}+{\bf u}_m)\text{d}m
\end{equation}
}
where ${\bf B}_m$ is the blur frame, ${\bf L}\in [-T,T]\times \Omega$ is a continuous latent video sequence over a time interval $[-T,T]$, $\tau$ is the duty cycle, ${\bf u}_m$ is the optical flow at $m$. We denote ${\bf L}_m({\bf x})={\bf L}(m,{\bf x})$. This leads to the discretized version of blur model in Eq.~(\ref{eq:blurmodel}) as
\begin{equation}\label{eq:convBlurKernel}
{\bf B}_m({\bf x}) = {\bf A}_m^{\bf x}{\bf L}_m,
\end{equation}
where ${\bf A}_m^{\bf x}$ is the blur kernel vector for the image at location ${\bf x}$. We obtain the blur kernel matrix ${\bf A}$ by stacking ${\bf A}^{\bf x}$. This leads to the blur model for the image as $\mathbf{B}_m = {\bf A}_m{\bf L}_m$.~In order to handle multiple types of blurs, Kim \etal~\cite{kim2014segmentation} approximated the pixel-wise blur kernel using bidirectional optical flows
{\small
\begin{equation}
\begin{aligned} \label{eq:kimblurKernel}
\hspace{-0.1cm}& k_{m,{\bf x}}(u,v)=\\
&\left\{
\begin{gathered}
\begin{aligned}
&\frac{\delta(u{\tilde{ v}_{m+}}-v\tilde{u}_{m+})}{\tau_m||{\tilde{\mathbf{u}}_{m+}}||},&\;if\;u\in [0,\frac{\tau_m}{2} \tilde{u}_{m+}], v\in [0,\frac{\tau_m}{2}{\tilde v}_{m+}]\\
&\frac{\delta(u\tilde{v}_{m-}-v\tilde{u}_{m-})}{\tau_m||{\tilde{\mathbf{u}}_{m-}}||},&\;\;if\;\; u\in (0,\frac{\tau_m}{2} \tilde{u}_{m-}],v\in (0,\frac{\tau_m}{2}\tilde{v}_{m-}]\\
& 0, & otherwise
\end{aligned}
\end{gathered}\right.
\end{aligned}
\end{equation}
}
where $k_{t,{\bf x}}$ is the blur kernel at ${\bf x}$,~$\delta$ denotes the Kronecker delta,~$\tilde{{\bf u}}_{m+} = (\tilde{u}_{m+},\tilde{v}_{m+})$ and $\tilde{{\bf u}}_{m-} = (\tilde{u}_{m-},\tilde{v}_{m-})$ are the bidirectional optical flows at frame $m$. In particular, ${\bf u}_{m+}={\bf u}_{m\rightarrow m+1}$ and ${\bf u}_{m-}={\bf u}_{m\rightarrow m-1}$. They  jointly estimated the optical flow and the deblurred images. In our setup, the stereo video provides the depth information for each frame. Based on our piece-wise planar assumptions on the scene, optical flows for pixels lying on the same plane are constrained by a single homography. In particular, we represent the scene in terms of superpixels and finite number of objects with rigid motions. We denote $\mathcal{S}$ and $\mathcal{O}$ as the set of superpixels and moving objects, respectively. Each superpixel $i\in S$ is associated with a region $\mathcal{R}_i$ in the image with a plane variable $\mathbf{n}_{i,k} \in \mathbb{R}^3$ in 3D ($\mathbf{n}_{i,k}^T\mathbf{X}=1$ for $\mathbf{X} \in \mathbb{R}^3$), where $k \in \left\{1,\cdots,|\mathcal{O}|\right\}$ denotes that superpixel $i$ is associated with object $k$ inheriting its corresponding motion parameters $\mathbf{o}_k=(\mathbf{R}_k, \mathbf{t}_k) \in \mathbb{SE}(3)$, where  $\mathbf{R}_k \in \mathbb{R}^{3 \times 3}$ is the rotation matrix and $\mathbf{t}_k \in \mathbb{R}^{3}$ is the translation vector. Note that $({\mathbf o}_k,{\mathbf n}_{i,k})$ encodes the scene flow information~\cite{menze2015object}. Given the parameters $({\mathbf o}_k,{\mathbf n}_{i,k})$, we can obtain the homography defined for superpixel $i$ as
\vspace{-1mm}
\begin{equation}
\mathbf{H}_i = {\bf K}(\mathbf{R}_k -\mathbf{t}_k \mathbf{n}^{T}_{i,k}){\bf K}^{-1}
\end{equation}
\vspace{-1mm}
where ${\mathbf K} \in \mathbb{R}^{3 \times 3}$ is the intrinsic matrix. The optical flow is then defined as
\vspace{-1mm}
\begin{equation}\label{eq:structuredflow}
\begin{aligned}
\mathbf{u}_{i,j} = \mathbf{H}_i\mathbf{x}_{i,j}-\mathbf{x}_{i,j}
\end{aligned}
\end{equation}
where ${\mathbf x}_{i,j}$ is the coordinate of pixel $j$ in superpixel $i$. This shows that the optical flows for pixels in a superpixel are constrained by the homography. Thus, it leads to a structured version of blur kernel defined in Eq.~(\ref{eq:kimblurKernel}). In Fig.~\ref{fig:flowkernel}, we compare our blur kernel estimation with the Kim and Lee \cite{hyun2015generalized} and Sellent \etal \cite{sellent2016stereo}. Our kernels are more structural, which also leads to more accurate scene flow estimation.

\subsection{Energy Minimization}
We formulate the problem in a single framework as a discrete-continuous optimization problem to jointly estimate the scene flow and deblur the images. In particular, our energy minimization model is formulated as
\begin{eqnarray}\label{eq:energy}
\mathbf{E}=\underbrace{\sum_{i\in \mathcal{S}} \mathbf{\phi}_{i}(\mathbf{n}_{i},\mathbf{o},\mathbf{L})}_{\text{data~term}} + \underbrace{\sum_{i,j}\mathbf{\phi}_{i,j}(\mathbf{n}_{i},\mathbf{n}_{j},\mathbf{o})}_{\substack{\text{scene flow}\\\text{smoothness term}}}+\underbrace{\sum_m\mathbf{\phi}_{m}(\mathbf{L})}_{\substack{\text{latent image}\\\text{regularisation}}}
\end{eqnarray}
which consists of a data term, a smoothness term for scene flow, and a spatial regularization term for latent clean images. Our model is initially defined on three consecutive pairs of stereo video sequences. It can also allow the input with two pairs of frames. Details are provided in Section~\ref{sec:experiments}. The energy terms are discussed in the following sections.


In Section~\ref{sec:optimization}, we solve the optimization problem in an alternative manner to handle mixed discrete and continuous variables, thus allowing us to jointly estimate the scene flow and deblur the images.
\begin{figure*}
\begin{center}
\includegraphics[width=.985\textwidth]{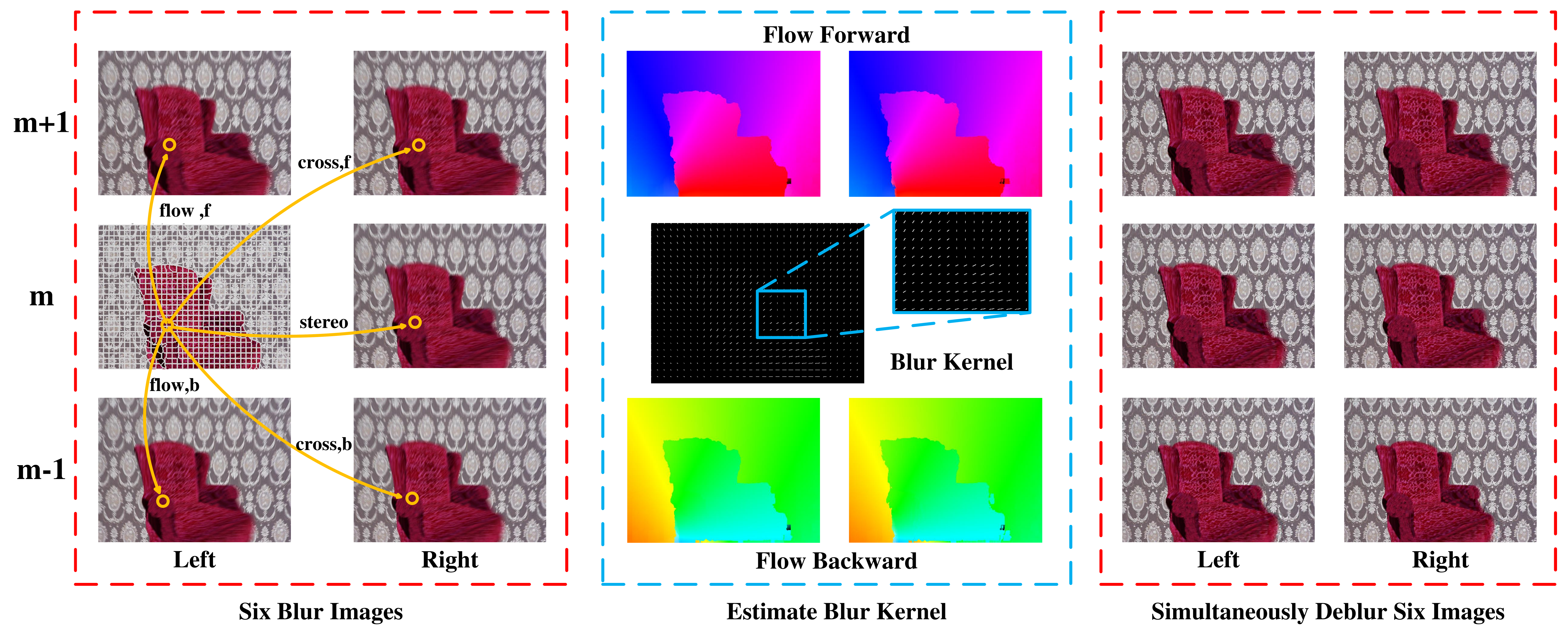}
\end{center}
\caption{
Illustration of our 'generalized stereo deblurring' method. We simultaneously compute four scene flows (in two directions and in two view), and deblur six images. In case the input contains only two images, we use the reflection of the flow forward as the flow backward in the deblurring part.}
\label{fig:framework}
\end{figure*}

\subsection{Data Term}
\label{sec:dataTerm}
Our data term involves mixed discrete and continuous variables, and are of three different kinds. The first kind encodes the fact that the corresponding pixels across the six latent images should have similar appearance (brightness constancy). This lets us write the term as
\begin{equation}
\phi_i^1(\mathbf{n}_{i,k},\mathbf{o}_{k},\mathbf{L}) = \theta_{1}|\mathbf L(\mathbf x)-\mathbf L^*(\mathbf H^*\mathbf x)|_1,
\end{equation}
where the superscript $* \in \left\{ \mathbf{stereo},\mathbf{flow}_{f,b},\mathbf{cross}_{f,b} \right\}$ denotes the warping direction to other images and $(\cdot)_{f,b}$ denotes the forward and backward direction, respectively (see Fig.~\ref{fig:framework}). We adopt the robust $\ell_1$ 
norm to enforce its robustness against noise and occlusions.

Our second potential, similar to one term used in~\cite{menze2015object}, is defined as
\[
\phi_i^2(\mathbf{n}_{i,k},\mathbf{o}_{k})=\left \{
  \begin{tabular}{cc}
  $\theta_2$$\rho_{\alpha_1}(||{\bf H}^{*}{\bf x}-\mathbf{x}^{*}||_2)$& if $\bf x\in \Pi_{\bf x}$, \\
  0 & otherwise. \nonumber \\
  \end{tabular}
\right.
\]
where $\rho_{\alpha}(\cdot)=\min(|\cdot|,\alpha)$ denotes the truncated $\ell_1$ penalty function. More specifically, it encodes the information that the warping of feature points $\bf x\in \Pi_{\bf x}$ based on ${\bf H}^{*}$ should match its extracted correspondences in  the target view. In particular, $\Pi_{\bf x}$ is obtained in a similar manner as \cite{menze2015object}.

The third data term, making use of the observed blur images, is defined as
\vspace{-1.5mm}
\begin{equation}
\mathbf{\phi}_i^3(\mathbf{n}_{i,k},\mathbf{o}_{k},\mathbf{L})= \theta_{3}\sum_m\sum_{\partial_* }\left\|{\partial_*}\mathbf{A}_{m}(\mathbf{n}_{i,k},\mathbf{o}_{k})\mathbf{L}_m - {\partial_*}\mathbf{B}_m\right\|_2^2 \nonumber
\end{equation}
\vspace{-0.5mm}
where $\partial_*$ are the Toeplitz matrices corresponding to the  horizontal and vertical derivative filters. This term encourages the intensity changes in the estimated blur images to be close to that of the observed blur images.


\subsection{Smoothness Term for Scene Flow}
\label{sec:smoothTerm}
Our energy model exploits a smoothness potential that involves the discrete and continuous variables. It is similar to the ones used in~\cite{menze2015object}. In particular, our smoothness term includes three different types. The first one is to encode the compatibility of two superpixels that share a common boundary by respecting the depth discontinuities. To this end, we define our potential function as
\begin{equation}
\phi_{i,j}^1(\mathbf{n}_{i},\mathbf{n}_{j})= \theta_{4}\sum_{\mathbf{x}\in \mathcal{B}_{i,j}}\rho_{\alpha_2}(\omega_{i,j}(\mathbf{n}_i,\mathbf{n}_j,\mathbf{x}))
\end{equation}
where $\omega(\mathbf{n}_i,\mathbf{x})$ is the disparity of pixel $\mathbf{x}$ in superpixel $i$ in the reference disparity map, $\omega_{i,j}(\mathbf{n}_i,\mathbf{n}_j,\mathbf{x}) = \omega(\mathbf{n}_i,\mathbf{x})-\omega(\mathbf{n}_j,\mathbf{x})$ are the distance of disparity for pixel ${\bf x} \in \mathcal{B}_{i,j}$ on the boundary.

The second potential is to encourage the neighbor superpixels to orient in the same direction. It is expressed as
\begin{equation}
\phi_{i,j}^2(\mathbf{n}_{i},\mathbf{n}_{j}) = \theta_{5}\rho_{\alpha_3} \left(1-\frac{|\mathbf{n}_{i}^T \mathbf{n}_{j}|}{\left\| \mathbf{n}_{i} \right\| \left\| \mathbf{n}_{j} \right\|}\right).
\end{equation}

The third potential is to encode the fact that the motion boundaries are co-aligned with disparity discontinuities. This potential can be expressed as
{\small
\begin{eqnarray}
\hspace{-0.8cm}&&\phi_{i,j}^3(\mathbf{n}_{i,k},\mathbf{n}_{j,k'}) \nonumber\\
\hspace{-0.8cm}&&=\theta_{6}\left\{\begin{tabular}{cc}
$\hspace{-0.1cm}\exp {\Big(}-\frac{\lambda}{|\mathcal{B}_{i,j}|}\sum\limits_{\mathbf{x} \in \mathcal{B}_{i,j}} \omega_{i,j}(\mathbf{n}_i,\mathbf{n}_j,\mathbf{x})^2 \frac{|\mathbf{n}_i^T \mathbf{n}_j|}{\left\| \mathbf{n}_i \right\| \left\| \mathbf{n}_j \right\|}{\Big)}$& if  $k \neq k'$,\nonumber\\
0 & else.
\end{tabular}
\right.
\end{eqnarray}
}
where $|\mathcal{B}_{i,j}|$ denotes the number of pixels shared along boundary between superpixels $i$ and $j$.

\subsection{Regularization Term for Latent Images}
\label{sec:regularization}
Spatial regularization has proven its importance in image deblurring \cite{krishnan2009fast, krishnan2011blind}. In our model, we use the total variation term to suppress the noise in the latent image while preserving edges, and penalize spatial fluctuations. Therefore, our potential takes the form
\begin{equation} \label{Eregularization1}
\phi_m = |\nabla \mathbf{L}_m|.
\end{equation}
Note that the total variation is applied to each color channel.

\section{Solution}\label{sec:optimization}
The optimization of our energy function defined in Eq.~(\ref{eq:energy}), involving both discrete and continuous variables, is challenging to solve. Recall that our model involves two set of variables, namely scene flow variables and latent images. Fortunately, given one set of variables, we can solve the other efficiently. Therefore, we perform the optimization iteratively by the following steps,
\begin{itemize}
\vspace{-2.0mm}
\item Fix latent image ${\bf L}$, solve scene flow by optimizing Eq.~(\ref{eq:sceneFlowEnergy}) (See Section~\ref{sec:sceneflow}).
\vspace{-2.0mm}
\item Fix scene flow parameters, ${\bf n}$ and ${\bf o}$, solve latent image by optimizing Eq.~(\ref{eq:latentImageEnergy}) (See Section~\ref{sec:deblurring}).
\end{itemize}
\vspace{-2.0mm}
In the following sections, we describe the details for each optimization step.

\subsection{Scene flow estimation}\label{sec:sceneflow}
We fix latent images, namely $\mathbf{L} = \tilde{\mathbf{L}}$, Eq.~(\ref{eq:energy}) reduces to
\begin{equation}\label{eq:sceneFlowEnergy}
\min_{{\bf n},{\bf o}}\sum_{i\in \mathcal{S}}\mathbf{\phi}_{i}^{1,2,3}(\mathbf{n}_{i},\mathbf{o},\tilde{\mathbf{L}}) + \sum_{i,j}\mathbf{\phi}_{i,j}^{1,2,3}(\mathbf{n}_{i},\mathbf{n}_{j},\mathbf{o}).
\end{equation}
which becomes a discrete-continuous CRF optimization problem. We use the sequential tree-reweighted message passing (TRW-S) method in \cite{menze2015object} to find the solution.


\subsection{Deblurring}\label{sec:deblurring}
Given the scene flow parameters, namely $\tilde{\mathbf{n}}$, and $\tilde{\mathbf{o}}$, the blur kernel matrix, ${\bf A}_m$ is derived based on Eq.~(\ref{eq:kimblurKernel}), and Eq.~(\ref{eq:structuredflow}). The objective function in Eq.~(\ref{eq:energy}) becomes convex with respect to $\mathbf{L}$ and is expressed as
\begin{equation}\label{eq:latentImageEnergy}
\min_{{\bf L}}\sum_{i\in \mathcal{S}} \mathbf{\phi}_{i}^1(\tilde{\mathbf{n}}_{i},\tilde{\mathbf{o}},\mathbf{L}) +\mathbf{\phi}_{i}^3(\tilde{\mathbf{n}}_{i,k},\tilde{\mathbf{o}}_k,\mathbf{L})+\sum_m\mathbf{\phi}_{m}(\mathbf{L}).
\end{equation}

In order to obtain sharp image $\mathbf{L}$, we adopt the conventional convex optimization method \cite{chambolle2011first} and derive the primal-dual updating scheme as follows
\begin{equation}
\left\{
\begin{gathered}
\begin{aligned}
& \mathbf{p}_m^{r+1}=\frac{\mathbf{p}_{m}^{r}+\gamma \nabla \mathbf{L}_m^r}{\mathbf{max}(1,\mathbf{abs}(\mathbf{p}_{m}^{r}+\gamma\nabla \mathbf{L}_m^r))}\\
& \mathbf{q}_{m,*}^{r+1}=\frac{\mathbf{q}_{m,*}^{r}+\gamma\theta_1 (\mathbf{L}_m^r-\mathbf{L}_{m,*}^{r})}{\max(1,\mathbf{abs}(\mathbf{q}_{m,*}^{r}+\gamma\theta_1 (\mathbf{L}_m^r-\mathbf{L}_{m,*}^{r}))}\\
& \mathbf{L}_{m}^{r+1} =\arg \min_{\mathbf{L}_m} \sum_i \theta_{3}\sum_{\partial_* }\left\|{\partial_*}\mathbf{A}_{m}\mathbf{L}_m - {\partial_*}\mathbf{B}_m\right\|_2^2 +\\
& \frac{\left\|[\mathbf{L}_m-\eta ((\nabla {\bf p}_{m}^{r+1})^{T}+\theta_1({\bf q}_{m,+*}^{r+1}-{\bf q}_{m,-*}^{r+1})^{T})]-\mathbf{L}_m^r\right\|^2}{2\eta}
\end{aligned}
\end{gathered}\right.
\end{equation}
where  $\mathbf{p}_m$, $\mathbf{q}_{m,*}$ are the dual variables, $\gamma$ and $\eta$ are the step variants which can be modified at each iteration, and $r$ is the iteration number.

\section{Experiments}\label{sec:experiments}


To demonstrate the effectiveness of our method, we evaluate it on two datasets: the synthetic chair sequence~\cite{sellent2016stereo} and KITTI dataset~\cite{geiger2013vision}. We discuss our results on both datasets in the following sections.

\subsection{Experimental Setup}
\noindent{\bf Initialization.} Our model is formulated on three consecutive stereo pairs. In particular, we treat the middle frame in the left view as the reference image. We adopt the StereoSLIC~\cite{yamaguchi2013robust} to generate the superpixels. Given the stereo images, we apply the approach in~\cite{geiger2011stereoscan} to obtain sparse feature correspondences. The traditional SGM~\cite{hirschmuller2008stereo} method is applied to obtain a disparity map which is used to initialize the plane parameters. The motion hypotheses are generated using RANSAC as implemented in~\cite{geiger2011stereoscan}. In order to obtain the model parameters $\{\theta\}$ and $\{\alpha\}$, we performed block-coordinate-descent on a subset of 30 randomly selected training images.

\noindent{\bf Evaluations.} Since our method estimates the scene flow and deblurs the images, we evaluate these two tasks separately. For the scene flow estimation results, we evaluate both the optical flow and disparity map by the same error metric, which is by counting the number of pixels having errors more than $3$ pixels and $5\%$ of its ground-truth. We adopt the PSNR to evaluate the deblurred image sequences for left and right view separately. Thus, for each sequence, we report three values: disparity errors for three stereo image pairs, flow errors in forward and backward directions, and PSNR values for six images.

\noindent\textbf{Baseline Methods.}
As for our scene flow results, we compare with piece-wise rigid scene flow method (PRSF)~\cite{vogel20153d}, which ranks the first on KITTI stereo and optical flow benchmark. Note that PRSF is used as a preprocessing stage in \cite{sellent2016stereo}.~We then compare our deblurring results with the state-of-the-art deblurring approach for monocular video sequence~\cite{hyun2015generalized}, and the approach for stereo videos~\cite{sellent2016stereo}.


\begin{figure*}
\begin{center}
\begin{small}
\begin{tabular}{cccc}
\hspace{0.0cm}
\includegraphics[width=0.21\textwidth]{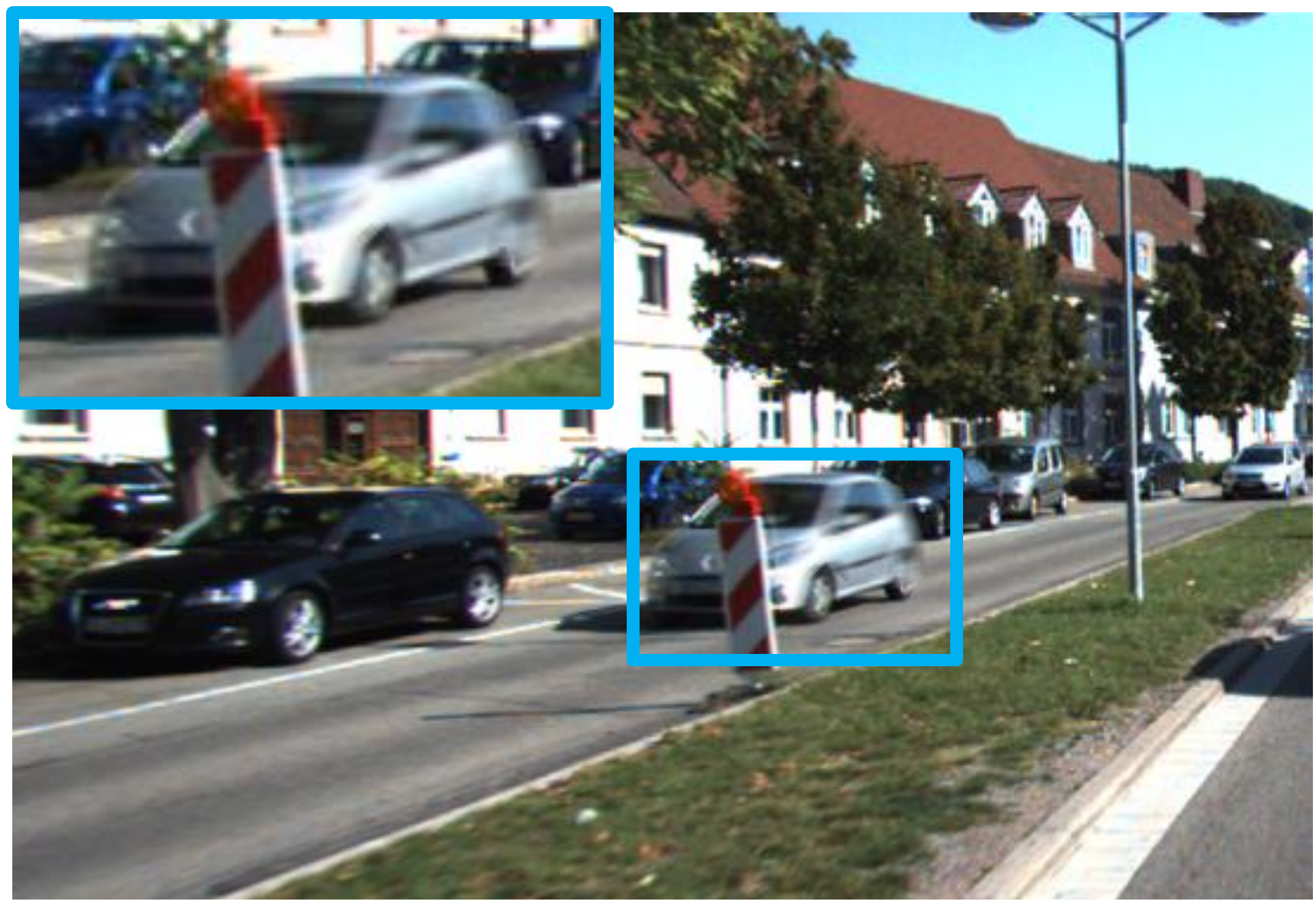}
&\hspace{0.0cm}
\includegraphics[width=0.21\textwidth]{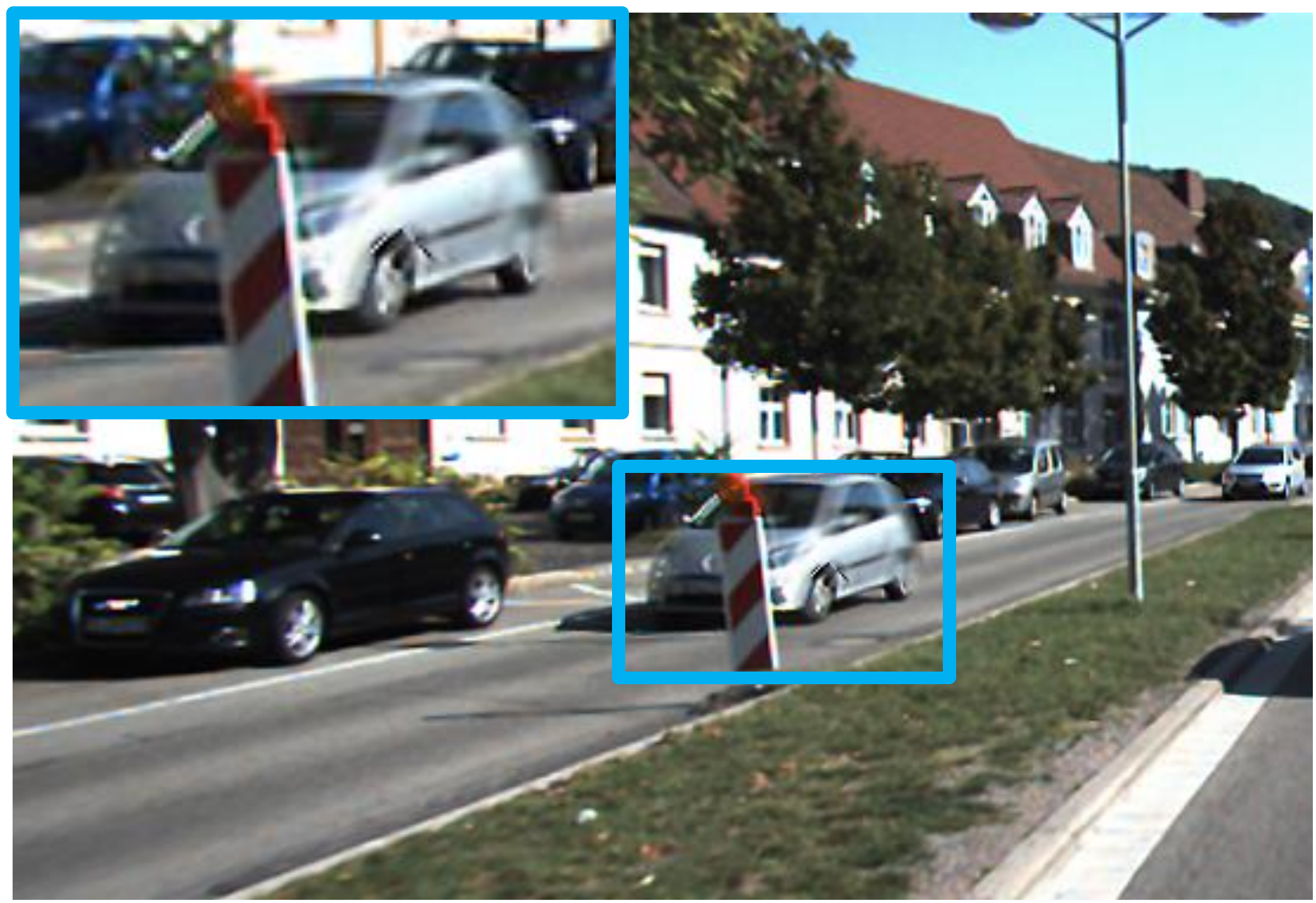}
&\hspace{0.0cm}
\includegraphics[width=0.21\textwidth]{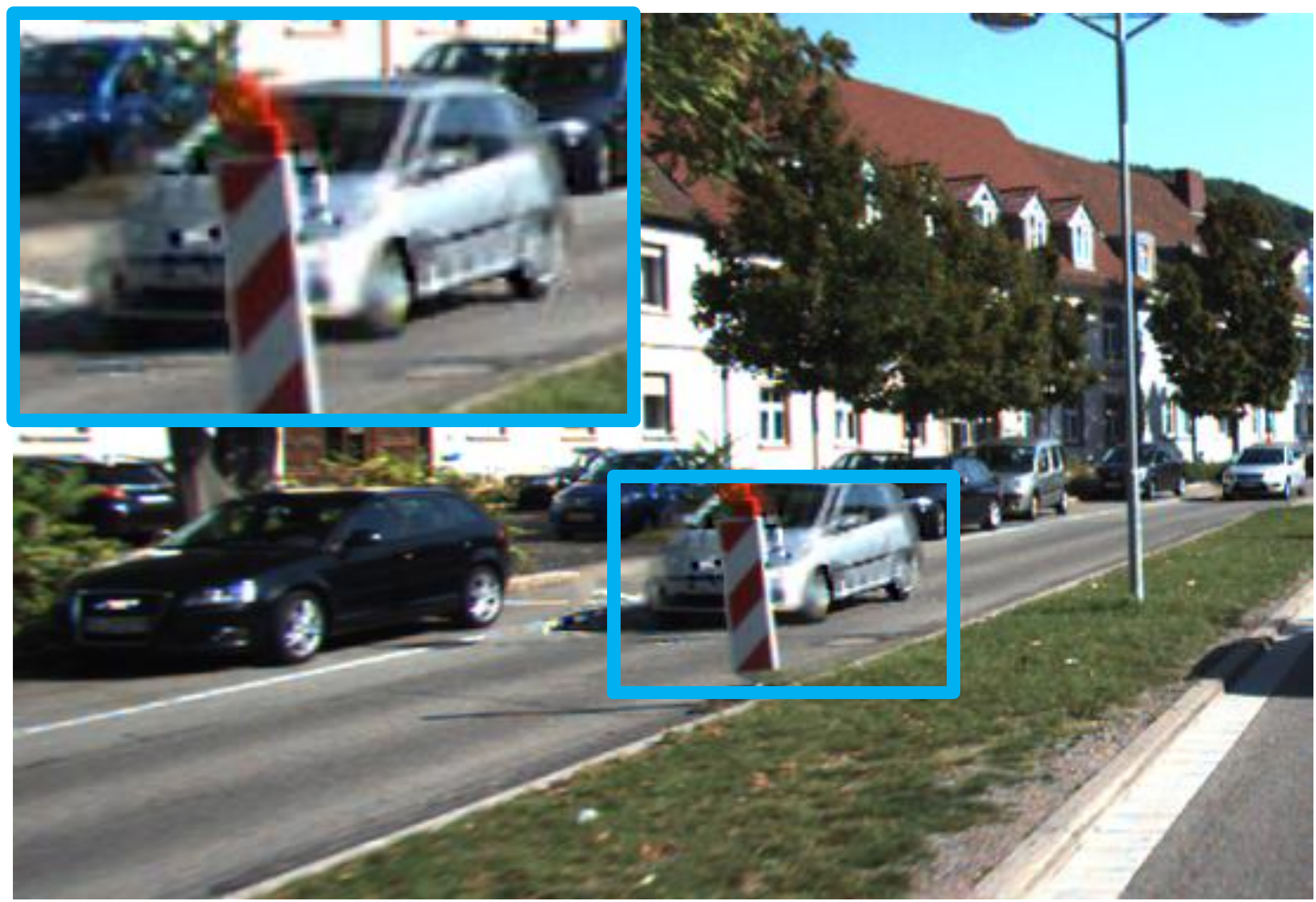}
&\hspace{0.0cm}
\includegraphics[width=0.21\textwidth]{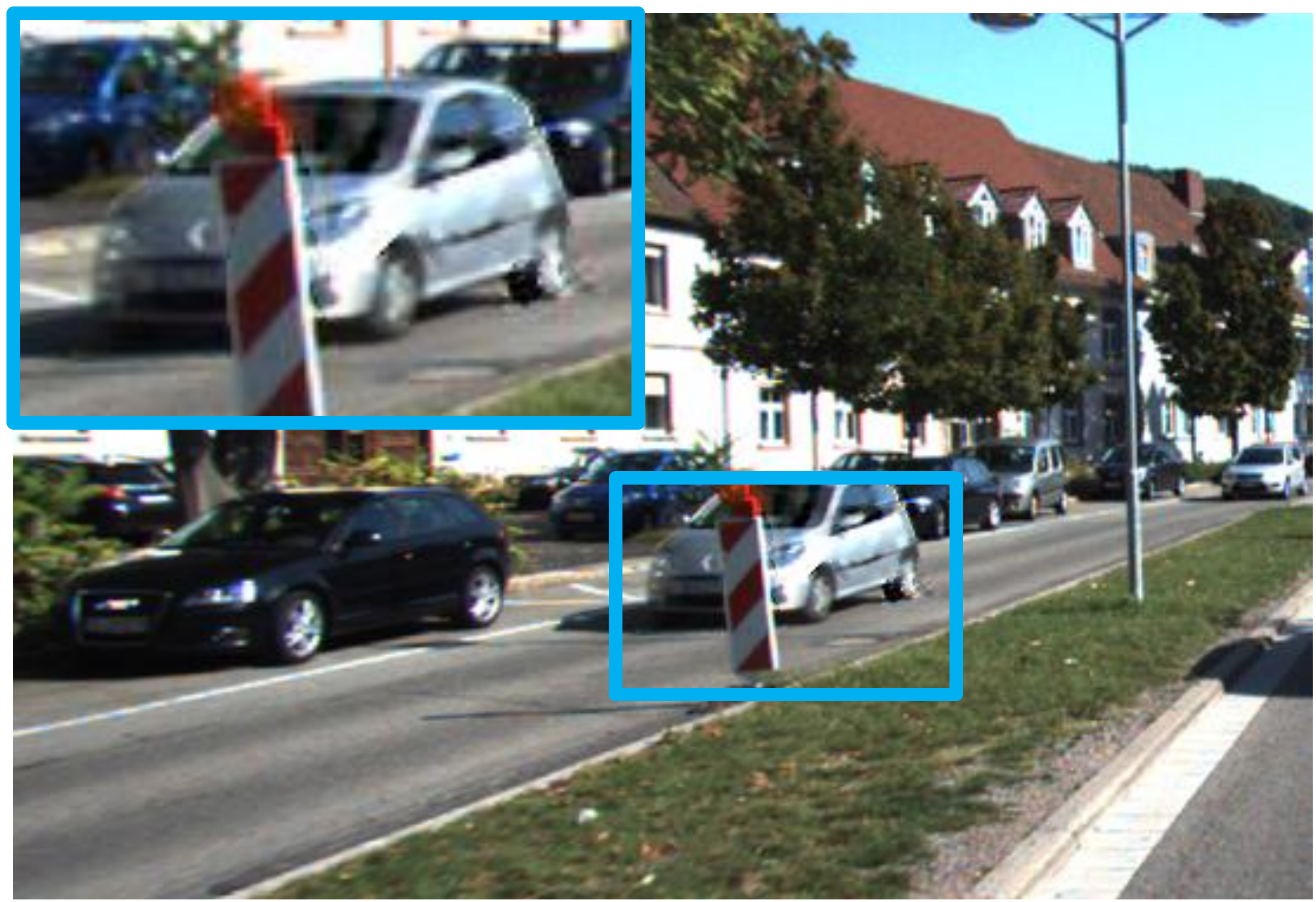}\\
\hspace{0.0cm}
\includegraphics[width=0.21\textwidth]{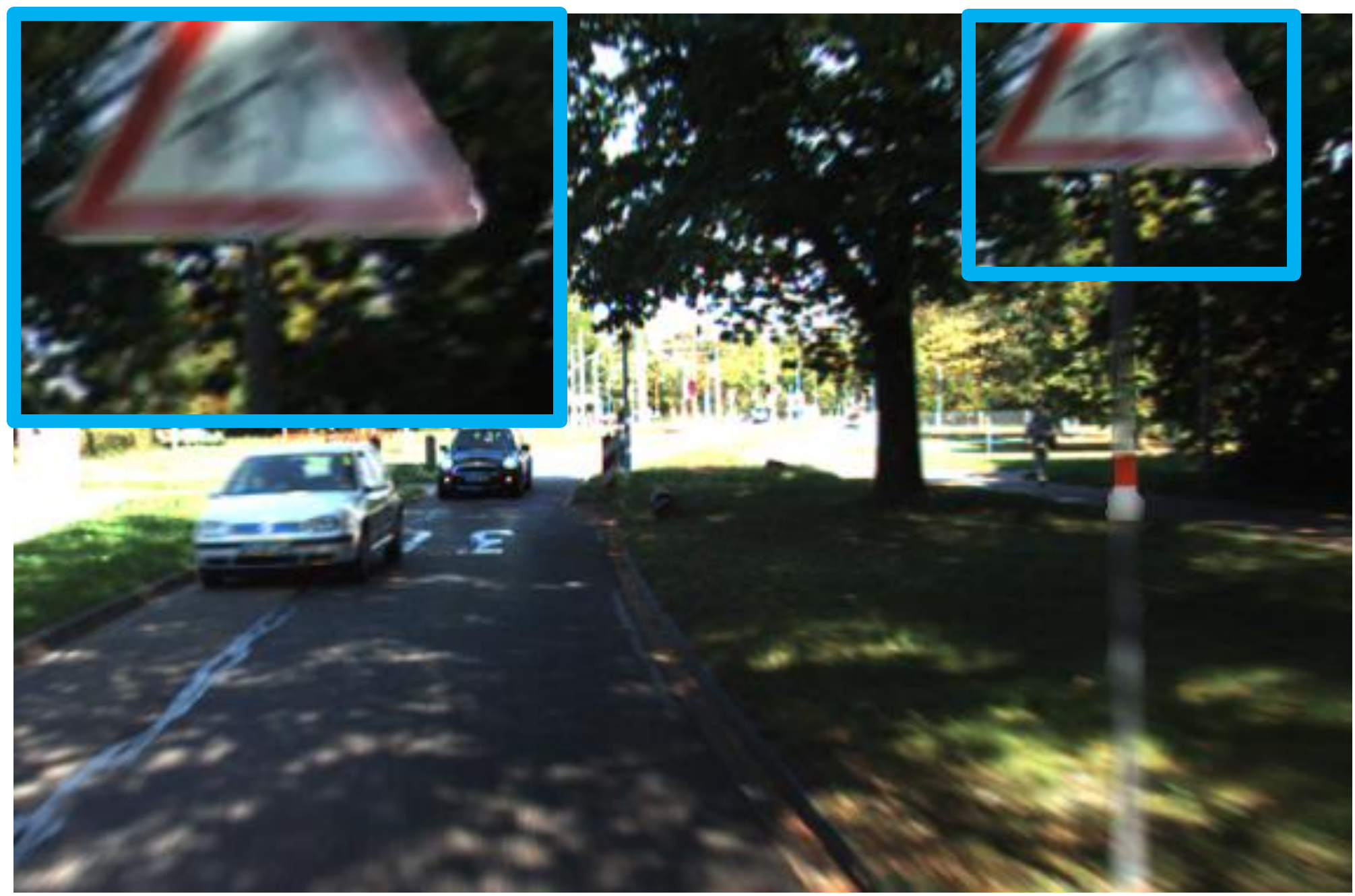}
&\hspace{0.0cm}
\includegraphics[width=0.21\textwidth]{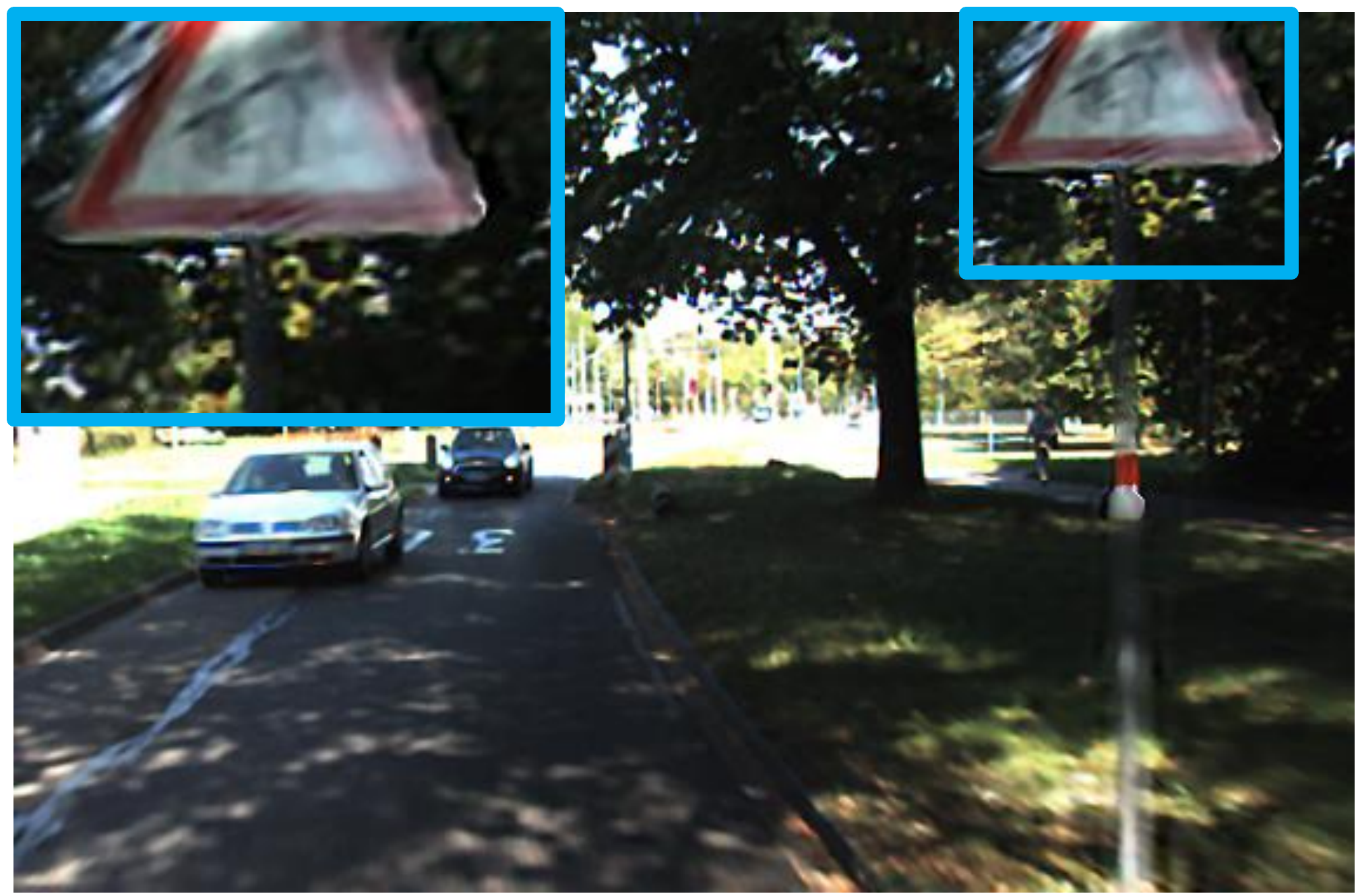}
&\hspace{0.0cm}
\includegraphics[width=0.21\textwidth]{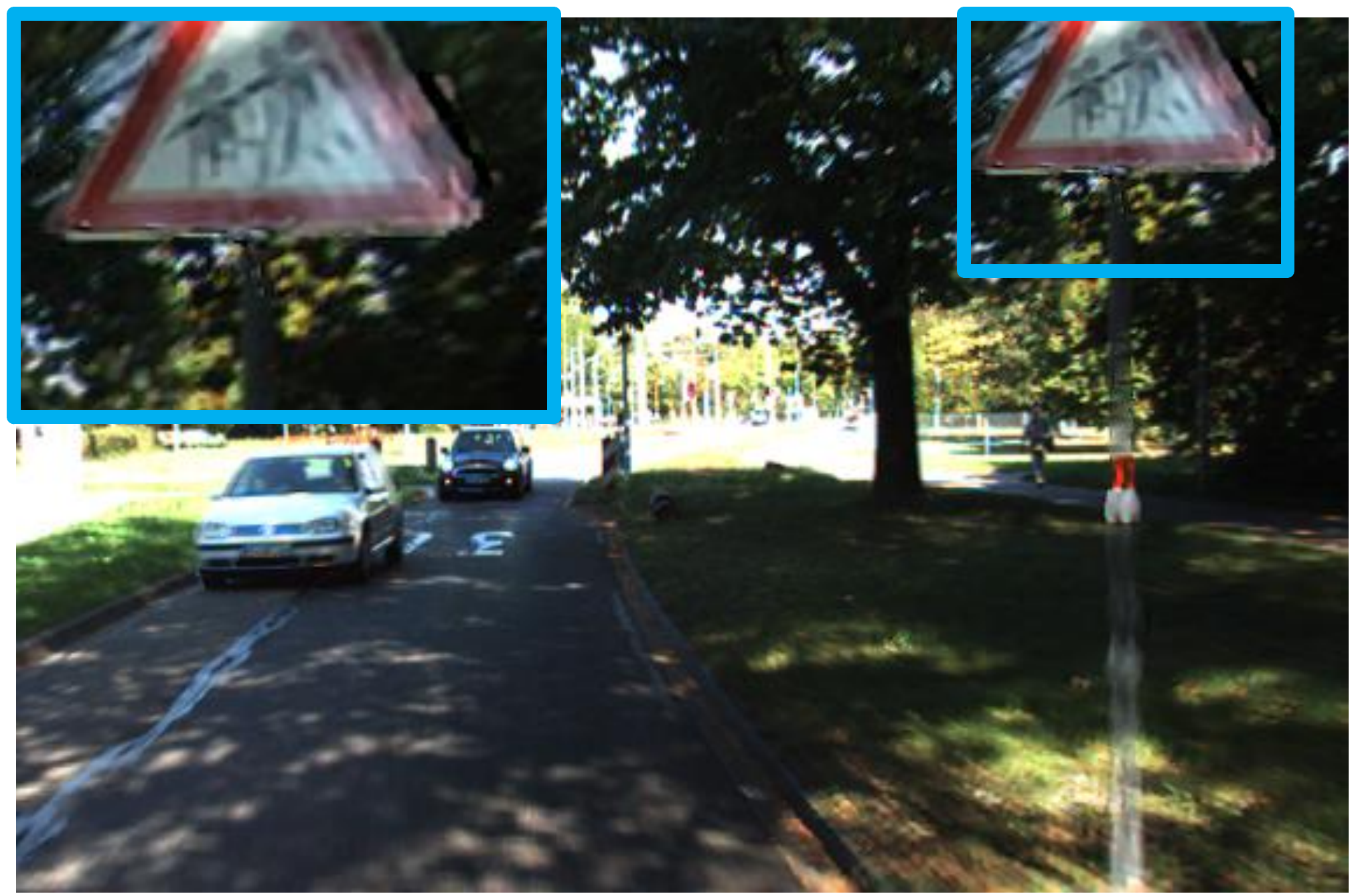}
&\hspace{0.0cm}
\includegraphics[width=0.21\textwidth]{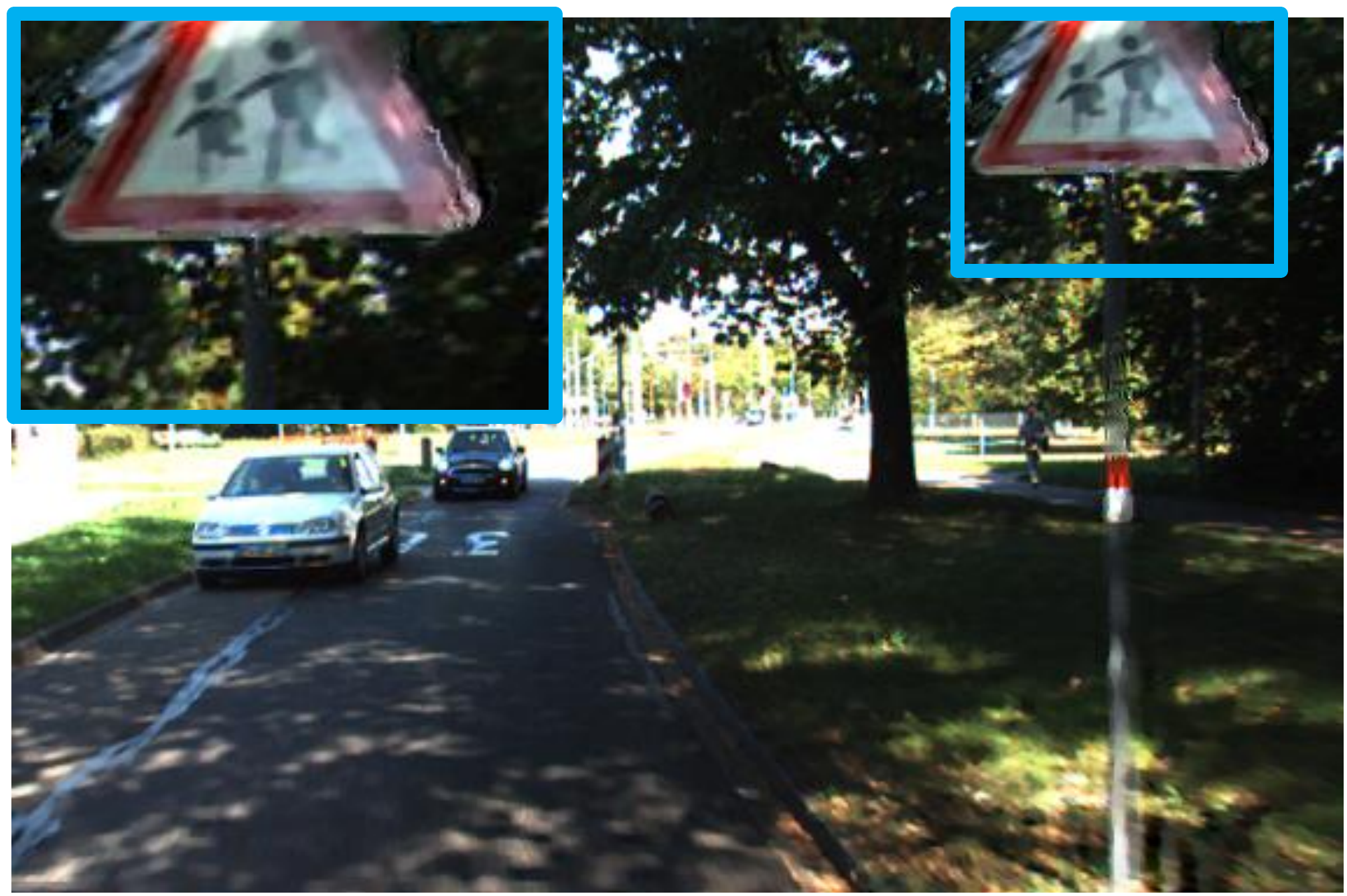}\\
\hspace{0.0cm}
\includegraphics[width=0.21\textwidth]{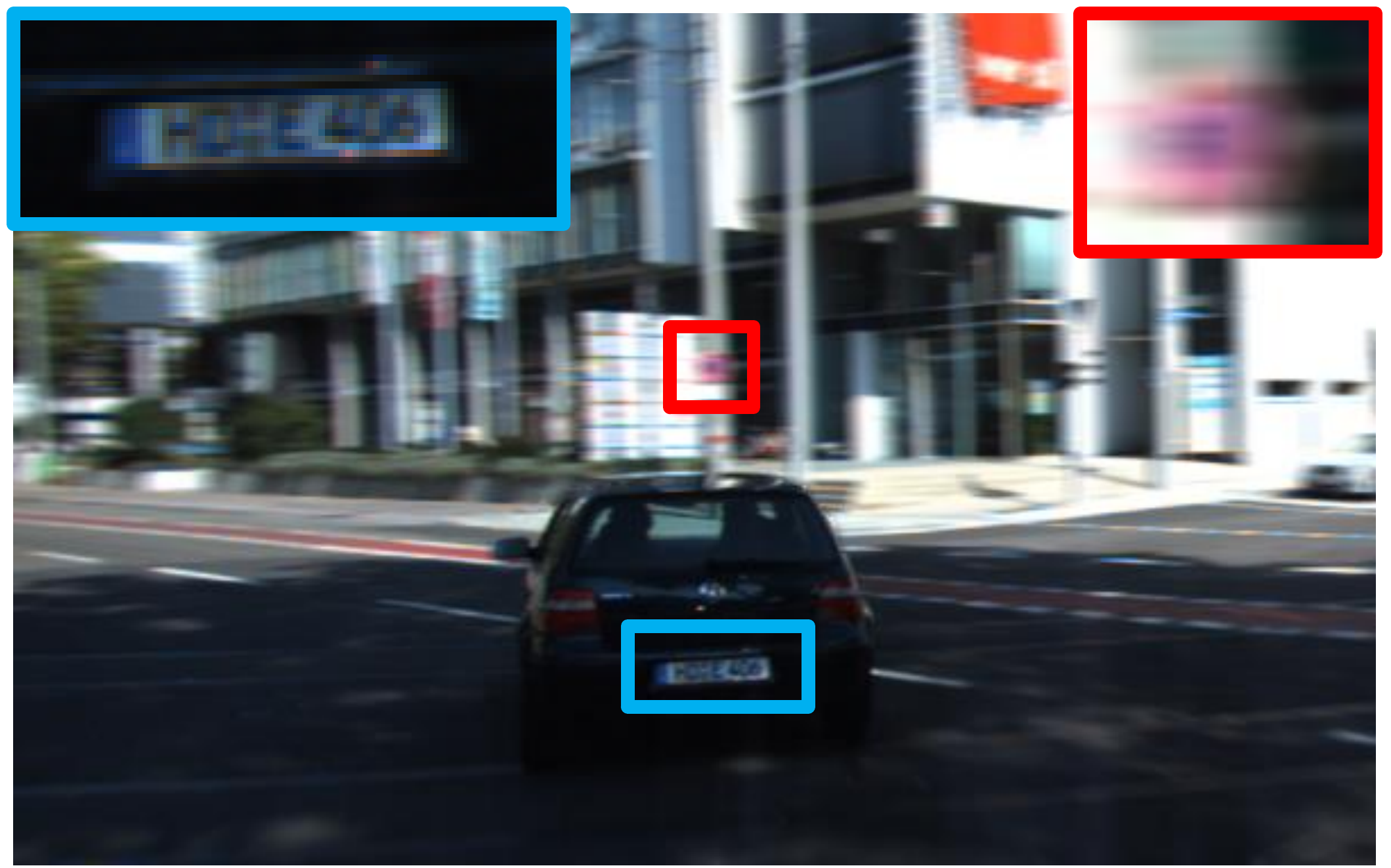}
&\hspace{0.0cm}
\includegraphics[width=0.21\textwidth]{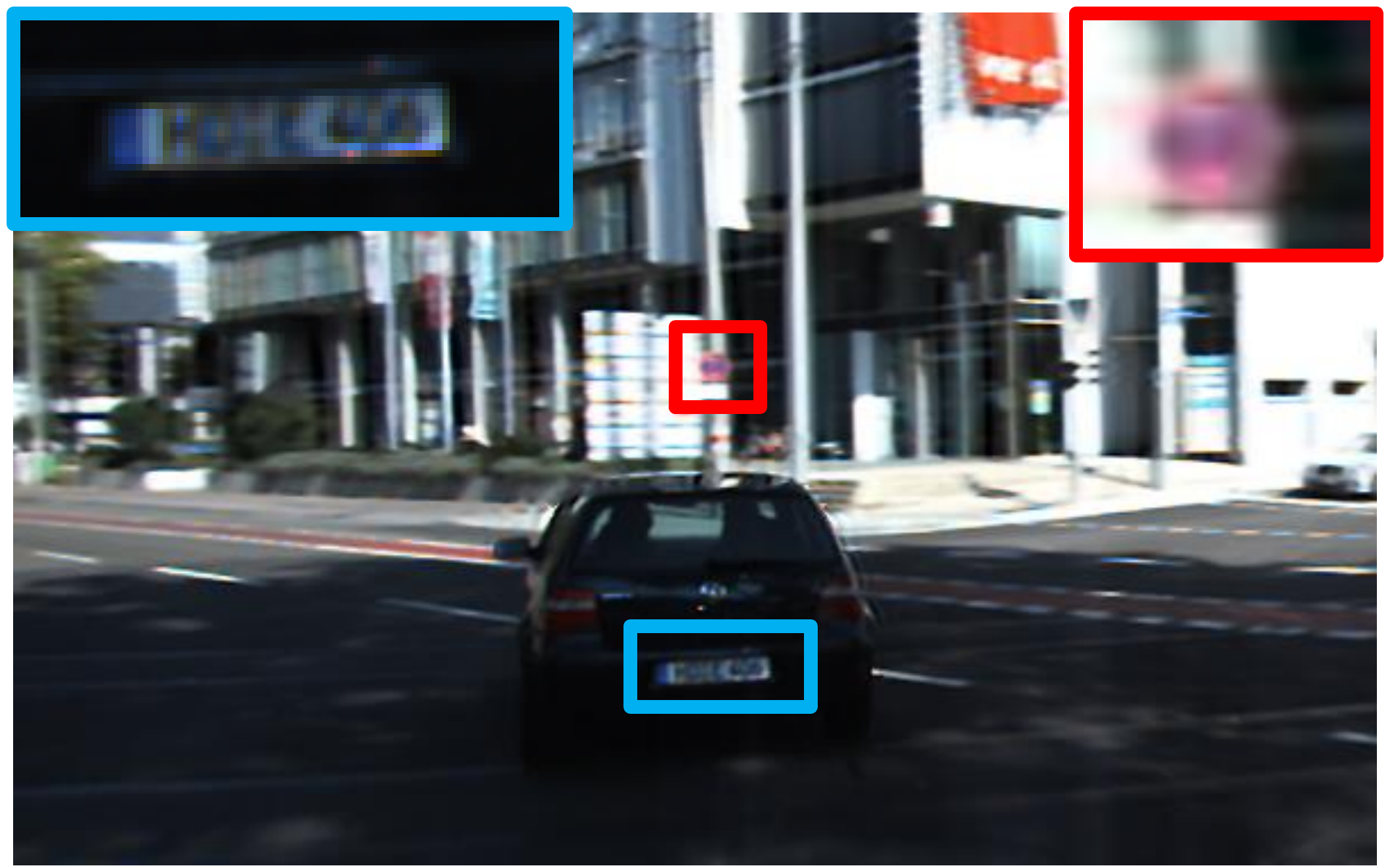}
&\hspace{0.0cm}
\includegraphics[width=0.21\textwidth]{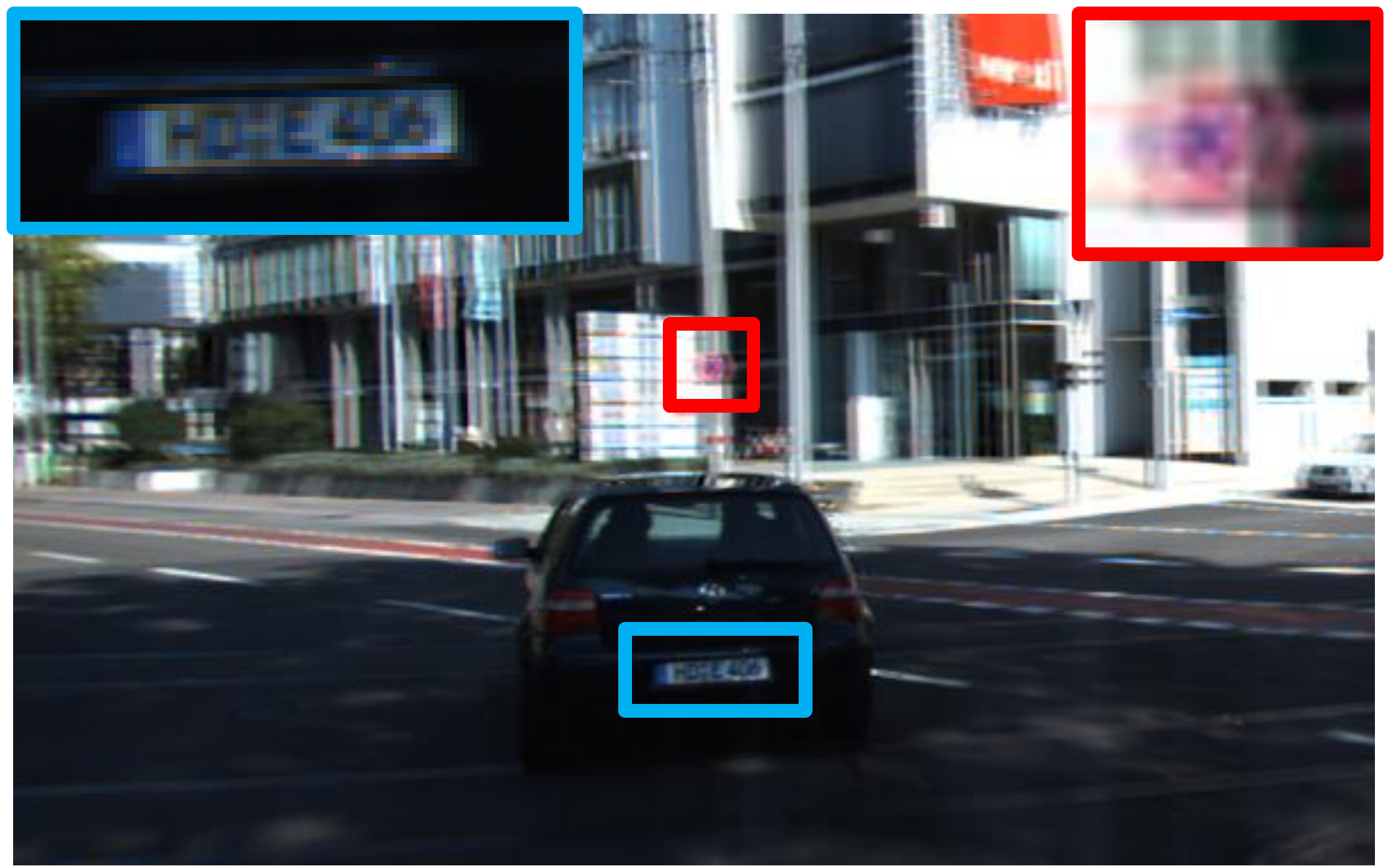}
&\hspace{0.0cm}
\includegraphics[width=0.21\textwidth]{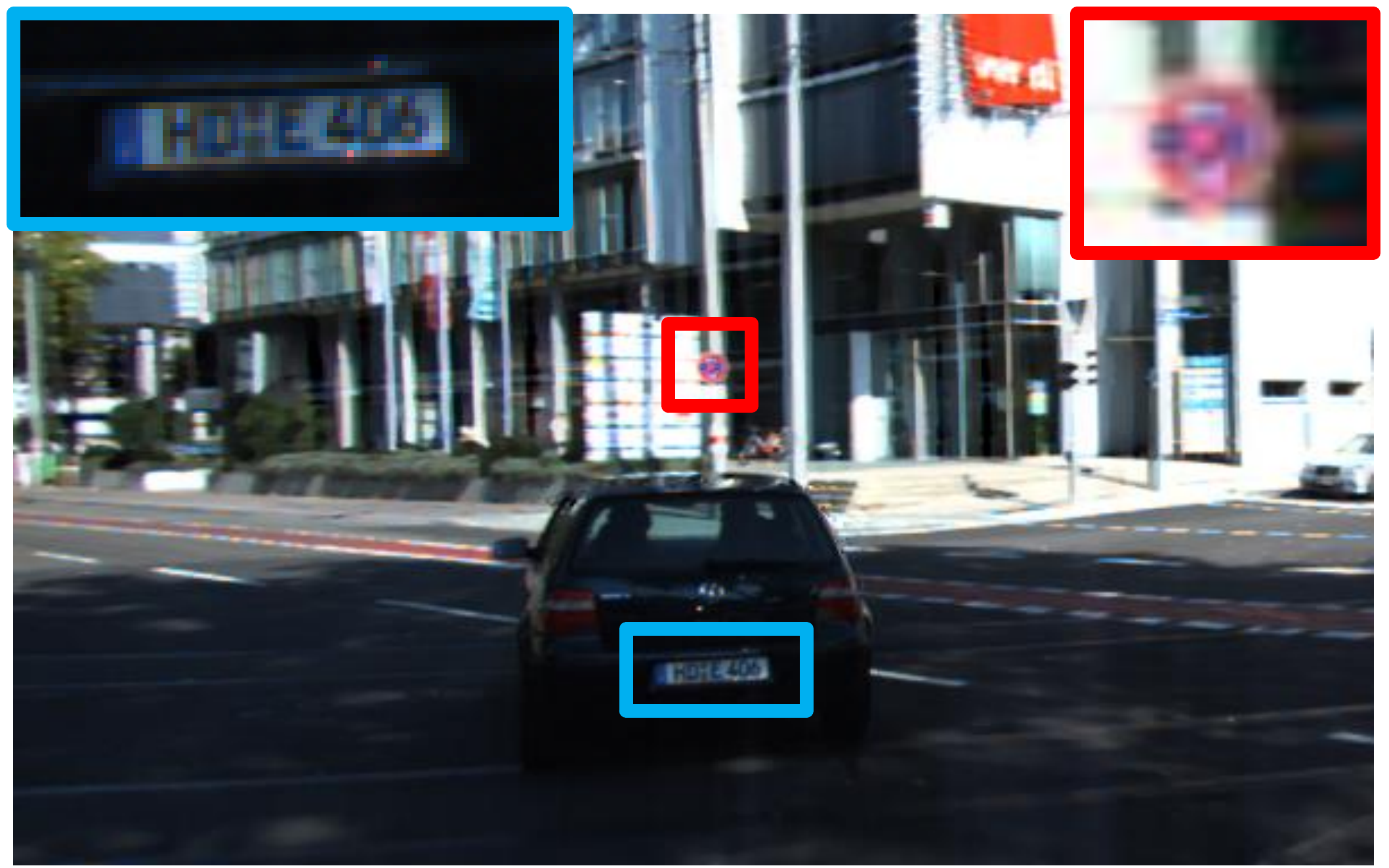}\\
\hspace{0.0cm}
\includegraphics[width=0.21\textwidth]{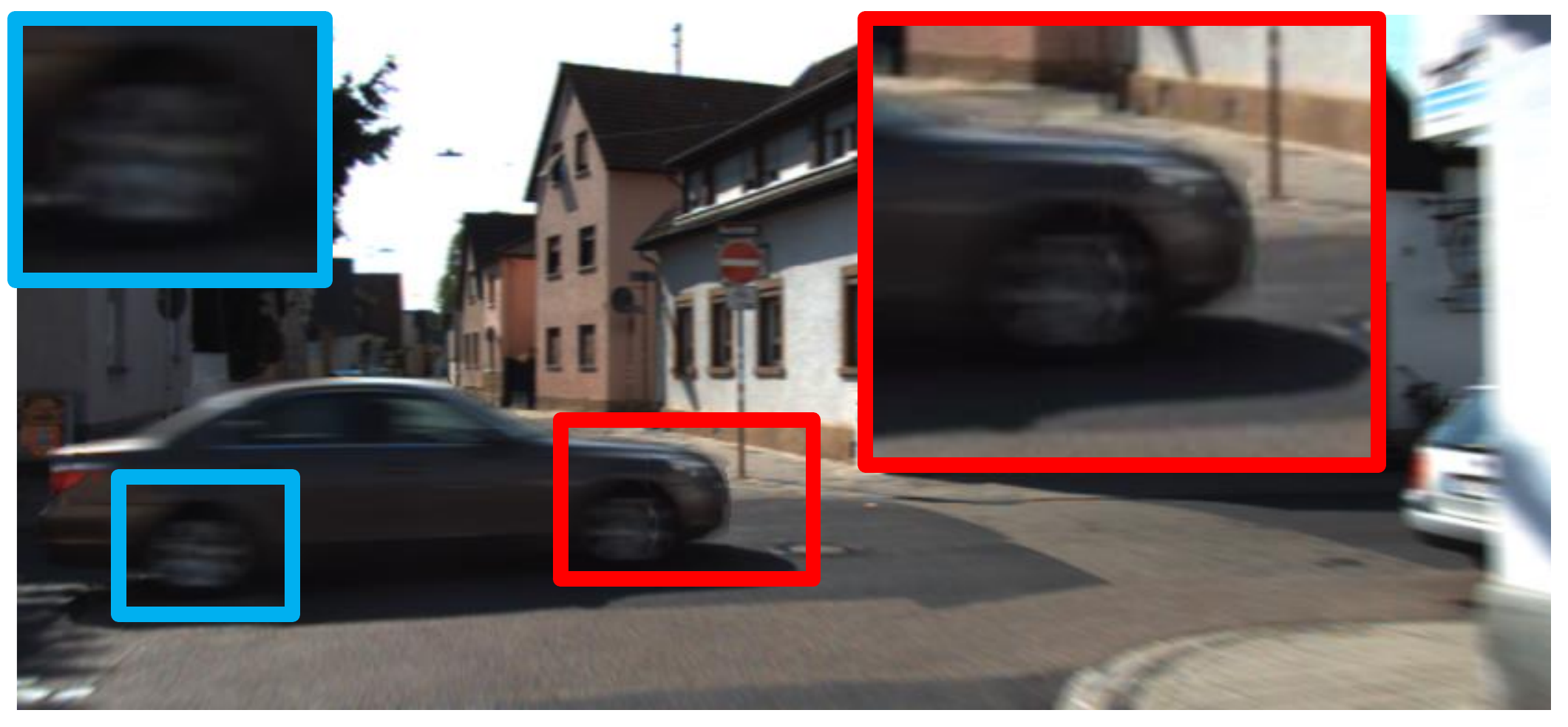}
&\hspace{0.0cm}
\includegraphics[width=0.21\textwidth]{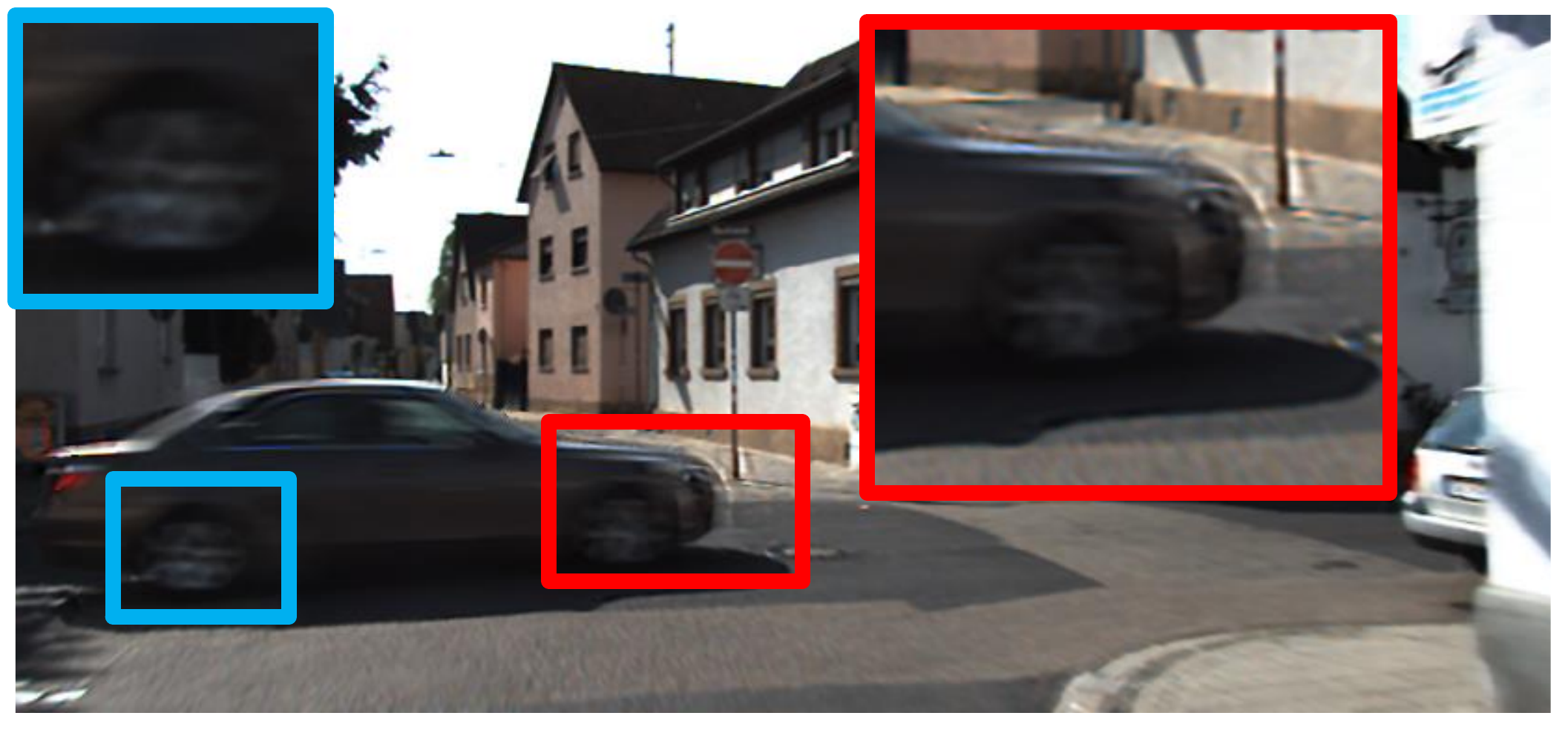}
&\hspace{0.0cm}
\includegraphics[width=0.21\textwidth]{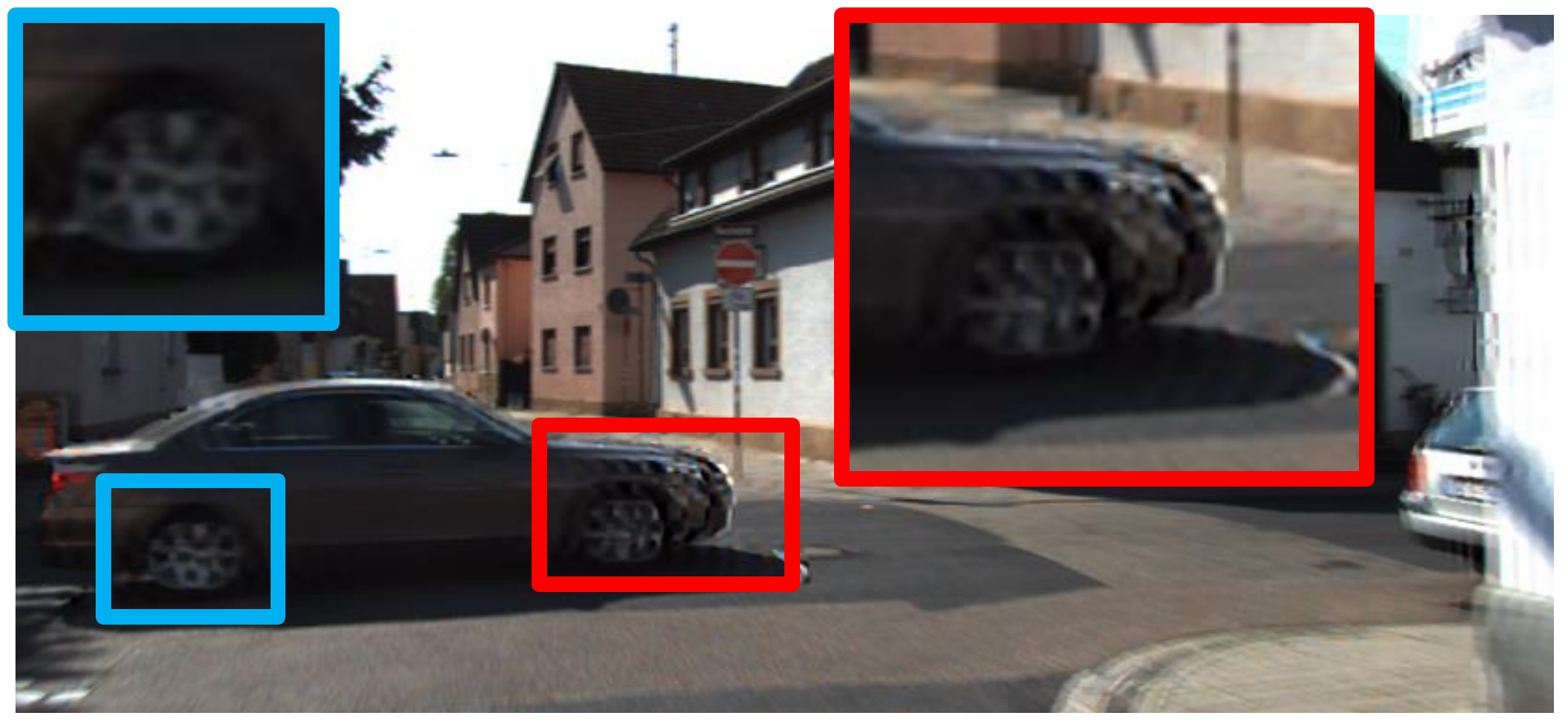}
&\hspace{0.0cm}
\includegraphics[width=0.21\textwidth]{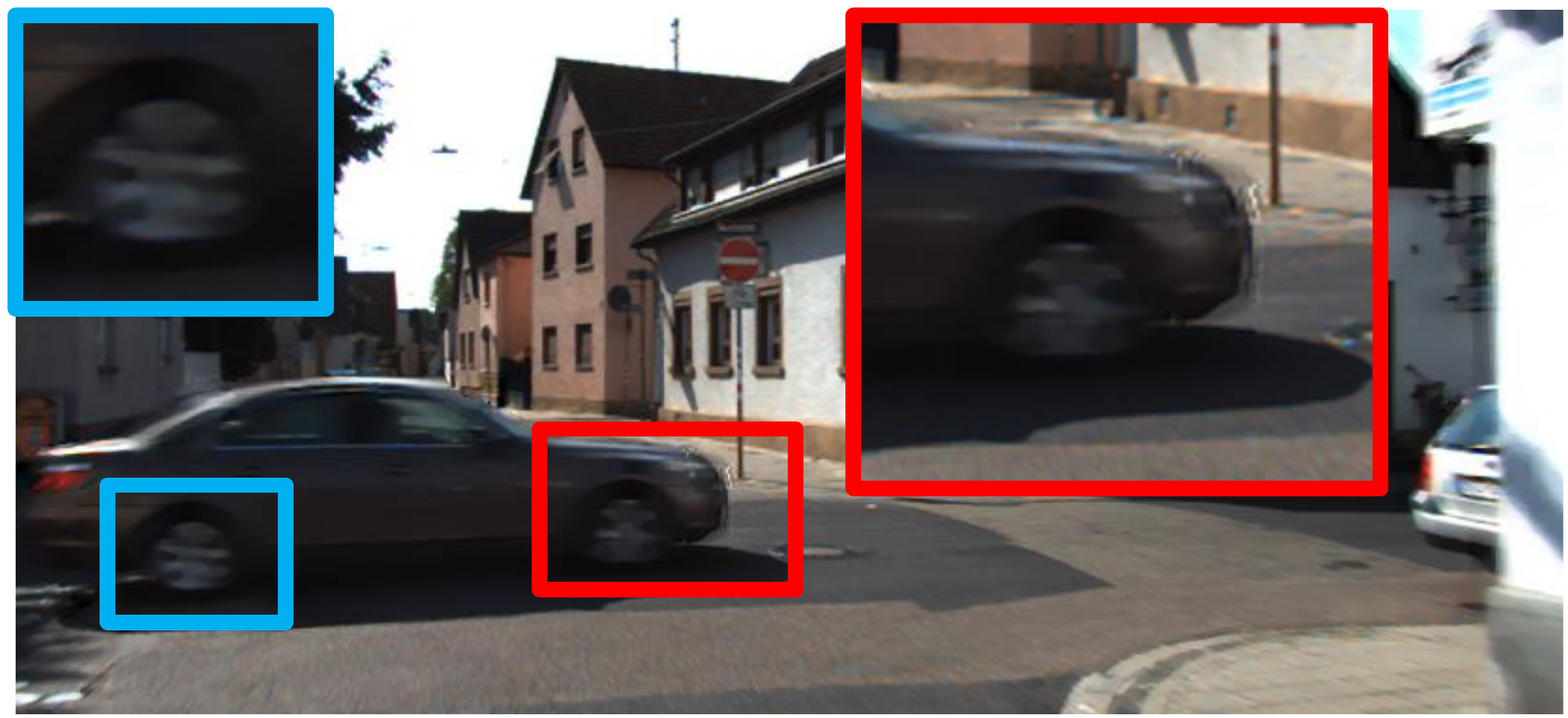}\\
\hspace{0.0cm}
\includegraphics[width=0.21\textwidth]{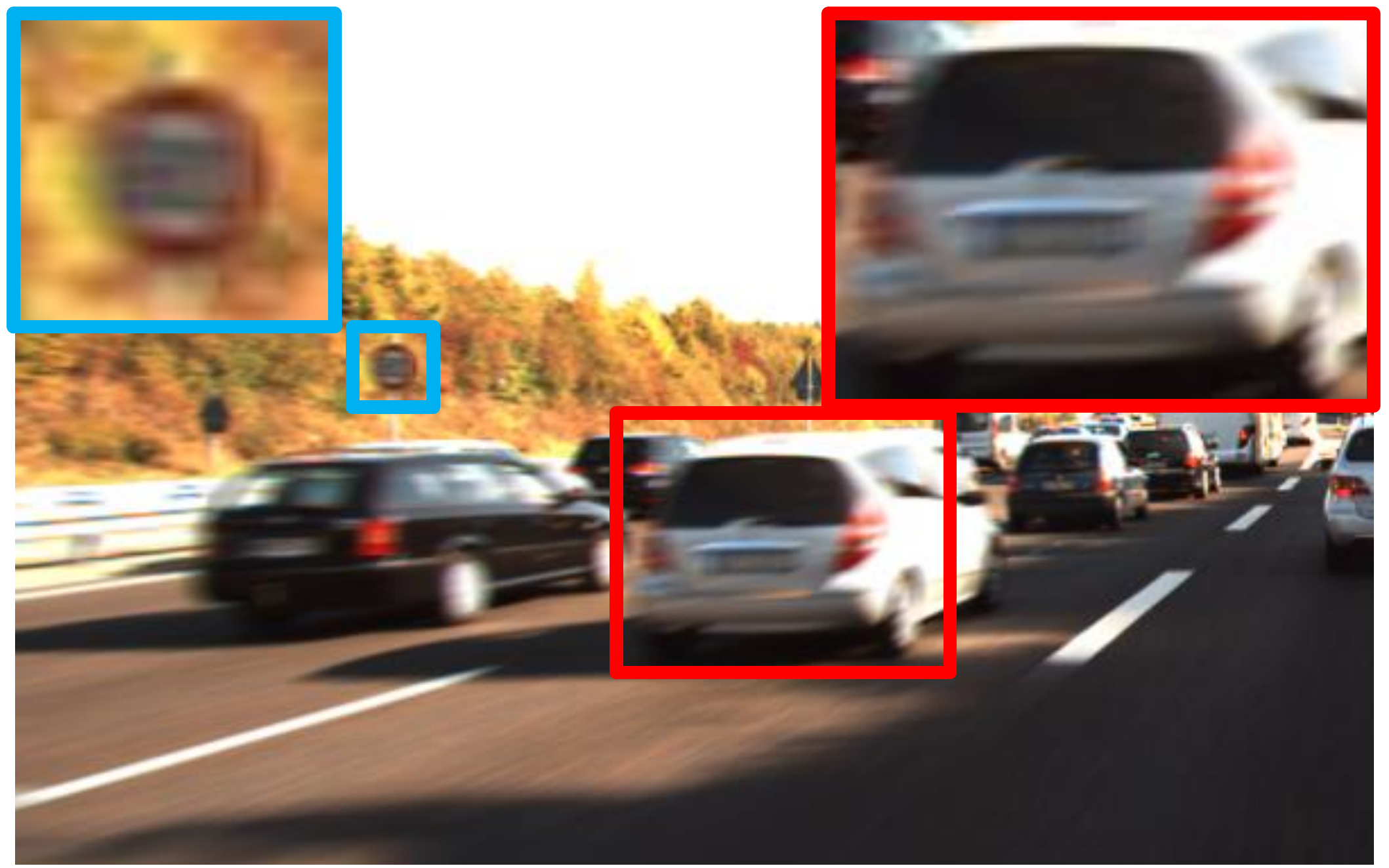}
&\hspace{0.0cm}
\includegraphics[width=0.21\textwidth]{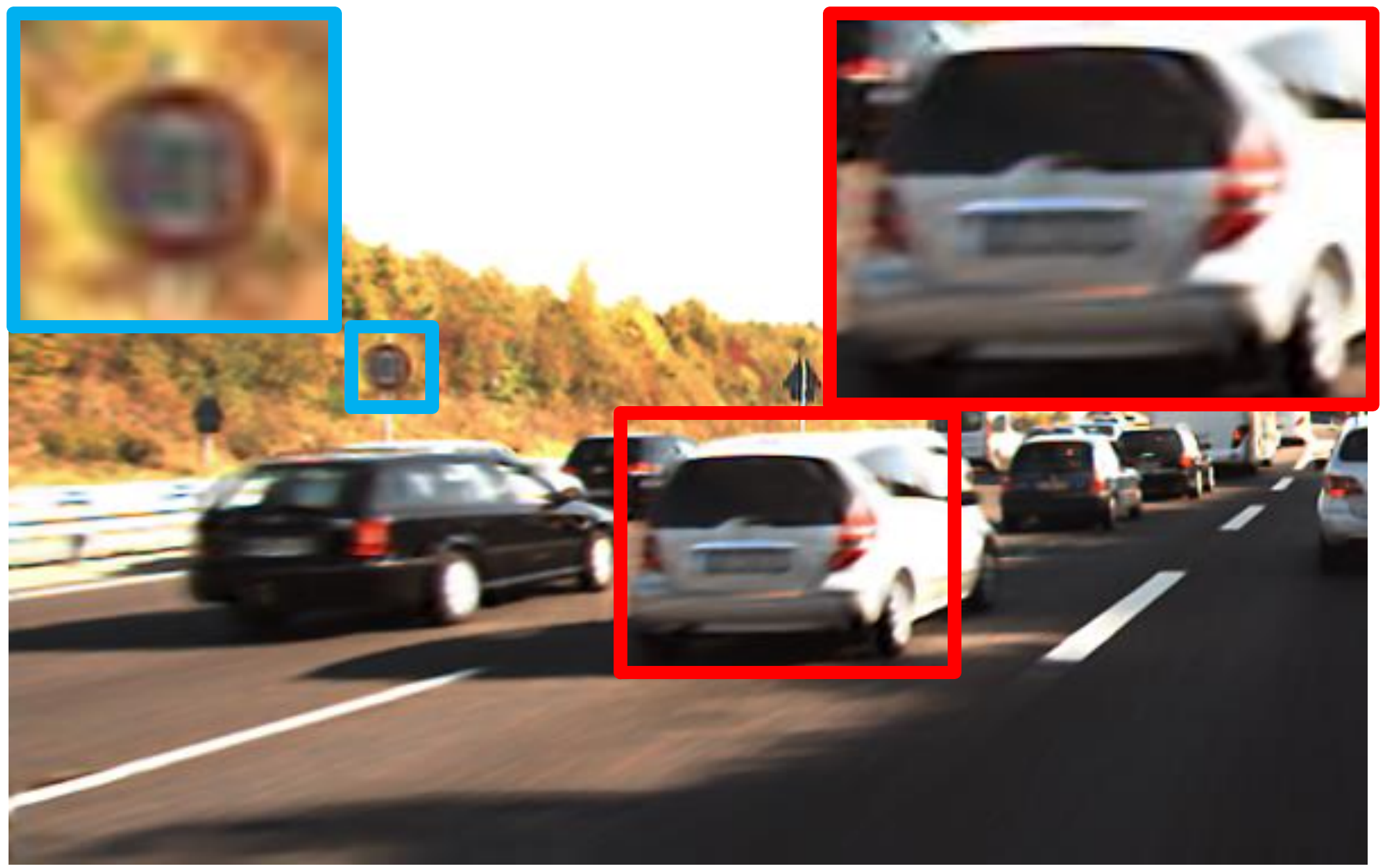}
&\hspace{0.0cm}
\includegraphics[width=0.21\textwidth]{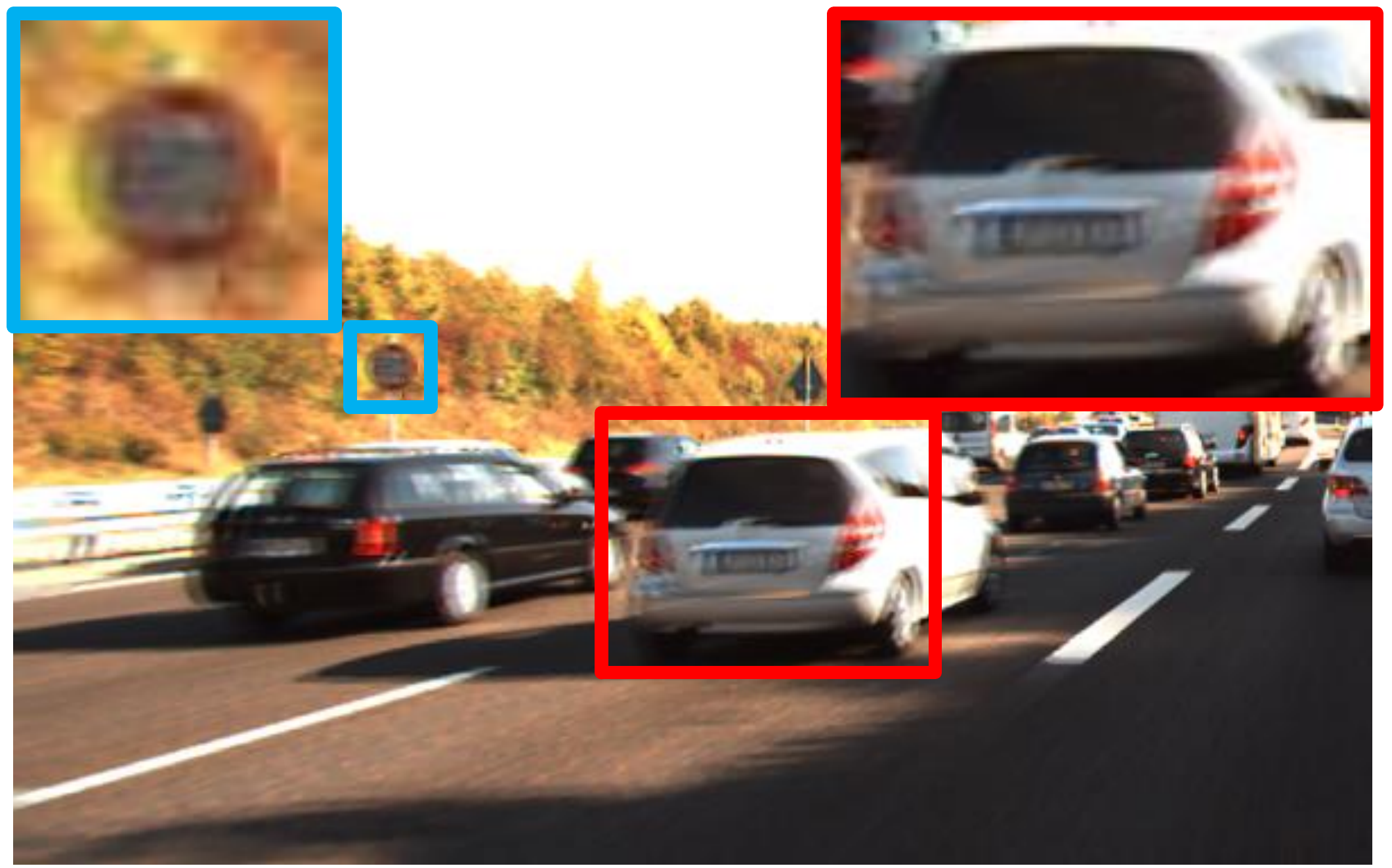}
&\hspace{0.0cm}
\includegraphics[width=0.21\textwidth]{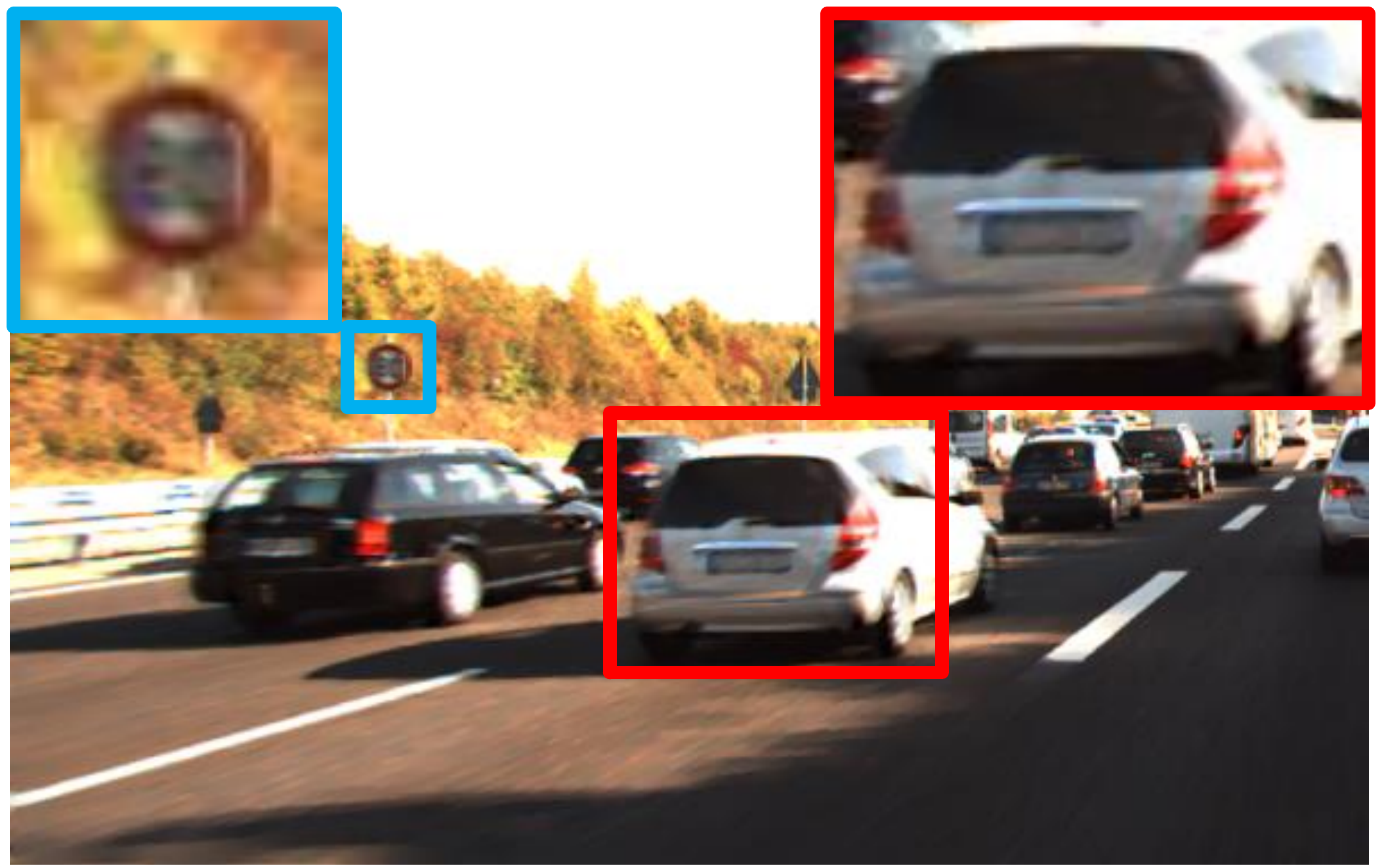}\\
(a) Blur Image & (b) Kim and Lee \cite{hyun2015generalized} & (c) Sellent \etal \cite{sellent2016stereo} & (d) Ours
\end{tabular}
\end{small}
\caption{Numerous outdoor blurry frames and our deblurring result compare with several baselines. Best Viewed on Screen.}
\label{fig:add}
\end{center}
\end{figure*}

\subsection{Experimental Results}
\noindent{\bf Results on KITTI.} To the best of our knowledge, there are no realistic benchmark datasets that provide blur and its corresponding ground-truth clear images and scene flow. In this paper, we take advantage of the KITTI dataset~\cite{geiger2013vision} to create a synthetic \textbf{Blurred KITTI dataset} (will be publicly available) on realistic scenery. It contains 199 scenes, each of which includes 6 images of size $375 \times 1242$. Since the KITTI benchmark does not provide dense ground-truth flow, we use a state-of-the-art scene flow method \cite{menze2015object} to generate dense ground-truth flows. Given the dense scene flow, the blur images are generated by using the piecewise linear 2D kernel, please refer to \cite{hyun2015generalized} and \cite{sellent2016stereo} for more details. The blur is caused by both objects motion and camera motion with occlusion and shadow.

\begin{figure}[t]\begin{center}
\includegraphics[width=0.38\textwidth]{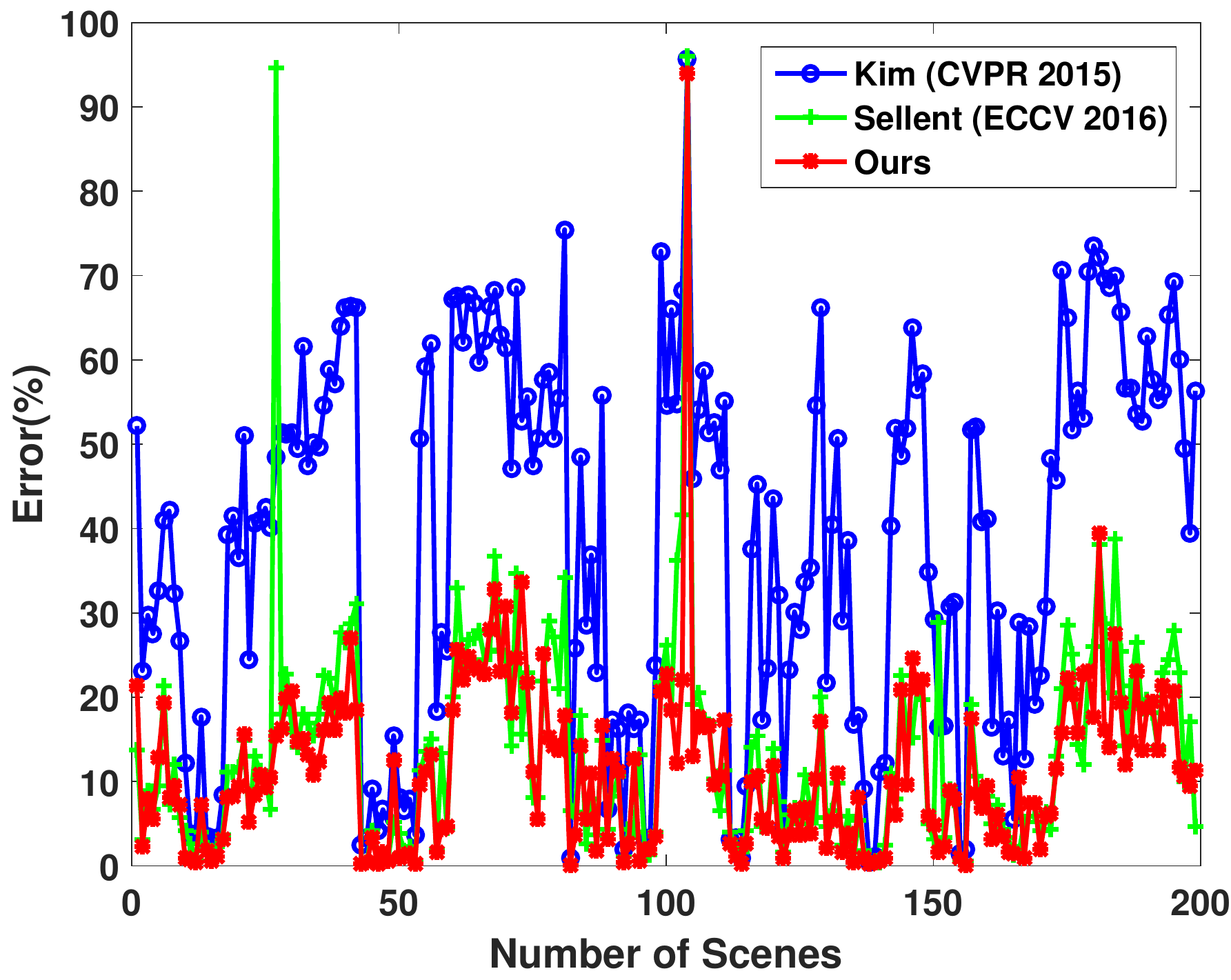}
\end{center}
\vspace{-3mm}
\caption{Flow estimation errors on the Blurred KITTI dataset. Our method clearly outperforms both monocular and stereo video deblurring methods.}
\label{fig:i_err_compare}
\end{figure}

\begin{figure}[t]
\begin{center}
\includegraphics[width=0.38\textwidth]{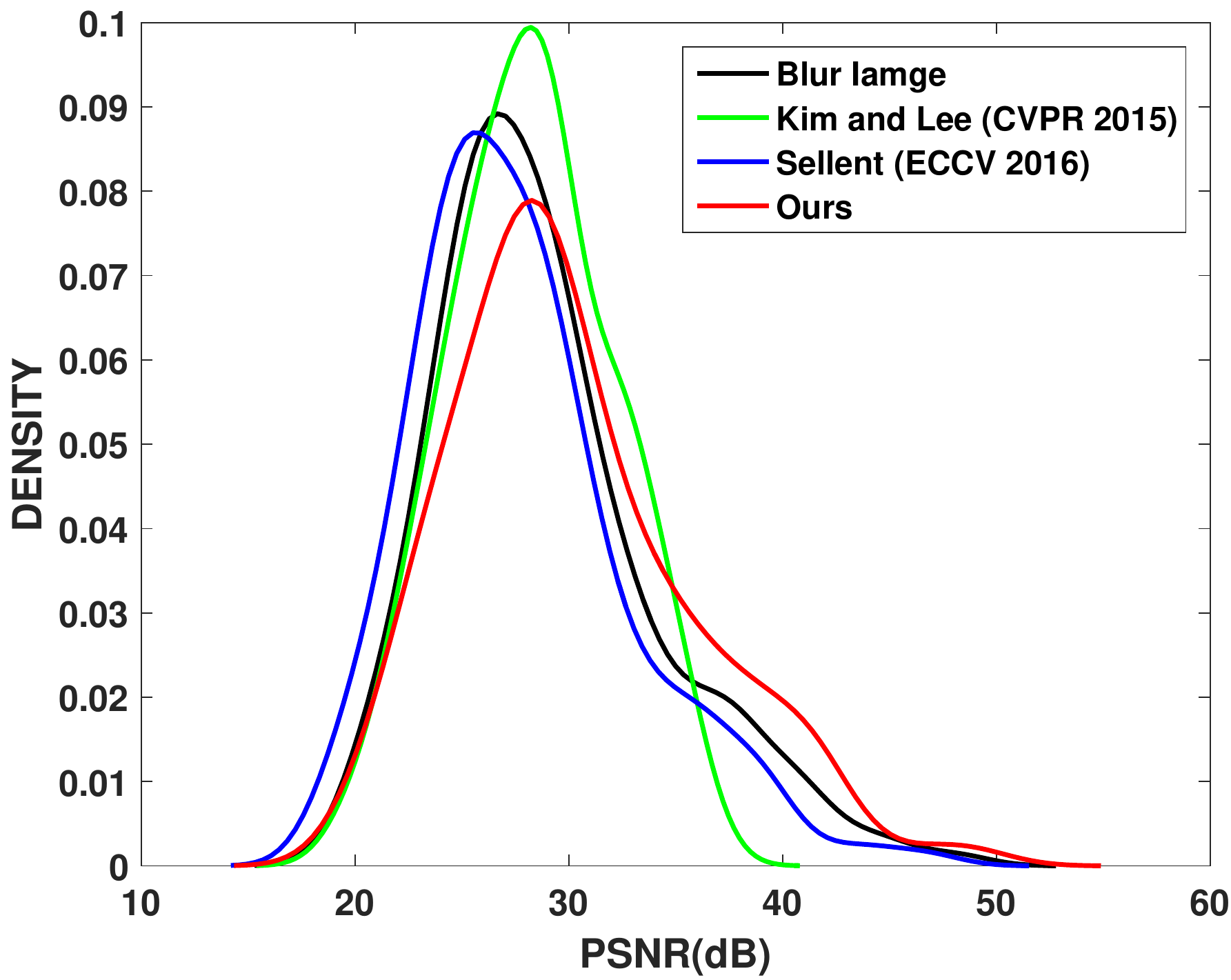}
\end{center}
\vspace{-3mm}
\caption{The distribution of the PSNR scores on the Blurred KITTI dataset. The probability distribution function for each PSNR was estimated using kernel density estimation with a normal kernel function. The heavy tail of our method means larger PSNR can be achieved using our method.}
\label{fig:psnr_compare}
\end{figure}

We evaluated results by averaging errors and PSNR scores over $m-1$ to $m+1$ stereo image pairs. Table~\ref{all_all} shows the PSNR values, disparity errors, and flow errors averaged over the Blurred KITTI dataset. Our method consistently outperforms all baselines. We achieve the minimum error scores of 10.01\% for optical flow and 6.82\% for disparity in the reference view.  In Fig.~\ref{fig:add}, we show qualitative results of our method and other methods on sample sequences from our dataset. Fig.~\ref{fig:i_err_compare} and Fig.~\ref{fig:psnr_compare} show the scene flow estimation and deblurring results of the Blurred KITTI dataset.

\begin{figure}[t]
\begin{center}
\includegraphics[width=0.234\textwidth]{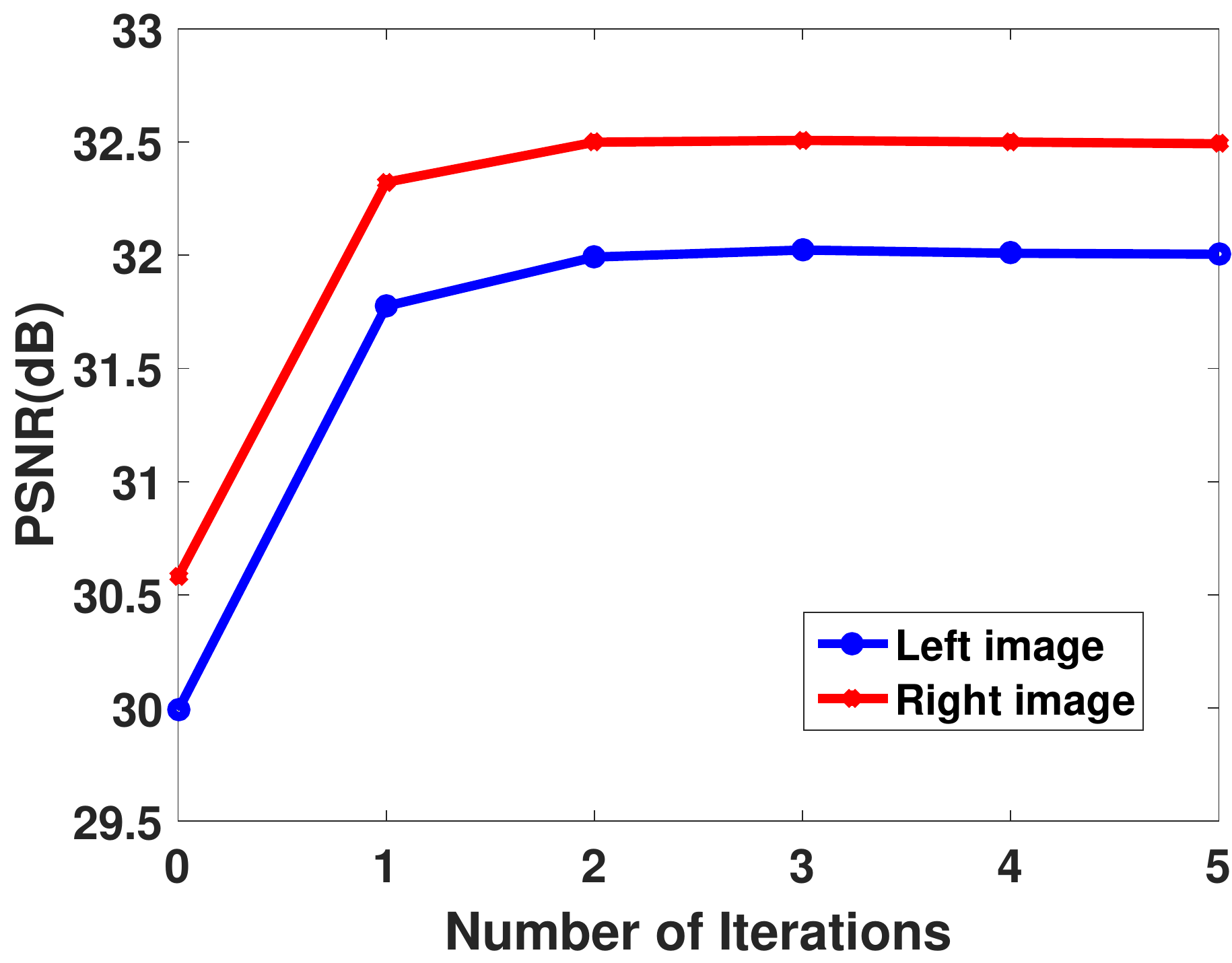}
%
\includegraphics[width=0.228\textwidth]{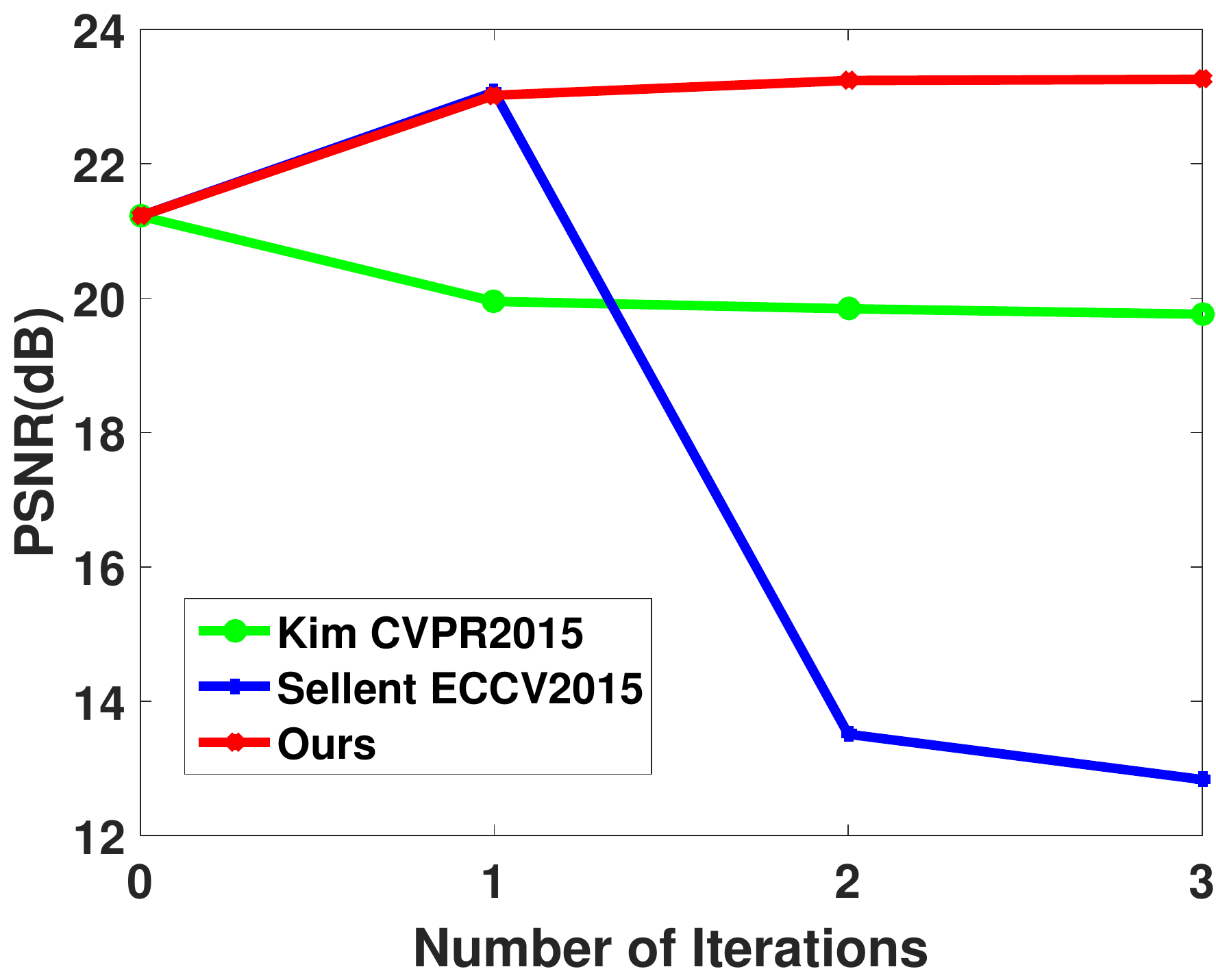}
\end{center}
\caption{Deblurring performance with respect to iterations. (left) Our method gains an improvement of 0.3dB between the first and the last iteration on the 50 challenging dataset. (right) Comparison between our method and other baselines on the 'Chair' sequence.}
\label{fig:iter_psnr}
\end{figure}

We then choose a subset of 50 more challenging sequences with large motion from the 199 scenes as test images, which contains daily traffic scenes covering urban areas (30 sequences), rural areas (10 sequences) and highway (10 sequences). Table~\ref{all_our} shows the PSNR values, disparity errors, and flow errors averaged over 50 test sequences on Blurred KITTI dataset. Fig.~\ref{fig:iter_psnr} (left) shows the performance of our deblurring stage with respect to the number of iterations. While we use 5 iterations for all our experiments, our experiments indicate that only 3 iterations are sufficient in most cases to achieve optimal performance under our model.
\begin{table}\footnotesize
\centering
\caption{Quantitative comparisons on the Blurred KITTI dataset. 
}
\label{all_all}
\begin{tabular}{c|c|c|c|c|c|c|c}
\hline
\multicolumn{2}{c|}{\multirow{2}{*}{KITTI Dataset}} & \multicolumn{2}{c|}{Disparity} & \multicolumn{2}{c|}{Flow} & \multicolumn{2}{c}{PSNR} \\ \cline{3-8}
\multicolumn{2}{c|}{}                                      & m              & m+1           & Left        & Right       & Left        & Right       \\ \hline
\multicolumn{2}{c|}{Vogal \etal \cite{vogel20153d}}        & 8.20           & 8.50          & 13.62       & 14.59       & /           & /           \\ \hline
\multicolumn{2}{c|}{Kim and Lee \cite{hyun2015generalized}}& /              & /             & 38.89       & 39.45       & 28.25       & 29.00       \\ \hline
\multicolumn{2}{c|}{Sellent \etal \cite{sellent2016stereo}}& 8.20           & 8.50          & 13.62       & 14.59       & 27.75       & 28.52       \\ \hline
\multirow{2}{*}{Ours}     & 2 Frames                       & 7.02           & 8.55          & 11.44       & 19.34       & \bf30.24    & \bf30.71       \\ \cline{2-8}
                          & 3 Frames                       & \bf 6.82       &  \bf 8.36     & \bf 10.01   & \bf 11.45   & 29.80      & 30.30   \\ \hline
\end{tabular}
\end{table}

\begin{table}\footnotesize
\centering
\caption{Quantitative comparisons on 50 challenging sequences. }
\label{all_our}
\begin{tabular}{c|c|c|c|c|c|c|c}
\hline
\multicolumn{2}{c|}{\multirow{2}{*}{Our Dataset}} & \multicolumn{2}{c|}{Disparity} & \multicolumn{2}{c|}{Flow} & \multicolumn{2}{c}{PSNR} \\ \cline{3-8}
\multicolumn{2}{c|}{}                                      & m              & m+1           & Left        & Right       & Left        & Right       \\ \hline
\multicolumn{2}{c|}{Vogal \etal \cite{vogel20153d}}        & 6.67           & 6.70          & 7.26       & 7.90        & /           & /           \\ \hline
\multicolumn{2}{c|}{Kim and Lee \cite{hyun2015generalized}}& /              & /             & 25.83      & 26.36       & 29.58       & 30.30       \\ \hline
\multicolumn{2}{c|}{Sellent \etal \cite{sellent2016stereo}}& 6.67           & 6.70          & 7.26       & 7.90        & 28.73       & 29.44       \\ \hline
\multirow{2}{*}{Ours}     & 2 Frames                       & 4.98           & 5.82          & \bf 6.12   & 13.06       & \bf 32.22   & \bf 32.62   \\ \cline{2-8}
                          & 3 Frames                       & \bf 4.90       & \bf 5.76      & 6.16   &  \bf 6.17       & 31.80       & 32.28       \\ \hline
\end{tabular}
\end{table}

\begin{figure}[h]
\begin{center}
\begin{small}
\begin{tabular}{cc}
\hspace{0.0cm}
\includegraphics[width=0.21\textwidth]{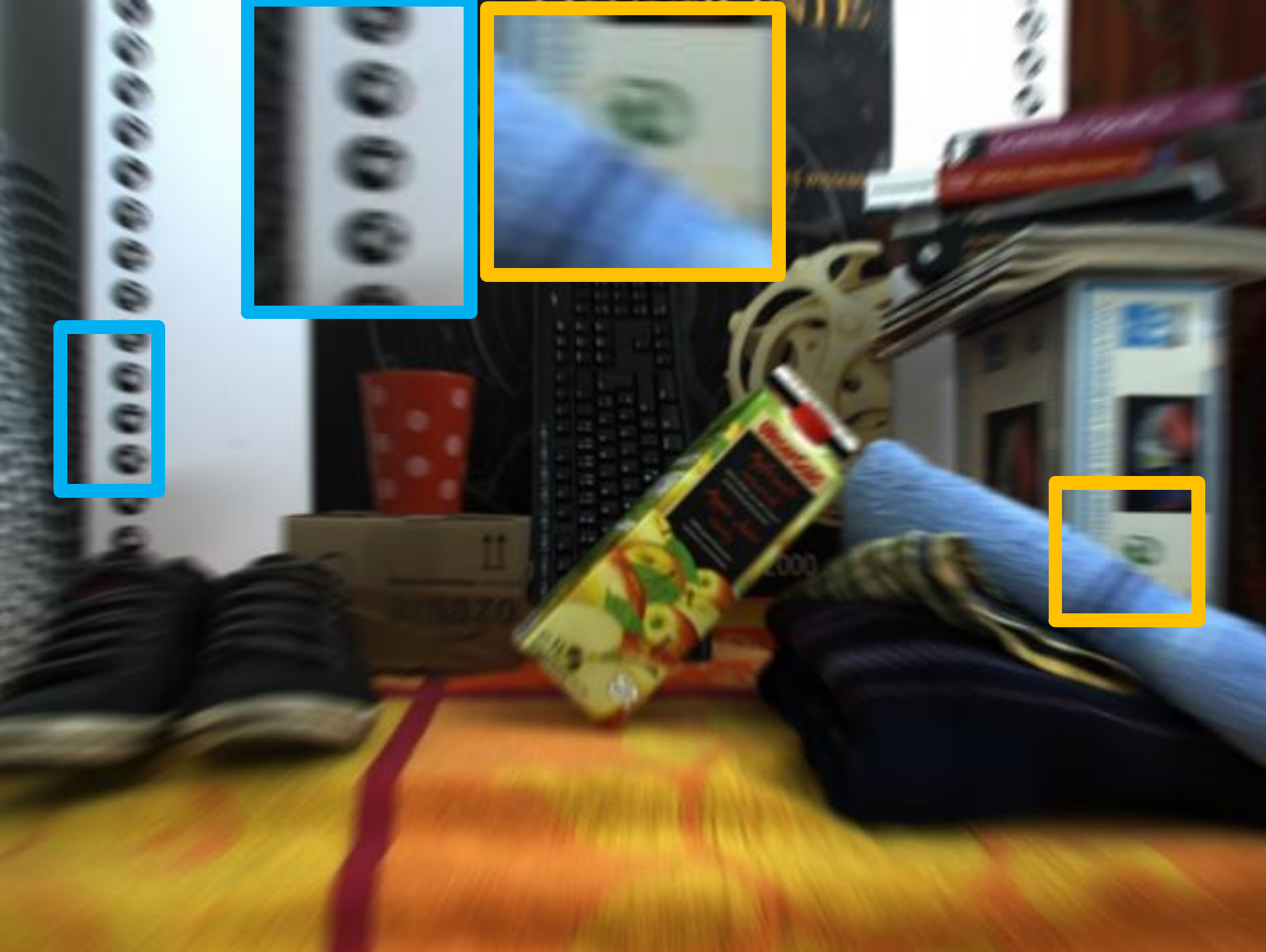}
&\hspace{0.0cm}
\includegraphics[width=0.21\textwidth]{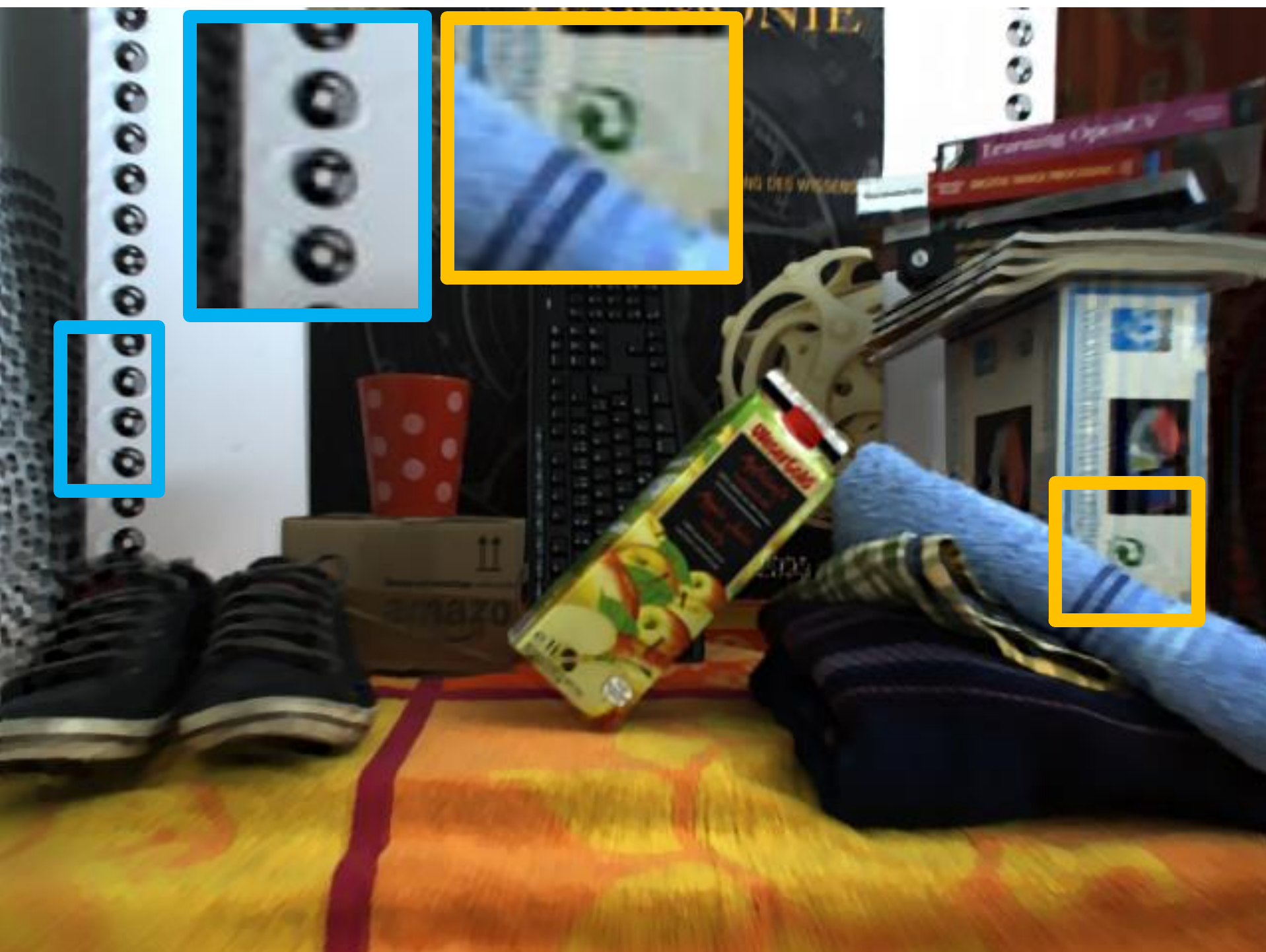}\\
\hspace{0.0cm}
\includegraphics[width=0.21\textwidth]{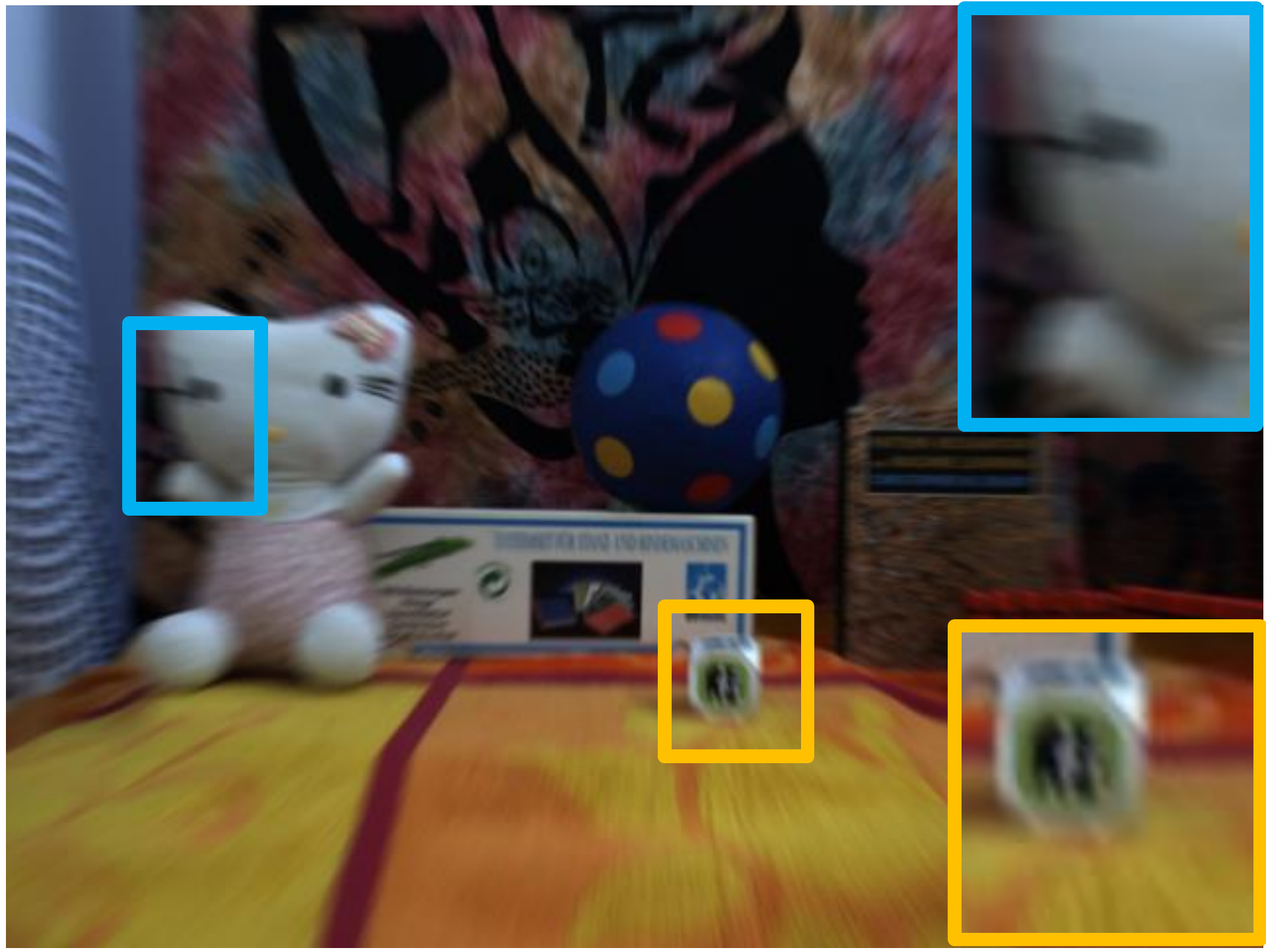}
&\hspace{0.0cm}
\includegraphics[width=0.21\textwidth]{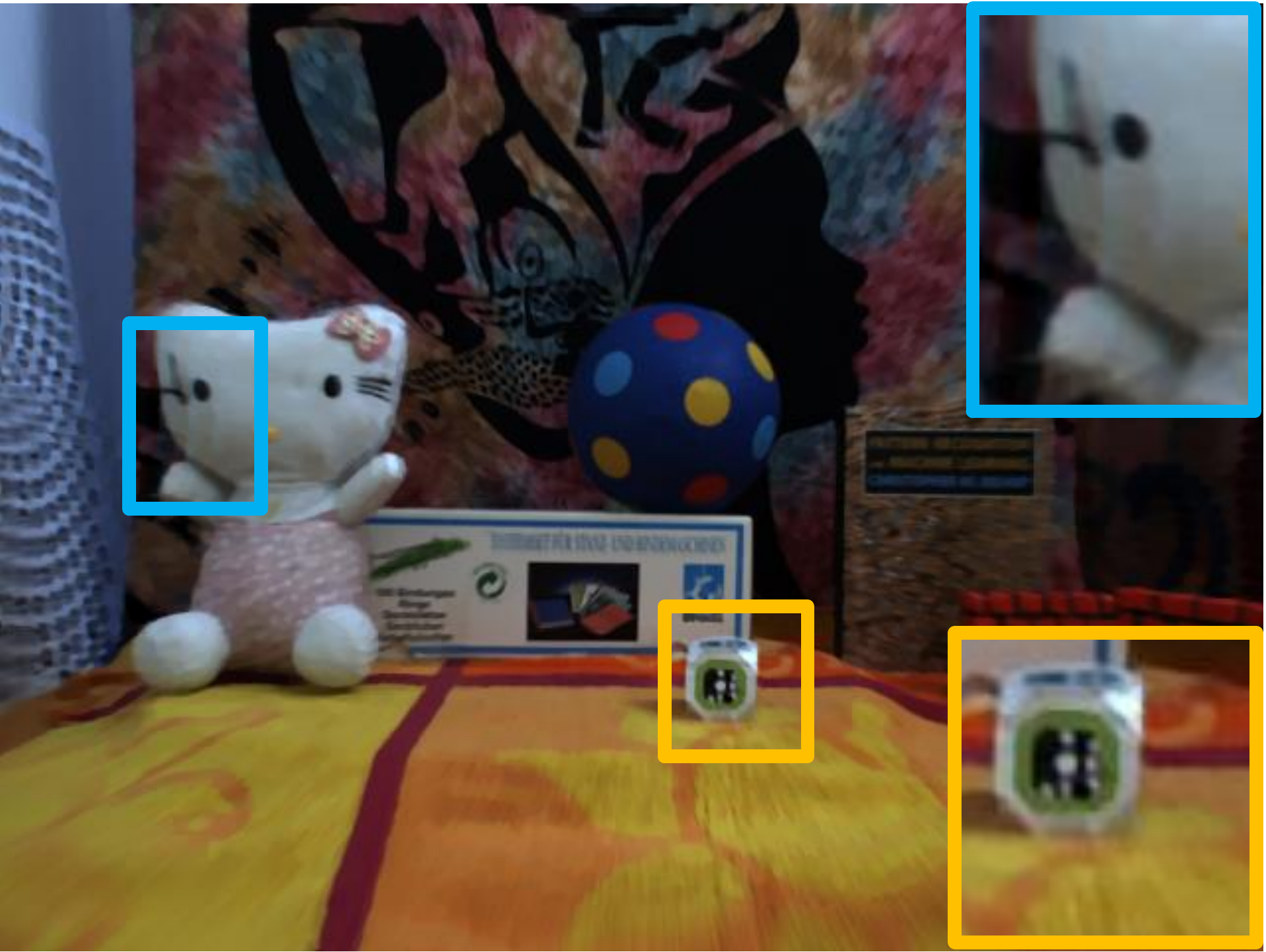}\\
(a) Blur Images & (b) Our results\\
\end{tabular}
\end{small}
\caption{Sample deblur results on the real image dataset from Sellent et al.\cite{sellent2016stereo}. Best Viewed on Screen.}
\label{fig:realeccv16}
\end{center}
\end{figure}

\noindent{\bf Results on Sellent \etal \cite{sellent2016stereo} dataset}
We further evaluate our approach on the dataset in \cite{sellent2016stereo} where the blur images are generated by 3D kernel model.~Those sequences contain four real and four synthetic scenes and each of them includes six blur images with its sharp images, where ground-truth scene flow is only available for the synthetic scene ``Chair''. We thus report the quantitative comparison in Table \ref{chair} on the scene ``Chair'' between our method and state-of-the-art methods, where the evaluation results are averaged over 4 images. We also present the qualitative results in Fig.~\ref{fig:realeccv16} for real images in this dataset. Fig.~\ref{fig:iter_psnr} (right) shows the performance comparison in deblurring between our method and other baselines with respect to iterations on scene ``Chair''. These results affirm our assumption that simultaneously solving scene flow and video deblur benefit each other and that a simple combination of two stages cannot achieve the targeted results.



\begin{table}\footnotesize
\centering
\caption{Performance comparisons on scene ``Chair'' \cite{sellent2016stereo}.
}
\label{chair}
\begin{tabular}{c|c|c|c|c}
\hline
\multicolumn{2}{c|}{Chair video}                            & Disparity(\%)   & Flow Error(\%) & PSNR(dB) \\ \hline
\multicolumn{2}{c|}{Menze \cite{menze2015object}}          & 1.17            & 9.33           & /        \\ \hline
\multicolumn{2}{c|}{Vogel \cite{vogel20153d}}               & 1.34            & 2.13           & /        \\ \hline
\multicolumn{2}{c|}{Kim  \cite{hyun2015generalized}}        & /               & 9.08           & 19.95    \\ \hline
\multicolumn{2}{c|}{Sellent \cite{sellent2016stereo}}       & 1.34            & 2.13           & 23.07    \\ \hline
\multirow{2}{*}{Ours}      & 2 Frames                       & 1.28            & 1.22           & 23.13    \\  \cline{2-5}
                           & 3 Frames                       & \bf 1.15        & \bf 1.18       & \bf 23.26\\ \hline
\end{tabular}
\end{table}

\begin{figure}
\begin{center}
\begin{small}
\begin{tabular}{cc}
\hspace{0.0cm}
\includegraphics[width=0.21\textwidth]{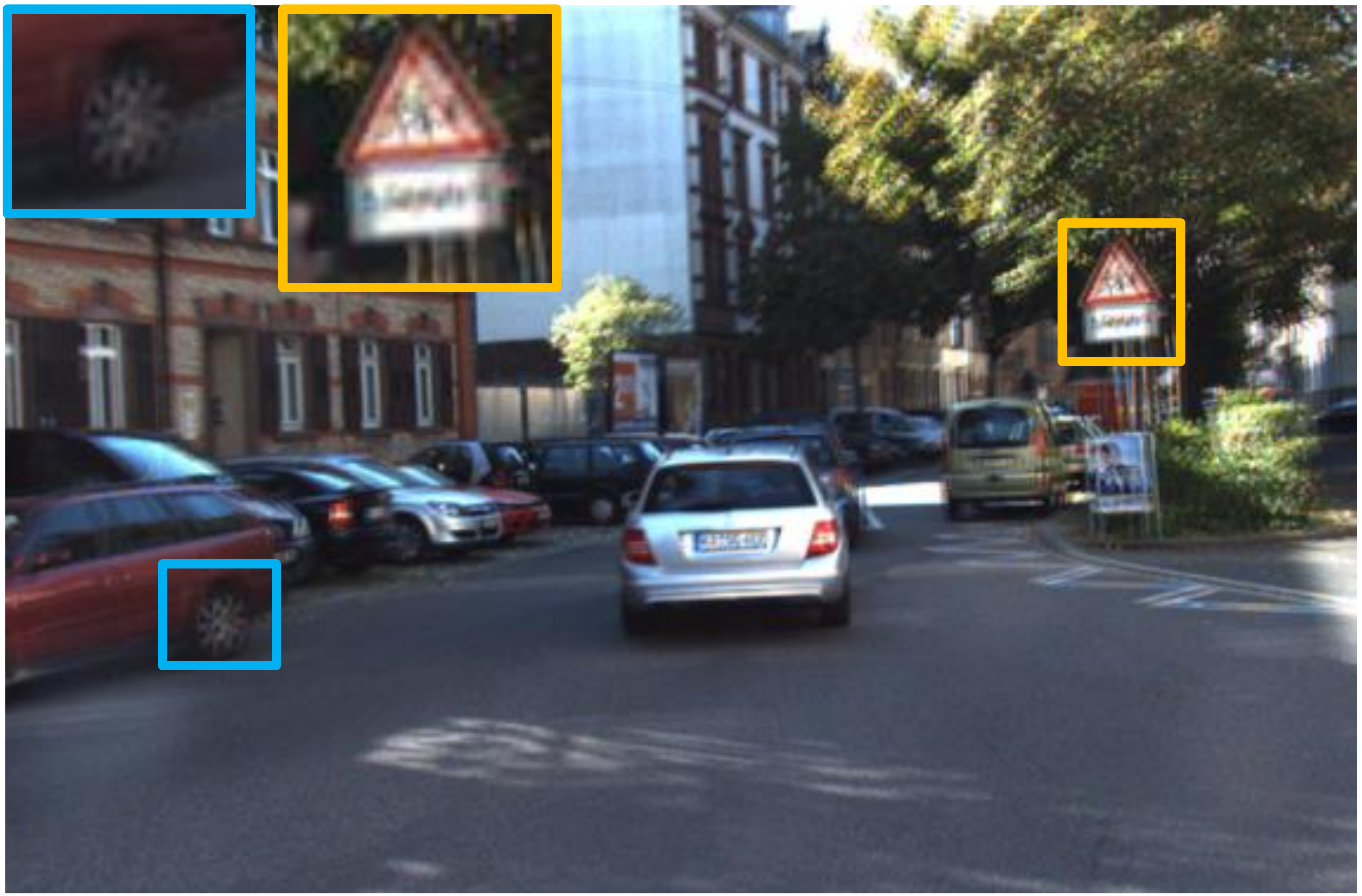}
&\hspace{0.0cm}
\includegraphics[width=0.21\textwidth]{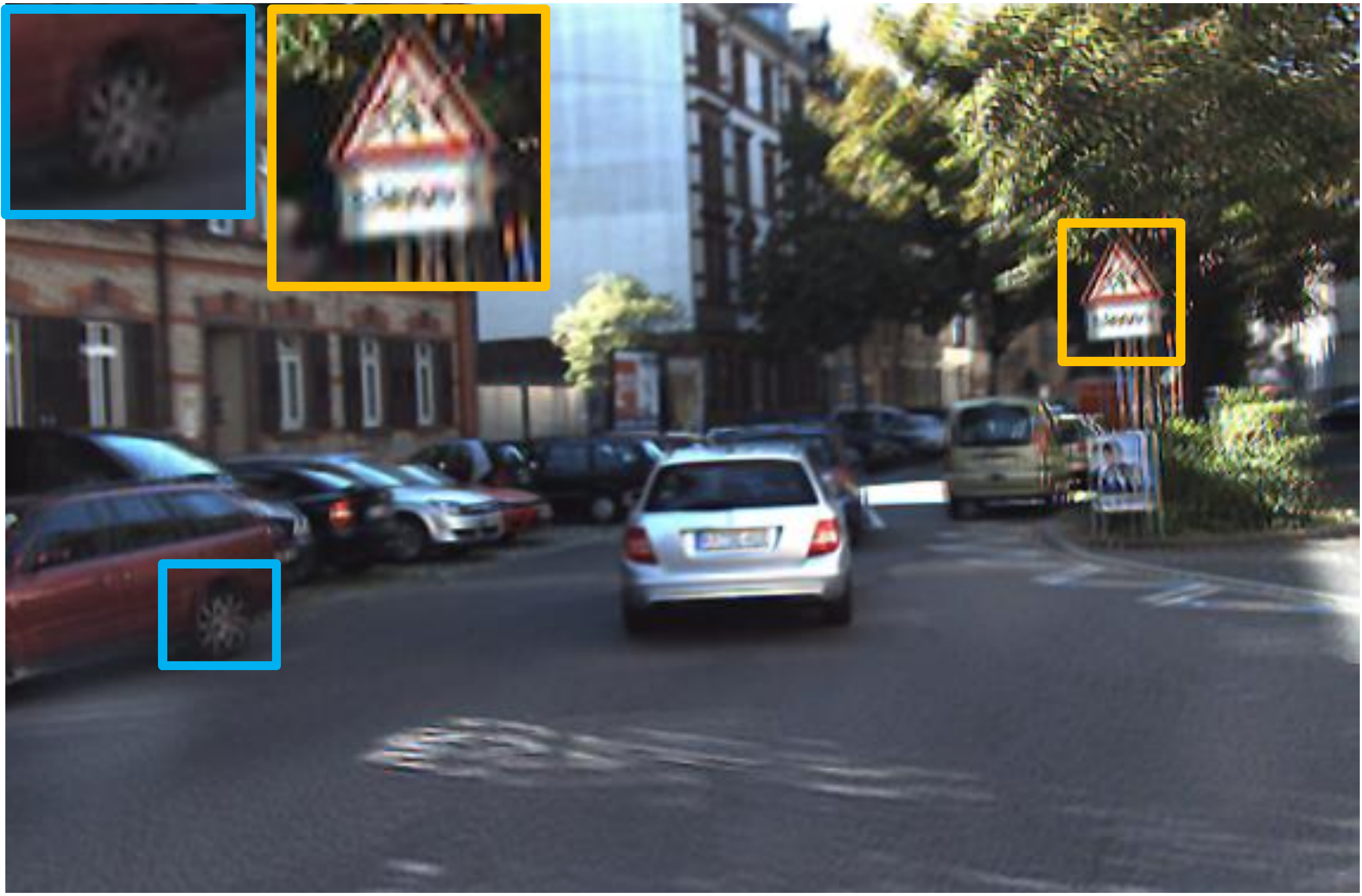}\\
(a) Blur Image & (b) Kim and Lee~\cite{hyun2015generalized} \\
\hspace{0.0cm}
\includegraphics[width=0.21\textwidth]{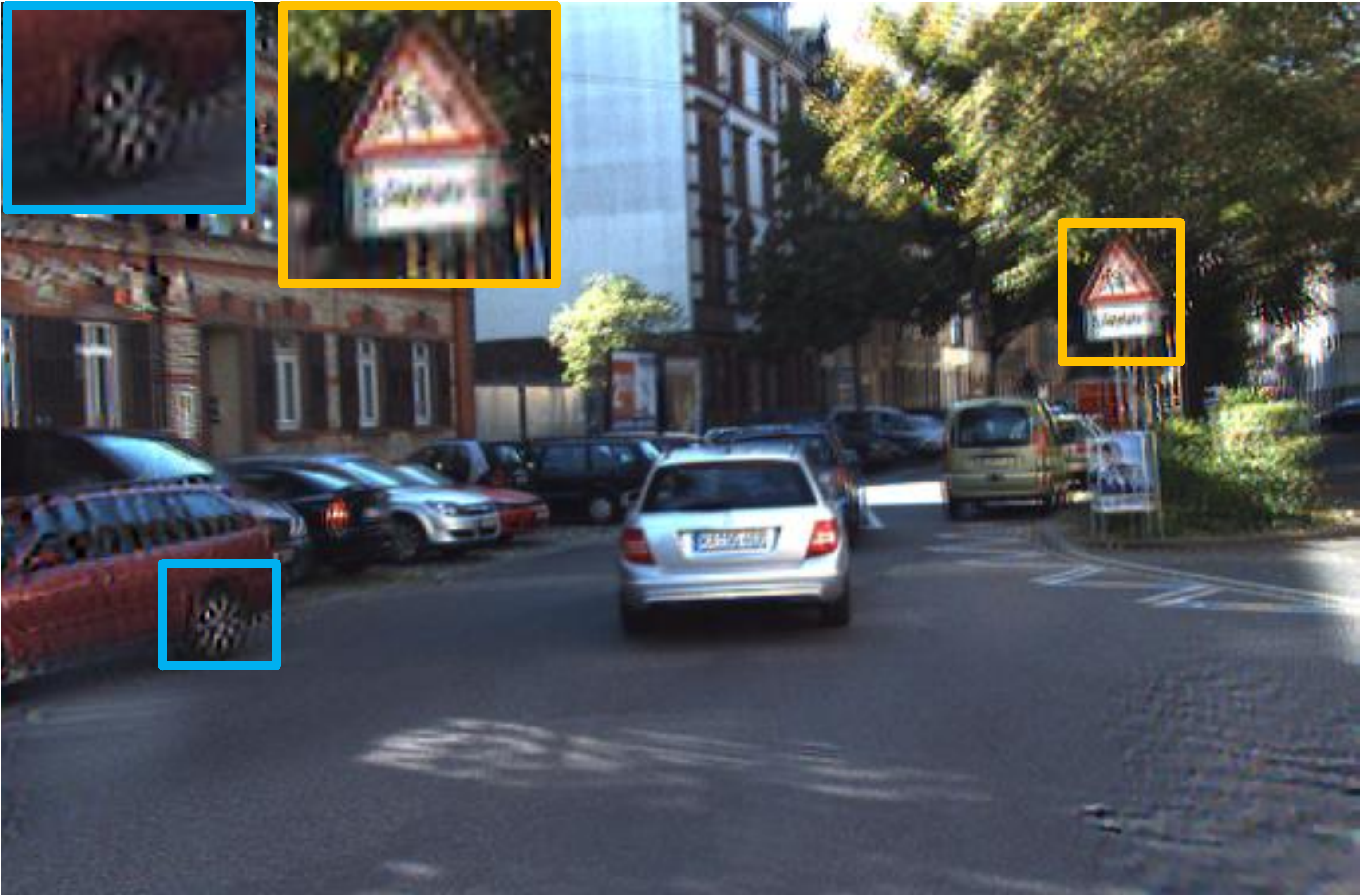}
&\hspace{0.0cm}
\includegraphics[width=0.21\textwidth]{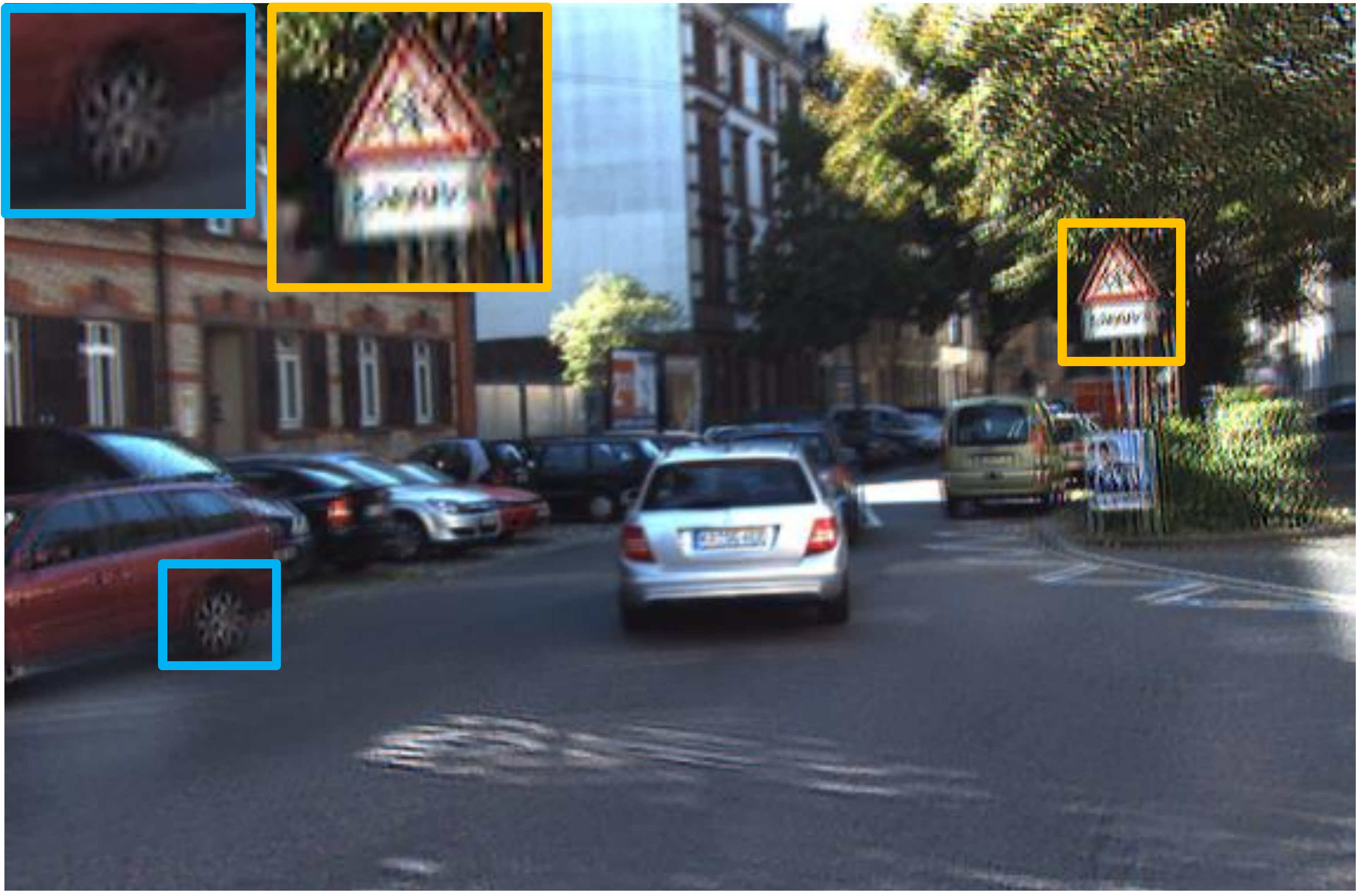}\\
((c) Sellent \etal \cite{sellent2016stereo} & (d) Ours
\end{tabular}
\end{small}
\caption{Quantitative evaluation on the KITTI dataset, where the blur images are generated by averaging three consecutive frames.}
\label{fig:averageblurmodel}
\end{center}
\end{figure}

\noindent\textbf{Results on another blur model:} We have also tested our method on another blur generation model, where the blurred image is an average of consecutive three frames \cite{gong2016motion,nah2016deep}. The results are shown in Table~\ref{kittiaver_psnr} and Fig.~\ref{fig:averageblurmodel} respectively, where our method again achieves the best performance.
\begin{table}[h]\footnotesize
\centering
\caption{Quantitative evaluation on the KITTI dataset where the blur images are generated by averaging three consecutive frames.}
\label{kittiaver_psnr}
\begin{tabular}{c|c|c|c}
\hline
             & Kim \cite{hyun2015generalized} & Sellent \etal \cite{sellent2016stereo}  & Ours     \\ \hline
PSNR(dB)     & 23.21                          & 23.31                                  & \bf23.89 \\ \hline
SSIM         & 0.781                          & 0.764                                  & \bf0.786 \\ \hline
\end{tabular}
\end{table}

\noindent\textbf{Runtime:}
In all experiments, we simultaneously compute two direction scene flow and restoration six blur images. Our MATLAB implementation with C++ wrappers requires a total runtime of 40 minutes for processing one scene(6 images, 3 iterations) on a single i7 core running at 3.6 GHz.

\section{Conclusion}
In this paper, we present a joint optimization framework to tackle the challenging task of stereo video deblurring where scene flow estimation and video deblurring are solved in a coupled manner. Under our formulation, the motion cues from scene flow estimation and blur information could reinforce each other, and produce superior results than conventional scene flow estimation or stereo deblurring methods. We have demonstrated the benefits our framework on extensive synthetic and real stereo sequences.

\vspace{-0.1cm}
\section*{Acknowledgement}
\vspace{-0.1cm}
This work was supported in part by China Scholarship Council (201506290130), Australian Research Council (ARC) grants (DP150104645, DE140100180), and Natural Science Foundation of China (61420106007, 61473230, 61135001), and Aviation fund of China (2014ZC5303). We thank all reviewers for their valuable comments.

{\small
\bibliographystyle{plain}
\bibliography{Flow_Deblur_Reference}
}

\end{document}